\documentclass{article} %
\usepackage{iclr2025_conference,times}

\usepackage{amsmath,amsfonts,bm}

\def\eqref#1{equation~\ref{#1}}

\def\1{\bm{1}}

\DeclareMathAlphabet{\mathsfit}{\encodingdefault}{\sfdefault}{m}{sl}
\SetMathAlphabet{\mathsfit}{bold}{\encodingdefault}{\sfdefault}{bx}{n}

\usepackage{hyperref}
\usepackage{url}
\usepackage{graphicx} 
\usepackage{float}
\usepackage{booktabs}
\usepackage[accsupp]{axessibility}  %
\usepackage{epsfig}
\usepackage{graphicx}
\usepackage{amsmath}
\usepackage{amssymb}
\usepackage{mathtools}
\usepackage{array}
\usepackage{soul}
\newcolumntype{P}[1]{>{\centering\arraybackslash}p{#1}}
\usepackage{pifont}%
\usepackage{multirow}
\usepackage{nicefrac}       %
\usepackage{microtype}  
\usepackage{tabularx}
\usepackage{array}
\usepackage{enumitem}
\usepackage{afterpage}
\usepackage{scalerel,xparse}
\usepackage{capt-of,etoolbox}
\usepackage{booktabs}       %
\usepackage{amsfonts}       %
\usepackage{nicefrac}       %
\usepackage{microtype}      %
\usepackage{xcolor}         %
\usepackage{graphics}
\usepackage{xfrac}
\usepackage{makecell}
\usepackage{tabularx}
\usepackage{tablefootnote}
\usepackage{ulem}
\usepackage{xspace}
\usepackage{wrapfig}

\usepackage{tikz}
\usepackage{subfigure}
\usepackage{pgfplots}
\pgfplotsset{compat=1.17}

\newcommand{\good}{\textcolor{green}{$\checkmark$}}
\newcommand{\medium}{\textcolor{orange}{$\sim$}}
\newcommand{\bad}{\textcolor{red}{$\times$}}

\newcommand{\augunet}{\mbox{\textsc{SD\textsubscript{aug}}} \xspace}
\newcommand{\augunetnospace}{\mbox{\textsc{SD\textsubscript{aug}}}}
\newcommand{\augold}{\mbox{\textsc{SD1-5\textsubscript{aug}}} \xspace}
\usepackage[capitalize]{cleveref}
\crefname{section}{Sec.}{Secs.}
\Crefname{section}{Section}{Sections}
\crefname{table}{Tab.}{Tabs.}
\Crefname{table}{Table}{Tables}

\title{Generative Models: What Do They Know? Do They Know Things? Let's Find Out!}

\author{Xiaodan Du\textsuperscript{1}\quad Nicholas Kolkin\textsuperscript{2} \quad Gregory Shakhnarovich\textsuperscript{1} \quad Anand Bhattad\textsuperscript{1} 
\\
\textsuperscript{1} Toyota Technological Institute at Chicago,
\textsuperscript{2} Adobe Research 
\\
\textsuperscript{1} \texttt{\{xdu,greg,bhattad\}@ttic.edu}, 
\textsuperscript{2} \texttt{kolkin@adobe.com}
}

\iclrfinalcopy %
\begin{document}

\makeatletter
\g@addto@macro\@maketitle{
    \begin{figure}[H]
    \scriptsize
        \setlength{\linewidth}{0.9\textwidth}
        \setlength{\hsize}{\textwidth}
        \vspace{-5mm}
    \centering
\includegraphics[width=\linewidth]{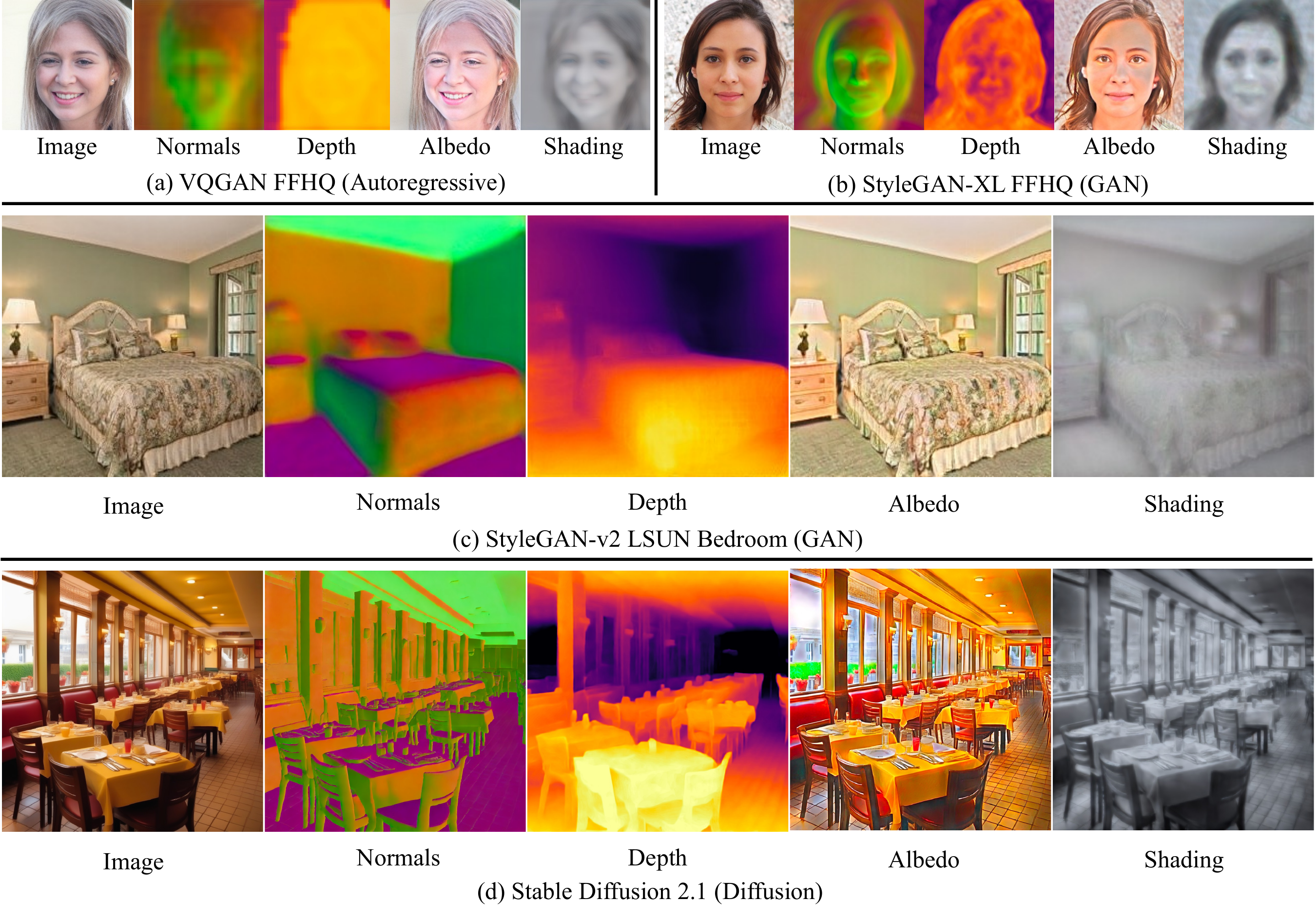}
\vspace{-13pt}
\caption{Generative models of various types---Autoregressive, GANs and Diffusion---implicitly encode intrinsic images as a by-product of generative training. We show that a model-agnostic approach, Low-Rank Adaptation (LoRA), can recover this intrinsic knowledge. Applying targeted, lightweight LoRA to attention layers in VQGAN (a) and Stable Diffusion (d), and affine layers in StyleGAN (b and c), allows us to recover fundamental intrinsic images---normals, depth, albedo and shading---directly from the models' learned representations, eliminating the need for additional task-specific decoding heads or layers.
}
\vspace{-10pt}
\label{fig:teaser2}
\end{figure}
}

\maketitle

\begin{abstract}
Generative models excel at mimicking real scenes, suggesting they might inherently encode important intrinsic scene properties. In this paper, we aim to explore the following key questions: (1) What intrinsic knowledge do generative models like GANs, Autoregressive models, and Diffusion models encode? (2) Can we establish a general framework to recover intrinsic representations from these models, regardless of their architecture or model type? (3) How minimal can the required learnable parameters and labeled data be to successfully recover this knowledge? (4) Is there a direct link between the quality of a generative model and the accuracy of the recovered scene intrinsics?

Our findings indicate that a small Low-Rank Adaptators (LoRA) can recover intrinsic images---depth, normals, albedo and shading---across different generators (Autoregressive, GANs and Diffusion) while using the same decoder head that generates the image. As LoRA is lightweight, we introduce very few learnable parameters (as few as 0.04\% of Stable Diffusion model weights for a rank of 2), and we find that as few as 250 labeled images are enough to generate intrinsic images with these LoRA modules. Finally, we also show a positive correlation between the generative model's quality and the accuracy of the recovered intrinsics through control experiments.
\end{abstract}
\section{Introduction}
\label{sec:intro}
Generative models can produce high-quality images that are almost indistinguishable from real-world photographs. They appear to profoundly understand the world, capturing object placement, appearance, and lighting conditions. Yet, it remains an open question how these models encode such detailed knowledge, and whether representations of scene intrinsics—such as depth, normals, albedo and shading—exist within these models and can be explicitly recovered, or if these models manipulate abstract representations of the world to generate these images.

\noindent\textbf{Why study intrinsic knowledge embedded in generative models?}
Understanding how generative models produce realistic outputs allows us to model the physical world better computationally, improving both image generation and interpretation across various applications. As we demonstrate in this paper, most generative models inherently encode intrinsic image representations as a byproduct of training on large-scale image data, and these can be easily recovered. By retrieving this embedded knowledge, we can enhance downstream tasks such as relighting, object compositing, and image editing without the need for large labeled datasets or extensive retraining of the models. 

Recent work has begun to study this question. 
\citet{bhattad2023stylegan} demonstrated that StyleGAN can encode important scene intrinsics. Similarly, 
\citet{zhan2023does} showed that diffusion models can understand 3D scenes in terms of geometry and shadows. 
\citet{chen2023beyond} found that Stable Diffusion's internal activations encode depth and saliency maps that can be extracted with linear probes. Three independent efforts~\citep{luo2023diffusion, tang2023emergent, hedlin2023unsupervised} discovered correspondences in diffusion models. However, these insights often pertain to specific models, leaving a gap in our understanding of whether such encoding is ubiquitous across generative architectures.

\noindent\textbf{Why study different models?}
While diffusion models~\citep{rombach2022high,saharia2022photorealistic}, have gained significant attention, other model types like GigaGAN~\citep{kang2023gigagan}, CM3leon~\citep{yu2023scaling}, and Parti~\citep{yu2022scaling} have shown they can produce similarly high-quality images. By investigating this wide range of models, we can create a general framework that not only applies to current generative models but is also adaptable to future developments and emerging architectures. To the best of our knowledge, this paper is the first to study generative models of all types.

\noindent\textbf{Why develop a general approach?} A general approach ensures broad applicability to emerging generative models. In this context, we find LoRA~\citep{hu2021lora} to be highly effective. LoRA can easily recover scene intrinsics across diverse architectures with minimal parameter updates and data. This general method lays the groundwork for future research that can build on our findings to explore intrinsic knowledge in new generative models. It is important to note that any approach capable of being applied to all generative models with minimal or no parameter updates and minimal data requirements is a reasonable and valid choice. {\it While we have identified one such method (LoRA) in this work, many others could also recover intrinsic representations across diverse generative models.}

\noindent\textbf{Why do we need minimal modification or minimal data to recover this knowledge?} Ideally, we recover intrinsic knowledge without any new learning, revealing what the model already ``knows.'' But achieving this purely with no learning is hard and non-trivial. Thus, we limit our approach to minimal fine-tuning, using little labeled data to avoid introducing new knowledge to the model. 

Previous approaches, such as \citet{bhattad2023stylegan}, have found codes in StyleGAN’s latent space for each intrinsic image, but such disentangled spaces have not yet been identified in models like diffusion and autoregressive models. Recent depth extraction from diffusion models often involves fine-tuning the entire model~\citep{zhao2023unleashing, ke2023repurposing} or applying linear probing~\citep{chen2023beyond}. Fine-tuning alters the model significantly, transforming it into a new version and potentially compromising its original image-generating capabilities. This raises the question of whether the depth perception was an innate quality of the model or a product of the fine-tuning process. A drawback of linear probing lies in probing each layer independently. As we show linear probes perform poorly, and our application of LoRA suggests that intrinsic information is distributed throughout the network. 

\noindent\textbf{Why analyze the correlation between recovered intrinsics and improved generative models?}
If higher-quality generative models consistently produce better intrinsic images, this suggests an alternative paradigm for improving these models. Instead of blindly scaling up with more data and parameters, we could focus on enhancing the model’s ability to capture and recover intrinsic properties. This approach could lead to more efficient improvements in model performance, driven by the quality of the intrinsic knowledge embedded within the model.

We find positive correlations in our experiments between the quality of recovered intrinsics and the improvements in generative model performance. Specifically, we observe this in Stable Diffusion versions 1.1, 1.2 and 1.5, as well as in improved face generators from various GAN and Autoregressive models, as measured by FID. A visual illustration of this correlation is in \cref{fig:corr_plot}. These results indicate that higher-quality generators tend to produce more accurate intrinsic representations.

Our contributions are showing that generative models encode intrinsic images across different architectures, including GANs, Autoregressive models and Diffusion models. Our findings are in \cref{tab:summary} and elaborated in \cref{sec:Exp}. We find a general approach using LoRA to recover these intrinsics, which are competitive, with minimal fine-tuning and data. This method obtains these properties using the same output head as the original image generation task.
Through control experiments, we find a positive correlation between the quality of the generative model and the accuracy of the recovered intrinsics, suggesting that better models naturally produce better intrinsic representations(\cref{fig:corr_plot}). This offers a new paradigm for model improvement beyond just scaling data and parameters.

\begin{figure*}
\centering
\LARGE
\vspace{-3pt}
\subfigure{
    \begin{tikzpicture}[scale=0.4]
        \begin{axis}[
            title={Surface Normal},
            xlabel={FID},
            xlabel style={at={(axis description cs:0.5,-0.1)}, anchor=north},
            ylabel={Error},
            xmin=0, xmax=12,
            ymin=12, ymax=22,
            xtick={0,2,4,6,8,10, 12},
            xticklabel style={ticklabel pos=lower},
            extra x ticks={2.19,3.62,9.6},
            extra x tick labels={\large SG-XL, \large SGv2, \large VQGAN},
            extra x tick style={
                grid=none,
                ticklabel pos=lower,
                major tick length=0pt,
                xticklabel style={rotate=75, anchor=east,
                xshift=-13pt,
                yshift=-5pt
                }},
            ytick={12,14,16,18,20,22},
            legend pos=south east,
            ymajorgrids=true,
            xmajorgrids=true,
            legend cell align={left},
            width=\textwidth,
            height=0.833333\textwidth,
            clip mode=individual,
        ]
        \addplot[
            color=blue,
            mark=*,
            ]
            coordinates {
            (2.19,15.28) (3.62,16.93) (9.6,19.97)
            };
        \addlegendentry{Mean Error}
    
        \addplot[
            color=red,
            mark=*,
            ]
            coordinates {
           (2.19,18.07) (3.62,19.60) (9.6,20.97)
            };
        \addlegendentry{Median Error}
        
        \addplot[
            color=yellow,
            mark=*,
            ]
            coordinates {
            (2.19,12.63) (3.62,13.87) (9.6,16.33)
            };
        \addlegendentry{L1 Error}
        
        \draw[dotted, thick] (axis cs:2.19,11.5) -- (axis cs:2.19,18.07);
        \draw[dotted, thick] (axis cs:3.62,11.5) -- (axis cs:3.62,19.6);
        \draw[dotted, thick] (axis cs:9.6,11.5) -- (axis cs:9.6,20.97);
    
        \end{axis}
    \end{tikzpicture}
}
\subfigure{
    \begin{tikzpicture}[scale=0.4]

        \begin{axis}[
            title={Depth},
            xlabel={FID},
            xlabel style={at={(axis description cs:0.5,-0.1)}, anchor=north},
            ylabel={Percentage},
            xmin=0, xmax=12,
            ymin=0, ymax=100,
            xtick={0,2,4,6,8,10, 12},
            xticklabel style={ticklabel pos=lower},
            extra x ticks={2.19,3.62,9.6},
            extra x tick labels={\large SG-XL, \large SGv2, \large VQGAN},
            extra x tick style={
                grid=none,
                ticklabel pos=lower,
                major tick length=0pt,
                xticklabel style={rotate=75, anchor=east,
                xshift=-13pt,
                yshift=-5pt
                }},
            ytick={0,20,40,60,80,100},
            yticklabel={\pgfmathprintnumber{\tick}\%},
            axis y line*=right,
            legend pos=south east,
            ymajorgrids=true,
            xmajorgrids=true,
            legend cell align={left},
            width=\textwidth,
            height=0.833333\textwidth,
            clip mode=individual,
        ]
        \addplot[
            color=cyan,
            mark=*,
            ]
            coordinates {
            (2.19,6.13) (3.62,9.26) (9.6,37.67)
            };
        \addlegendentry{Percentage $\delta > 1.25$}
        
        \end{axis}

        \begin{axis}[
            ylabel={Error},
            xmin=0, xmax=12,
            ymin=0, ymax=0.2,
            ytick={0.00,0.04,0.08,0.12,0.16,0.20},
            scaled y ticks=false,
            axis x line=none, %
            axis y line*=left, %
            scaled y ticks=false, %
            yticklabel={\pgfmathprintnumber[fixed, precision=2]{\tick}},
            clip mode=individual,
            legend pos=north west,
            height=0.8333333\textwidth,
            width=\textwidth,
        ]
        \addplot[
            color=purple,
            mark=*,
            ]
            coordinates {
           (2.19,0.1337) (3.62,0.1530) (9.6,0.1819)
            };
        \addlegendentry{Root Mean Squared Error}
        \draw[dotted, thick] (axis cs:2.19,-0.01) -- (axis cs:2.19,0.1337);
        \draw[dotted, thick] (axis cs:3.62,-0.01) -- (axis cs:3.62,0.1530);
        \draw[dotted, thick] (axis cs:9.6,-0.01) -- (axis cs:9.6,0.1819);
        \end{axis}
        
    \end{tikzpicture}
}
\vspace{-17pt}
\caption{FID vs. metrics of intrinsics recovered from different generative models traind on FFHQ. Enhancements in image generation quality correlate positively with intrinsic recovery capabilities.}
\label{fig:corr_plot}
\vspace{-12pt}
\end{figure*}
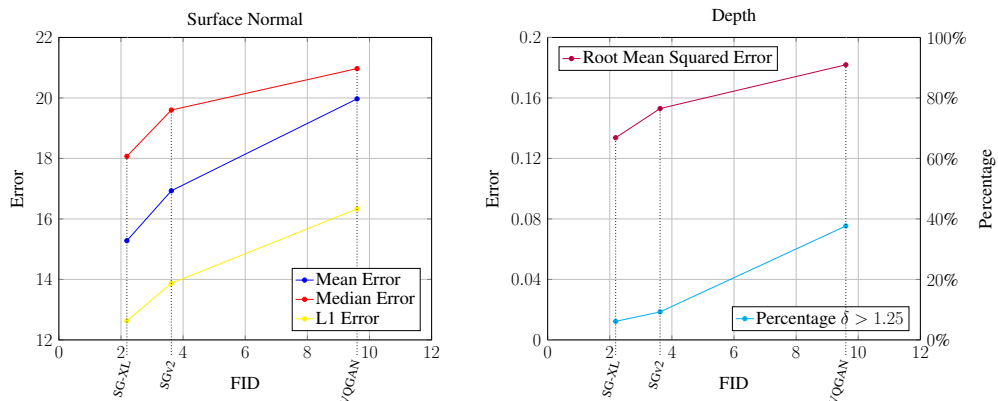
\begin{table}[t!]
    \centering
        \caption{%
    Summary of scene intrinsics found across different generative models without changing generator head. \good: Intrinsics can be recovered with high quality. \medium: Intrinsics cannot be recovered with high quality. \bad: Intrinsics cannot be recovered.
    }        
    \resizebox{0.9\linewidth}{!}{%
    \begin{tabular}{c  c  c  c c c c  }
        \toprule
        Model & Pretrain Type & Domain & Normal & Depth  & Albedo & Shading  \\
        \midrule
        VQGAN~\citep{esser2020taming} & Autoregressive & FFHQ & \medium & \medium & \good & \good \\
        SG-v2~\citep{karras2020analyzing} & GAN & FFHQ & \good & \medium & \good & \good  \\
        SG-v2~\citep{yu2021dual} & GAN & LSUN Bed & \good & \good & \good & \good \\
        SG-XL~\citep{sauer2022stylegan} & GAN & FFHQ & \good & \medium  & \good & \good   \\
        SG-XL~\citep{sauer2022stylegan} & GAN & ImageNet & \bad & \bad  & \bad & \bad   \\   
        SD-UNet (single-step)~\citep{rombach2022high} & Diffusion & Open & \good & \good  & \good & \good \\
        \augunet (multi-step)~\citep{rombach2022high}  & Diffusion & Open & \good & \good  & \good & \good \\

        \bottomrule
    \end{tabular}
    }
 \vspace{-15pt}
    \label{tab:summary}
\end{table}

\section{Related Work}
\label{sec:related}
{\bf Generative Models:} 
Generative Adversarial Networks (GANs)~\citep{goodfellow2014generative} have been widely used for generating realistic images. Variants like StyleGAN~\citep{karras2019style}, StyleGAN2~\citep{karras2020analyzing} and GigaGAN~\citep{kang2023gigagan} have pushed the boundaries in terms of image quality and control. Some work has explored the interpretability of GANs~\citep{bau2018gan, bhattad2023stylegan}, but little is known about their ability to capture scene intrinsics. 

Diffusion models~\citep{vincent2011connection, gutmann2010noise} are popular at the moment for generative tasks ~\citep{karras2022elucidating,ho2020denoising,rombach2022high}. These models have been shown to understand complex scene intrinsics like geometry and shadows~\citep{zhan2023does}, but the generalizability of this understanding across different scene intrinsics is largely unexplored. 

Autoregressive models~\citep{van2016pixel, van2016conditional} generate images pixel-by-pixel, offering fine-grained control but at the cost of computational efficiency. VQ-VAE-2~\citep{razavi2019generating} and VQGAN~\citep{esser2020taming} have combined autoregressive models with vector quantization to achieve high-quality image synthesis. While these models are powerful, their ability to capture and represent scene intrinsics is yet to be investigated.

\noindent{\bf Intrinsic Image Recovery:} 
\citet{barrow1978recovering} highlighted several fundamental scene intrinsics including depth, albedo, shading, and surface normals. A large body of work has focused on extracting some related properties like depth and normals, from images~\citep{eigen2014depth, long2015fully, eftekhar2021omnidata, kar20223d, ranftl2021vision, bhat2023zoedepth} using labeled data.  Labeled albedo and shading are hard to find and as the recent review in~\citet{forsyth2021intrinsic} shows, methods involving little or no learning have remained competitive. However, these methods often rely on supervised learning and do not recover intrinsic images from generative models. 

Many recent studies have used generative models as pre-trained feature extractors or scene prior learners. They use generated images to enhance downstream discriminative models, fine-tune the original generative model for a new task, learn new layers or decoders to produce desired scene intrinsics~\citep{abdal2021labels4free, jahanian2021generative, zhang2021datasetgan, li2021semantic, RGBDGAN, bao2022generative, xu2023odise, sariyildiz2023fake,
zhao2023unleashing, ke2023repurposing}. InstructCV~\citep{gan2023instructcv} executes computer vision tasks via natural language instructions, abstracting task-specific design choices. However, it requires re-training of the entire diffusion model. In contrast, we show that many generative models capture intrinsic image knowledge implicitly and do not require specialized training to recover this information.

\noindent{\bf Knowledge in Generative Models:} Several studies have explored the extent of StyleGAN's knowledge, particularly for 3D information about faces~\citep{pan20202d, zhang2020image}. 
\citet{yang2021semantic} show GANs encode hierarchical semantic information across different layers. Further research has demonstrated that manipulating offsets in StyleGAN can lead to effective relighting of images~\citep{bhattad2023StylitGAN, bhattad2023make} and extraction of scene intrinsics~\citep{bhattad2023stylegan}. 
\citet{chen2023beyond} found internal activations of the LDM encode linear representations of both depth data and a salient-object / background distinction. 
\citet{wu2023datasetdm} also demonstrate rich latent codes of diffusion models can be easily mapped to annotations with small amount of training samples. \citet{tang2023emergent, luo2023diffusion, hedlin2023unsupervised} found correspondence emerges in image diffusion models.~
\citet{sarkar2023shadows} showed generative models fail to replicate projective geometry.

\citet{luo2023readoutguidance} explored training task-specific ``readout'' networks to extract signals like pose, depth, and edges from feature maps in Stable Diffusion models for controlling image generation. 
Our goals are different: We are interested in understanding intrinsic knowledge encoded in these models, while the aim of \citet{luo2023readoutguidance} is controlling image generation. Our use of LoRA offers notable advantages in parameter efficiency: itis approximately 5 times more parameter-efficient than readout networks in their application to SD v1-5 (compare 8.5M vs 1.59M). Lastly, the broad applicability of ``readout'' networks across various generative model types remains uncertain.

A concurrent work~\citet{lee2023dmp} applies a LoRA-like approach to adapt a pre-trained diffusion model for dense semantic tasks. Our work differs from theirs in several aspects: First, their goal is to use pre-trained diffusion models as strong priors for dense prediction. Second, their tasks are within restricted domains, such as bedrooms. Finally, they do not extend to the wide range of generative models our study explores. Our paper not only demonstrates intrinsic knowledge encoded in different architectures but also explores its application in a diverse scene contexts including real images.

\section{Approach}\label{sec:ilora}
A  generative model $G$ maps noise/conditioning information $z$ to an RGB image $G(z) \in \mathbb{R}^{H\times W\times 3}$. We add to $G$ with a small set of parameters $\theta$ that allow us to produce, using the same architecture as $G$, an image-like map with up to three channels, representing scene intrinsics like surface normals.

\begin{figure}[t!]
\centering
\includegraphics[width=0.75\linewidth]{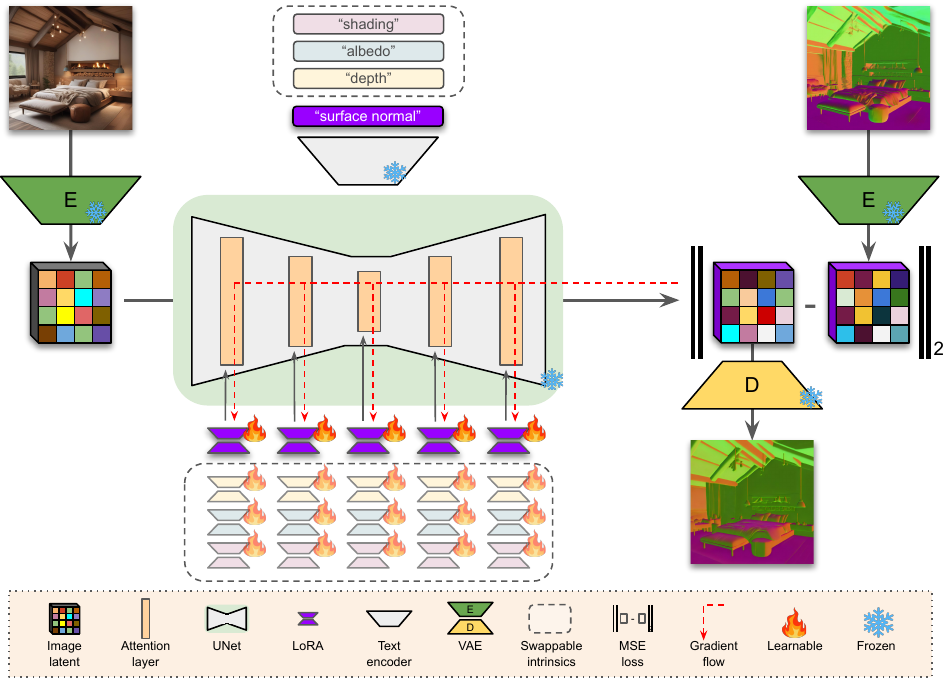}
\vspace{-10pt}
\caption{Overview of our framework applied to Stable Diffusion's UNet in a single-step manner. We adopt an efficient fine-tuning approach, low-rank adaptors (LoRA) corresponding to key feature maps --  attention matrices -- to reveal scene intrinsics. Distinct adaptors are optimized for each intrinsic ($\color{violet}{\textbf{\textit{violet}}}$ adaptors for surface normals; swappable with other intrinsics). We use a few labeled examples for this fine-tuning and directly obtain scene intrinsics using the \underline{same} decoder that generates images, circumventing the need for specialized decoders or comprehensive model re-training. 
}
\label{fig:lora_pipeline}
\vspace{-10pt}
\end{figure}

\noindent\textbf{Our Framework.} We recover intrinsic properties of an image (such as depth) using a small number of labeled examples (image/depth map pairs) as supervision. In cases where we do not have access to the actual intrinsic properties, we use models trained on large datasets to generate estimated intrinsics (such as estimated depth for an image) as pseudo-ground truth, used as training targets for $G_\theta$. To optimize $\theta$ of $G_\theta$ using a pseudo-ground truth predictor $\Phi$, 
we minimize the objective:
\begin{equation} \label{eq:obj1}
\vspace{-2pt}
    \min_\theta \mathbb{E}_{z} [ d(G_\theta(z), \Phi(G(z))) ],
\end{equation}
where $d$ is a distance metric that depends on the intrinsics we wish to learn.

Diffusion models require special treatment since their input and output are with the same dimension. 
During inference, diffusion models repeatedly receive a noisy image as input. Thus instead of conditioning noise $z$ we feed an image $x$(generated or real) to a diffusion model $G$. In this case, given a real image $x$, our objective function becomes $
    \min_\theta \mathbb{E}_{x} [ d(G_\theta(x), \Phi(x))]
$. 

For surface normals $\Phi$ is Omnidatav2~\citep{kar20223d}. To generate pseudo ground truth for depth we use ZoeDepth~\citep{bhat2023zoedepth} as the predictor $\Phi$.  For Albedo and Shading $\Phi$ is Paradigms~\citep{forsyth2021intrinsic, bhattad2022cut}. For SG2, SGXL and VQGAN, $d$ in Eq.$ \ref{eq:obj1}$ is 
\begin{equation}
    d(x,y)= 1-cos(x,y) + \|x-y\|_1
\end{equation}
for normal and MSE for other intrinsics. For latent diffusion, there isn't a clear physical meaning to the relative angle of latent vectors in encoded normals, so we use the standard MSE for all intrinsics. 

We use LoRA, a parameter-efficient adaptation technique, to recover image intrinsics from generative models. LoRA introduces a low-rank weight matrix \( W^* \), which has a lower rank than the original weight matrix \( W \in \mathbb{R}^{d_1 \times d_2} \). This is achieved by factorizing \( W^* \) into two smaller matrices \( W^*_u \in \mathbb{R}^{d_1 \times d^*} \) and \( W^*_l \in \mathbb{R}^{d^* \times d_2} \), where \( d^* \) is chosen such that \( d^* \ll \min(d_1, d_2) \).
The output \( o \) for an input activation \( a \) is then given by:
\begin{equation}
o = Wa + W^* a = Wa + W^*_u W^*_l a.
\end{equation}
To preserve the original model's behavior at initialization, \( W^*_u \) is initialized to zero. 

\noindent{\bf Applying LoRA for diffusion models}, LoRA adaptors are learned atop cross-attention and self-attention layers. The UNet is utilized as a dense predictor, transforming an RGB input into intrinsics in one step. This approach, favoring simplicity and effectiveness, delivers superior quantitative results. Depending on the intrinsic of interest, the textual input varies among ``surface normal'', ``depth'', ``albedo'', or ``shading''. 
\cref{fig:lora_pipeline} shows our pipeline. For \textbf{GANs}, LoRA modules are integrated with the affine layers that map from w-space to s-space~\citep{wu2021stylespace}. In the case of \textbf{VQGAN, an autoregressive model}, LoRA is applied to the convolutional attention layers within the decoder.

\section{Experiments} \label{sec:Exp}
In this section, we outline our findings. \cref{sec:generality} and \cref{sec:efficiency} demonstrate LoRA's general applicability across generative models and efficiency in terms of parameters and labeling, respectively. In \cref{sec:control}, we conduct control experiments and discover a strong correlation between the quality of a generator and the accuracy of its recovered intrinsics (\cref{sec:control}). Additional ablation studies and baseline comparisons further confirm LoRA's robustness (\cref{sec:ablations}). Note: our analysis in \cref{sec:efficiency}
uses a single-step approach for intrinsic image recovery from stable diffusion. In \cref{sec:p2i}, we discuss the challenge of naively applying LoRA to a multi-step Stable Diffusion model. To address this, we propose a simple modification to the architecture
. 
We refer to this modified model as \augunetnospace.

\begin{figure*}[t!]
\centering
\scriptsize
  \setlength\tabcolsep{0pt}
  \renewcommand{\arraystretch}{0}
\begin{tabular}{cccccccccc}
&&\multicolumn{2}{c}{Surface Normals} & \multicolumn{2}{c}{Depth} & \multicolumn{2}{c}{Albedo} & \multicolumn{2}{c}{Shading} \\
\multirow{2}{*}{\rotatebox[origin=c]{90}{\parbox[c]{0.5cm}{\centering VQGAN}}} & 
\includegraphics[width=0.107\linewidth]{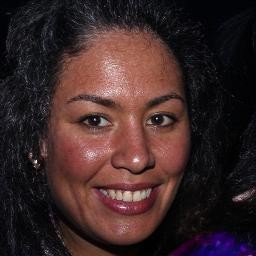} & 
\includegraphics[width=0.107\linewidth]{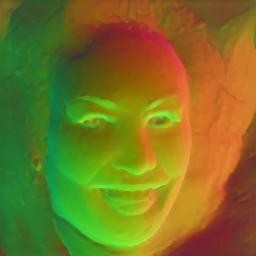} & 
\includegraphics[width=0.107\linewidth]{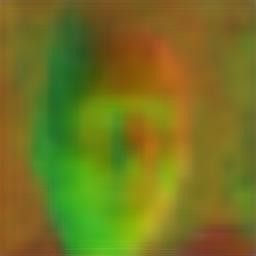} & 
\includegraphics[width=0.107\linewidth]{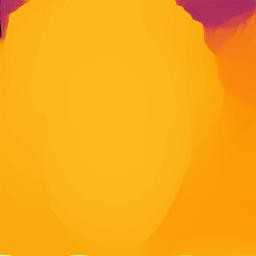} & 
\includegraphics[width=0.107\linewidth]{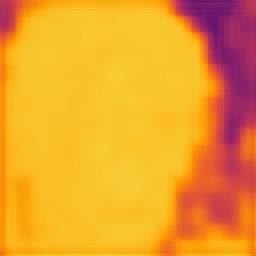} & 
\includegraphics[width=0.107\linewidth]{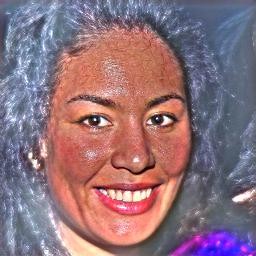} & 
\includegraphics[width=0.107\linewidth]{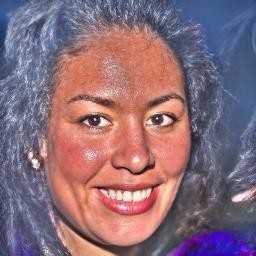} & 
\includegraphics[width=0.107\linewidth]{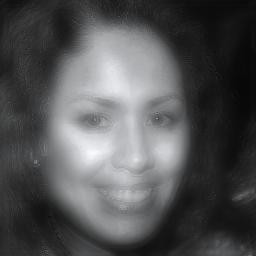} & 
\includegraphics[width=0.107\linewidth]{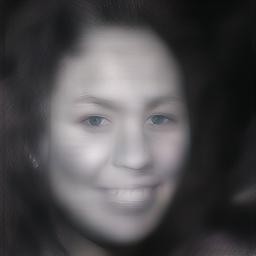}  \\
& \includegraphics[width=0.107\linewidth]{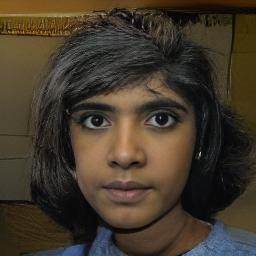} & 
\includegraphics[width=0.107\linewidth]{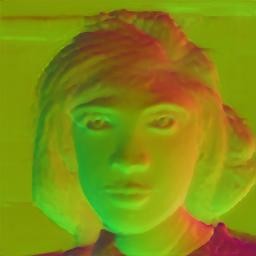} & 
\includegraphics[width=0.107\linewidth]{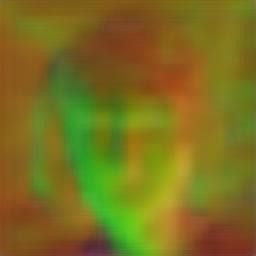} & 
\includegraphics[width=0.107\linewidth]{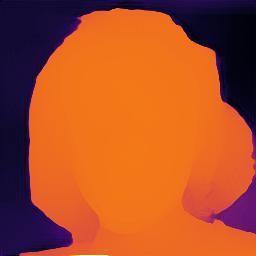} & 
\includegraphics[width=0.107\linewidth]{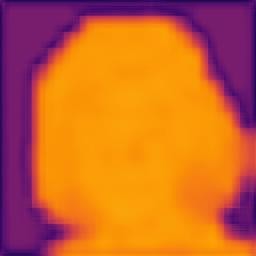} & 
\includegraphics[width=0.107\linewidth]{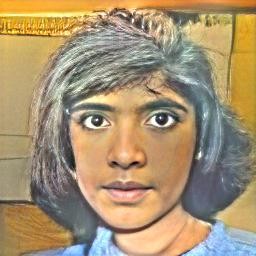} & 
\includegraphics[width=0.107\linewidth]{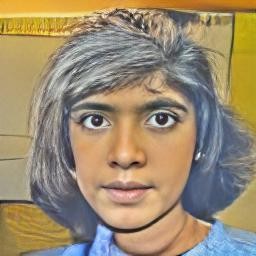} & 
\includegraphics[width=0.107\linewidth]{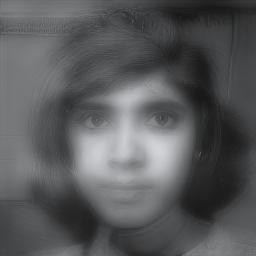} & 
\includegraphics[width=0.107\linewidth]{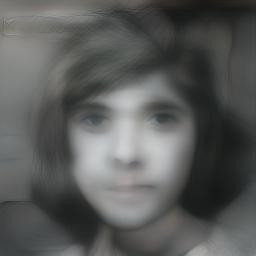}  \\
\midrule
\multirow{2}{*}{\rotatebox[origin=c]{90}{\parbox[c]{.7cm}{\centering StyleGANv2}}}
& \includegraphics[width=0.107\linewidth]{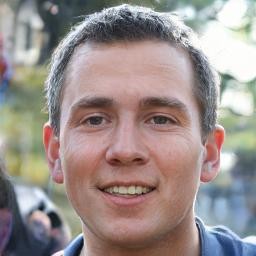} & 
\includegraphics[width=0.107\linewidth]{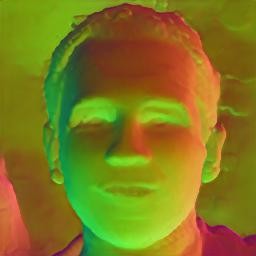} & 
\includegraphics[width=0.107\linewidth]{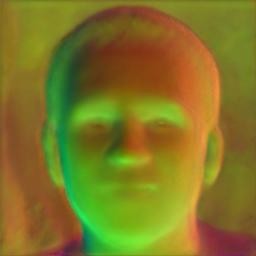} & 
\includegraphics[width=0.107\linewidth]{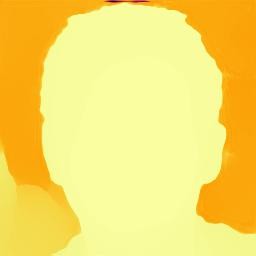} & 
\includegraphics[width=0.107\linewidth]{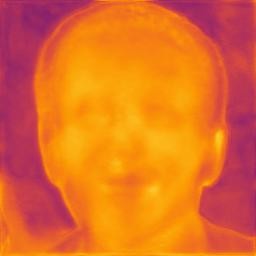} & 
\includegraphics[width=0.107\linewidth]{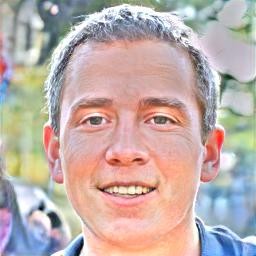} & 
\includegraphics[width=0.107\linewidth]{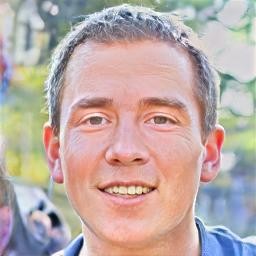} & 
\includegraphics[width=0.107\linewidth]{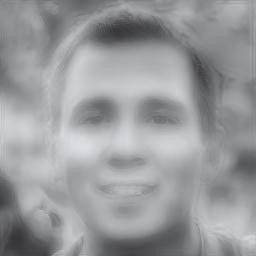} & 
\includegraphics[width=0.107\linewidth]{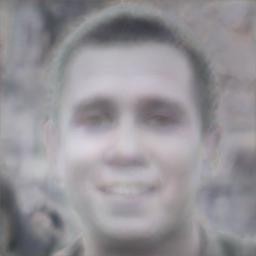} \\
& \includegraphics[width=0.107\linewidth]{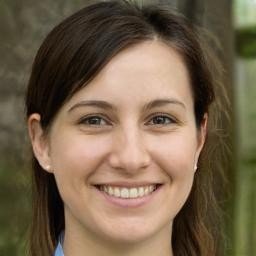} & 
\includegraphics[width=0.107\linewidth]{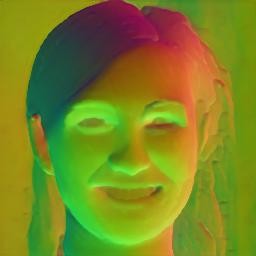} & 
\includegraphics[width=0.107\linewidth]{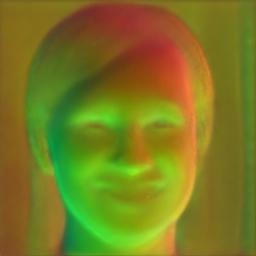} & 
\includegraphics[width=0.107\linewidth]{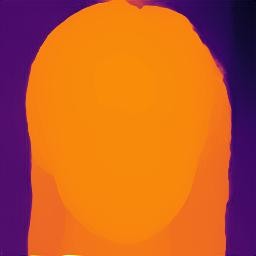} & 
\includegraphics[width=0.107\linewidth]{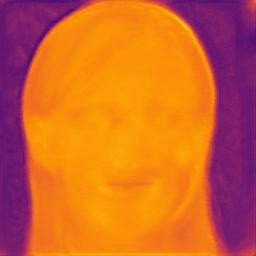} & 
\includegraphics[width=0.107\linewidth]{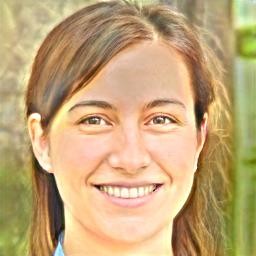} & 
\includegraphics[width=0.107\linewidth]{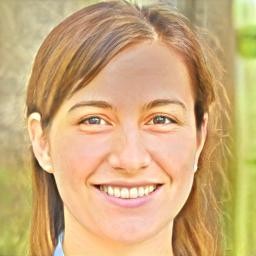} & 
\includegraphics[width=0.107\linewidth]{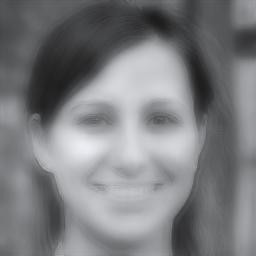} & 
\includegraphics[width=0.107\linewidth]{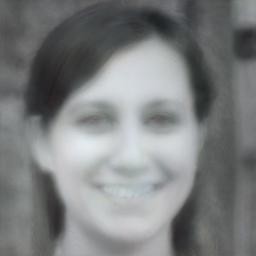} \\
\midrule
\multirow{2}{*}{\rotatebox[origin=c]{90}{\parbox[c]{.7cm}{\centering StyleGANXL}}} & \includegraphics[width=0.107\linewidth]{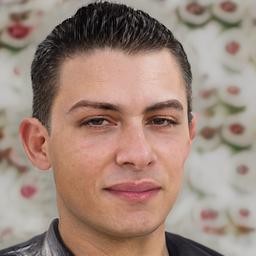} & 
\includegraphics[width=0.107\linewidth]{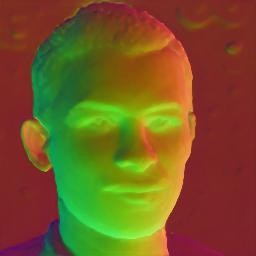} & 
\includegraphics[width=0.107\linewidth]{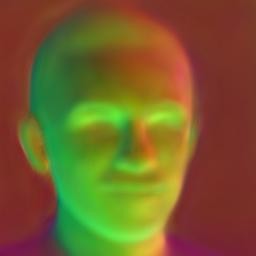} & 
\includegraphics[width=0.107\linewidth]{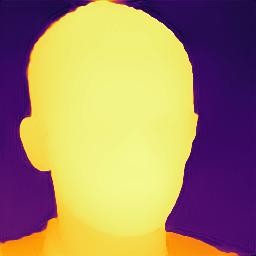} & 
\includegraphics[width=0.107\linewidth]{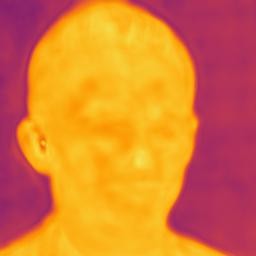} & 
\includegraphics[width=0.107\linewidth]{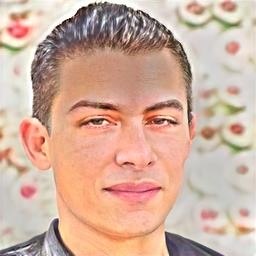} & 
\includegraphics[width=0.107\linewidth]{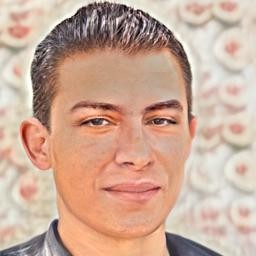} & 
\includegraphics[width=0.107\linewidth]{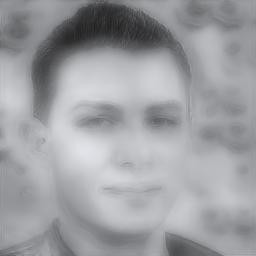} & 
\includegraphics[width=0.107\linewidth]{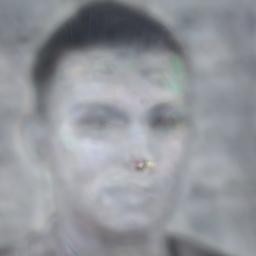} 
\\
& \includegraphics[width=0.107\linewidth]{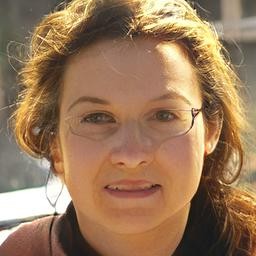} & 
\includegraphics[width=0.107\linewidth]{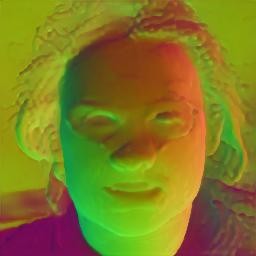} & 
\includegraphics[width=0.107\linewidth]{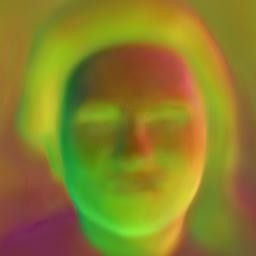} & 
\includegraphics[width=0.107\linewidth]{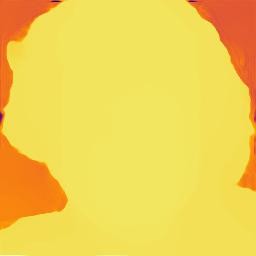} & 
\includegraphics[width=0.107\linewidth]{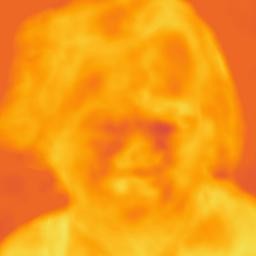} & 
\includegraphics[width=0.107\linewidth]{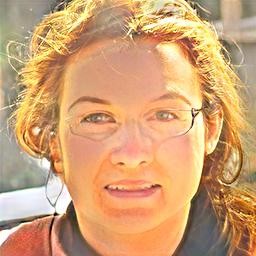} & 
\includegraphics[width=0.107\linewidth]{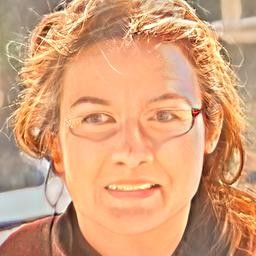} & 
\includegraphics[width=0.107\linewidth]{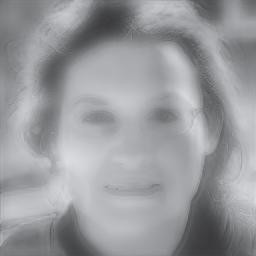} & 
\includegraphics[width=0.107\linewidth]{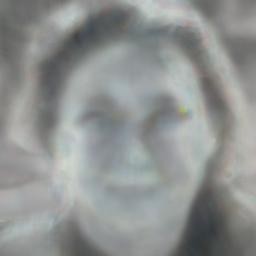} 
\\
\vspace{2pt}
\\
& \tiny Image & \multicolumn{1}{m{0.107\textwidth}}{\centering \tiny ~\citet{kar20223d}} & \tiny  \textbf{Recovered} & \multicolumn{1}{m{0.107\textwidth}}{\centering \tiny ~\citet{bhat2023zoedepth}} & \tiny \textbf{Recovered} & \multicolumn{1}{m{0.107\textwidth}}{\centering \tiny ~\citet{bhattad2022cut}} & \tiny  \textbf{Recovered} & \multicolumn{1}{m{0.107\textwidth}}{\centering \tiny  ~\citet{bhattad2022cut}} & \tiny  \textbf{Recovered} 
\end{tabular}
\vspace{-5pt}
\caption{ Scene intrinsics from VQGAN, StyleGAN-v2, and StyleGAN-XL -- trained on FFHQ dataset: The ``image'' column shows the synthetic images produced by each model. Other columns show four scene intrinsics predicted by a SOTA non-generative model and recovered by LoRA. 
}
\label{fig:generators_comparison}
\vspace{-5pt}
\end{figure*}

\begin{figure*}[t!]
\centering
\scriptsize
  \setlength\tabcolsep{0pt}
  \renewcommand{\arraystretch}{0}
\begin{tabular}{ccccccccc}
&\multicolumn{2}{c}{Surface Normals} & \multicolumn{2}{c}{Depth} & \multicolumn{2}{c}{Albedo} & \multicolumn{2}{c}{Shading} 
\\
 \includegraphics[width=0.11\textwidth]{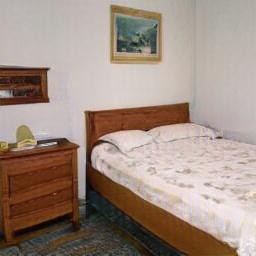} & 
\includegraphics[width=0.11\textwidth]{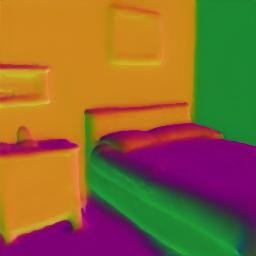} & 
\includegraphics[width=0.11\textwidth]{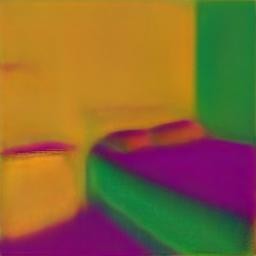} & 
\includegraphics[width=0.11\textwidth]{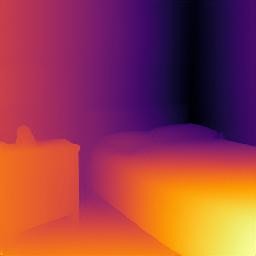} & 
\includegraphics[width=0.11\textwidth]{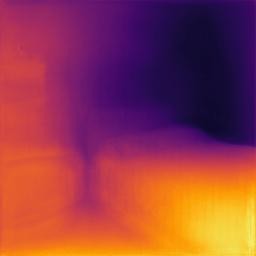} & 
\includegraphics[width=0.11\textwidth]{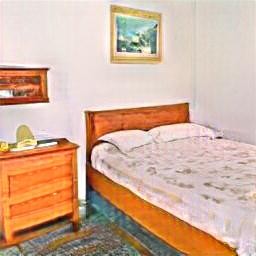} & 
\includegraphics[width=0.11\textwidth]{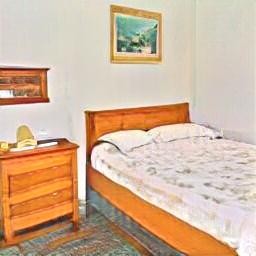} & 
\includegraphics[width=0.11\textwidth]{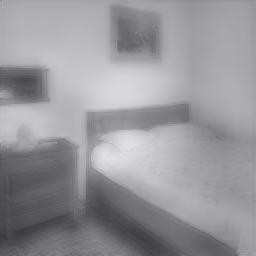} & 
\includegraphics[width=0.11\textwidth]{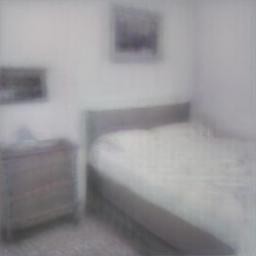} 
\\
\includegraphics[width=0.11\textwidth]{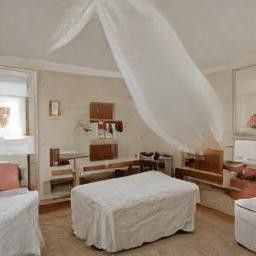} & 
\includegraphics[width=0.11\textwidth]{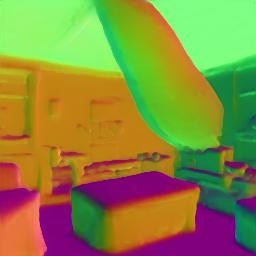} & 
\includegraphics[width=0.11\textwidth]{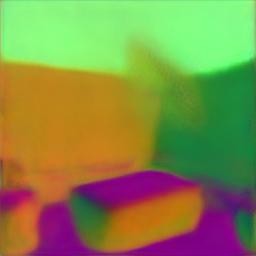} & 
\includegraphics[width=0.11\textwidth]{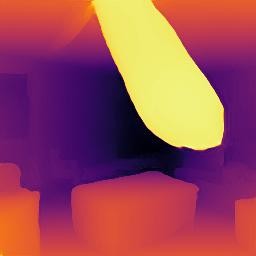} & 
\includegraphics[width=0.11\textwidth]{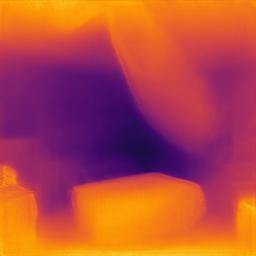} & 
\includegraphics[width=0.11\textwidth]{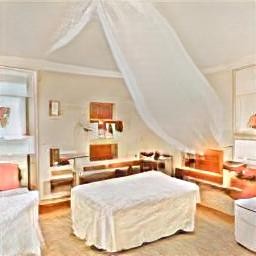} & 
\includegraphics[width=0.11\textwidth]{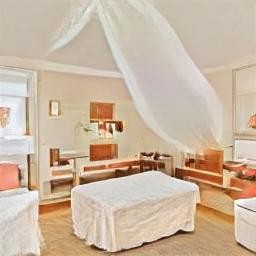} & 
\includegraphics[width=0.11\textwidth]{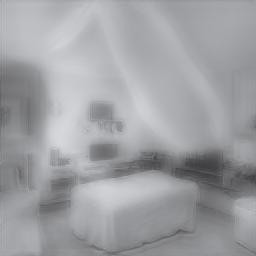} & 
\includegraphics[width=0.11\textwidth]{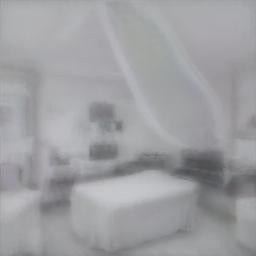} 
\\
\vspace{2pt}
\\
 \tiny Image & \multicolumn{1}{m{0.11\textwidth}}{\centering \tiny ~\citet{kar20223d}} & \tiny  \textbf{Recovered} & \multicolumn{1}{m{0.11\textwidth}}{\centering \tiny ~\citet{bhat2023zoedepth}} & \tiny \textbf{Recovered} & \multicolumn{1}{m{0.11\textwidth}}{\centering \tiny ~\citet{bhattad2022cut}} & \tiny  \textbf{Recovered} & \multicolumn{1}{m{0.11\textwidth}}{\centering \tiny  ~\citet{bhattad2022cut}} & \tiny  \textbf{Recovered} 
\end{tabular}\vspace{-5pt}
\caption{Our recovered scene intrinsics from StyleGAN-v2 trained on LSUN bedroom images. 
}
\label{fig:bedroom}
\vspace{-10pt}
\end{figure*}

\newcolumntype{L}{>{\centering\arraybackslash}m{0.11\textwidth}}
\begin{figure*}[t!]
\centering
\tiny
  \setlength\tabcolsep{0pt}
  \renewcommand{\arraystretch}{0}
\begin{tabular}{ccccccccc}
&\multicolumn{2}{c}{Surface Normals} & \multicolumn{2}{c}{Depth} & \multicolumn{2}{c}{Albedo} & \multicolumn{2}{c}{Shading} \\
 \includegraphics[width=0.11\textwidth]{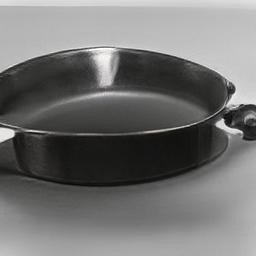} & 
\includegraphics[width=0.11\textwidth]{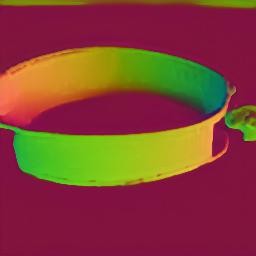} & 
\includegraphics[width=0.11\textwidth]{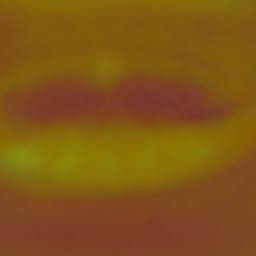} & 
\includegraphics[width=0.11\textwidth]{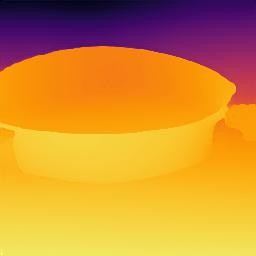} & 
\includegraphics[width=0.11\textwidth]{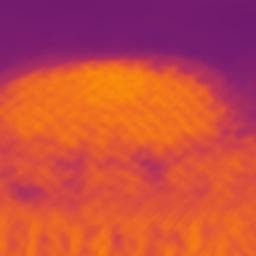} & 
\includegraphics[width=0.11\textwidth]{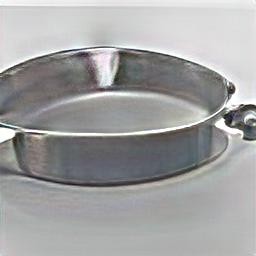} & 
\includegraphics[width=0.11\textwidth]{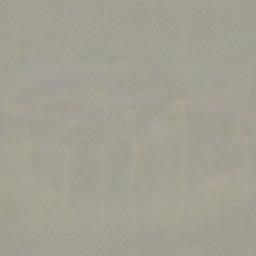} & 
\includegraphics[width=0.11\textwidth]{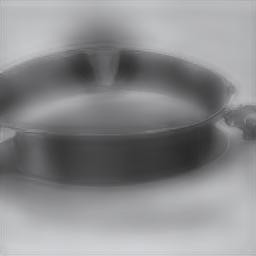} & 
\includegraphics[width=0.11\textwidth]{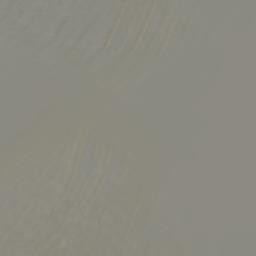} \\
 \includegraphics[width=0.11\textwidth]{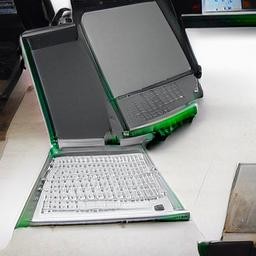} & 
\includegraphics[width=0.11\textwidth]{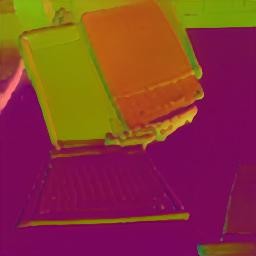} & 
\includegraphics[width=0.11\textwidth]{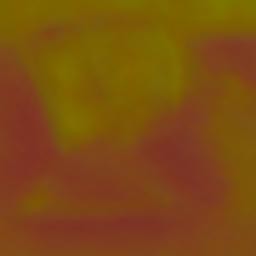} & 
\includegraphics[width=0.11\textwidth]{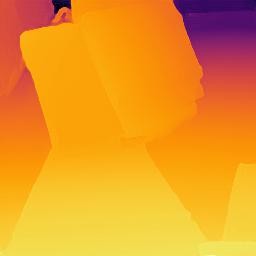} & 
\includegraphics[width=0.11\textwidth]{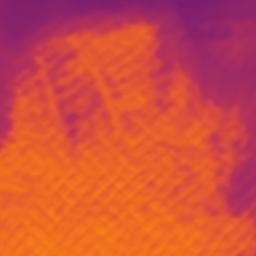} & 
\includegraphics[width=0.11\textwidth]{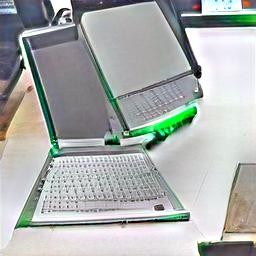} & 
\includegraphics[width=0.11\textwidth]{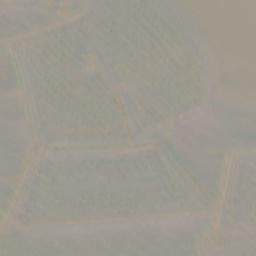} & 
\includegraphics[width=0.11\textwidth]{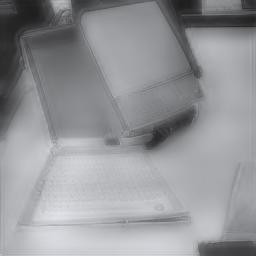} & 
\includegraphics[width=0.11\textwidth]{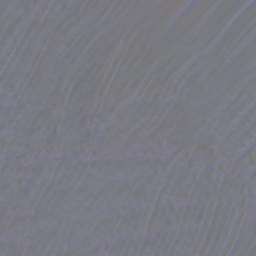} \\
\vspace{2pt}
\\
 \tiny Image & \multicolumn{1}{m{0.11\textwidth}}{\centering \tiny ~\citet{kar20223d}} & \tiny  \textbf{Recovered} & \multicolumn{1}{m{0.11\textwidth}}{\centering \tiny ~\citet{bhat2023zoedepth}} & \tiny \textbf{Recovered} & \multicolumn{1}{m{0.11\textwidth}}{\centering \tiny  ~\citet{bhattad2022cut}} & \tiny  \textbf{Recovered} & \multicolumn{1}{m{0.11\textwidth}}{\centering \tiny  ~\citet{bhattad2022cut}} & \tiny  \textbf{Recovered} 
\end{tabular}
\vspace{-8pt}
\caption{StyleGAN-XL on ImageNet. 
Recovered surface normals and depth maps, while capturing the basic shape and volume, lack precise detail and display artifacts. Albedo and Shading recoveries fail. These results are correlated with the overall bad image generation quality.}
\label{fig:imagenet}
\vspace{-5pt}
\end{figure*}

\subsection{Finding 1: Intrinsic images are encoded across generative models, and LoRA is a general approach for recovering them}\label{sec:generality}

\begin{figure*}[t!]
\centering
\tiny
  \setlength\tabcolsep{0pt}
  \renewcommand{\arraystretch}{0}
\begin{tabular}{ccccccccc}
&\multicolumn{2}{c}{Surface Normals} & \multicolumn{2}{c}{Depth} & \multicolumn{2}{c}{Albedo} & \multicolumn{2}{c}{Shading} 
\\

\includegraphics[width=0.11\textwidth]{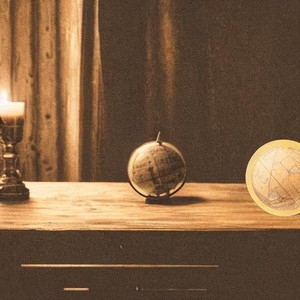} & 
\includegraphics[width=0.11\textwidth]{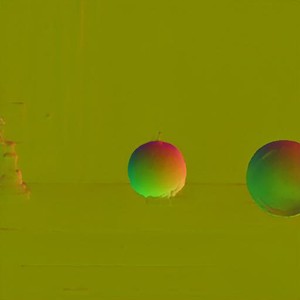} & 
\includegraphics[width=0.11\textwidth]{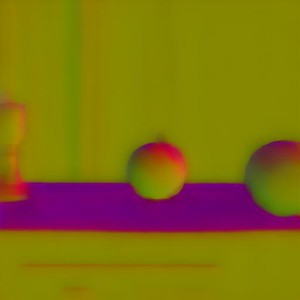} & 
\includegraphics[width=0.11\textwidth]{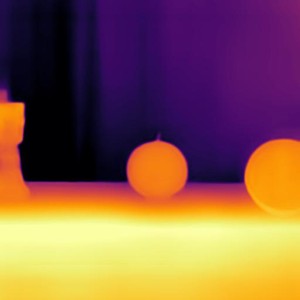} & 
\includegraphics[width=0.11\textwidth]{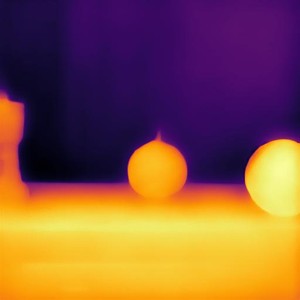} & 
\includegraphics[width=0.11\textwidth]{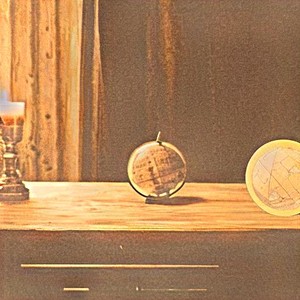} & 
\includegraphics[width=0.11\textwidth]{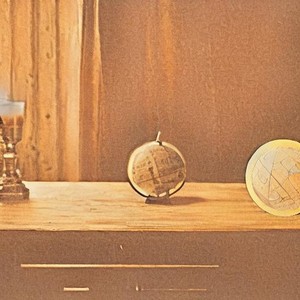} & 
\includegraphics[width=0.11\textwidth]{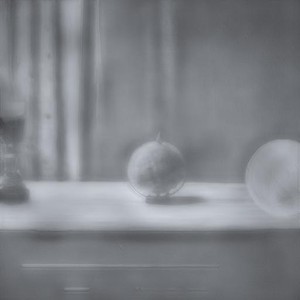} & 
\includegraphics[width=0.11\textwidth]{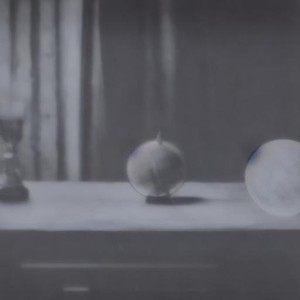} 
\\
\includegraphics[width=0.11\linewidth]{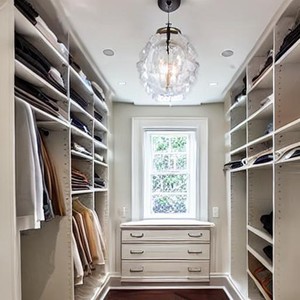} & 
\includegraphics[width=0.11\linewidth]{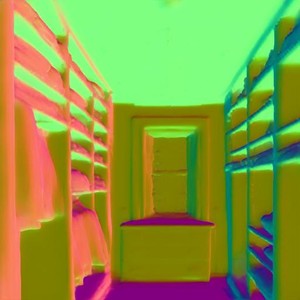} & 
\includegraphics[width=0.11\linewidth]{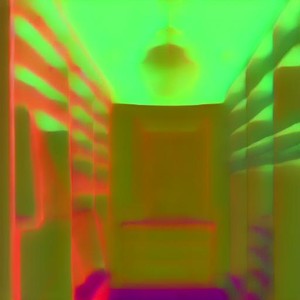} & 
\includegraphics[width=0.11\linewidth]{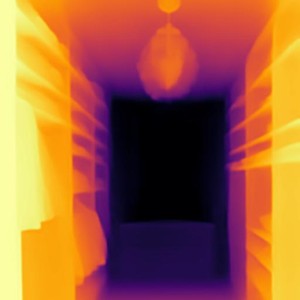} & 
\includegraphics[width=0.11\linewidth]{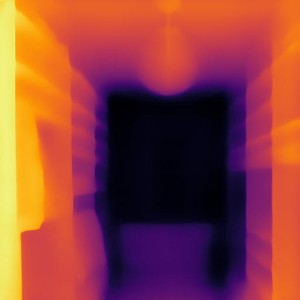} & 
\includegraphics[width=0.11\linewidth]{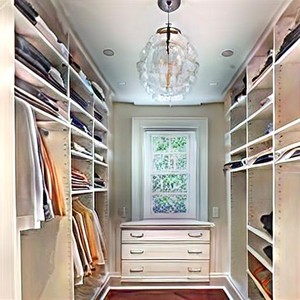} & 
\includegraphics[width=0.11\linewidth]{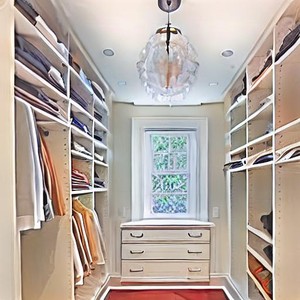} & 
\includegraphics[width=0.11\linewidth]{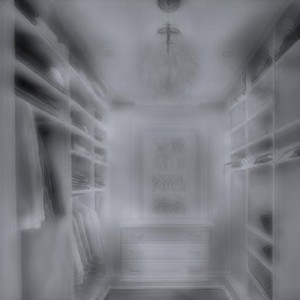} & 
\includegraphics[width=0.11\linewidth]{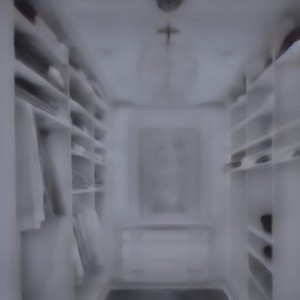} 
\\
\vspace{2pt}
\\
\tiny Image & \multicolumn{1}{m{0.11\textwidth}}{\centering \tiny ~\citet{kar20223d}} & \tiny  \textbf{Recovered} & \multicolumn{1}{m{0.11\textwidth}}{\centering \tiny ~\citet{bhat2023zoedepth}} & \tiny \textbf{Recovered} & \multicolumn{1}{m{0.11\textwidth}}{\centering \tiny ~\citet{bhattad2022cut}} & \tiny  \textbf{Recovered} & \multicolumn{1}{m{0.11\textwidth}}{\centering \tiny ~\citet{bhattad2022cut}} & \tiny  \textbf{Recovered}
\end{tabular}
\vspace{-8pt}
\caption{Scene intrinsics recovered from randomly generated stable diffusion images using LoRA. Recovered intrinsics appear to be better. For example, the table's normal in the first row is more accurate compared to~\citet{kar20223d}.
The rightmost globe also appears to be closer to the camera in recovered depth compared to~\citet{bhat2023zoedepth}. In the second row, ceiling lamp normals are visible in recovered intrinsics but not in~\citet{kar20223d}. 
These comparison highlights that the recovered intrinsics can closely align with, and sometimes surpass, these supervised monocular predictors.
}
\label{fig:sdsingle}
\vspace{-10pt}
\end{figure*}

We aim to recover intrinsic images across diverse generative models, including StyleGAN-v2~\citep{yu2019inverserendernet}, StyleGAN-XL~\citep{sauer2022stylegan}, and VQGAN~\citep{esser2020taming}, trained on datasets like FFHQ~\citep{karras2020analyzing}, LSUN Bedrooms~\citep{yu2015lsun}, and ImageNet~\citep{deng2009imagenet}. LoRA adapters are tailored to each model and dataset to recover intrinsics: surface normals, depth, albedo, and shading, demonstrating broad applicability and robustness in both qualitative assessments (\cref{fig:teaser2}, \ref{fig:generators_comparison}, \ref{fig:bedroom}, \ref{fig:sdsingle}) and quantitative (\cref{tab:model_performance_generated_images} on generated images, \cref{tab:model_performance_real} on real images). In all experiments -- covering both generated and real images -- we use pseudo-ground truth from pre-trained models as a supervisory signal for fine-tuning LoRA adapters to discover scene intrinsics within generative models as previously mentioned in \cref{sec:ilora}. We use LoRA with Rank 8 as default for all generative models if not otherwise mentioned.

We find LoRA can recover intrinsic images from almost all models tested. The notable exception is StyleGAN-XL trained on ImageNet, where it yields qualitatively poor results, which we attribute to the model's limited ability to generate realistic images (\cref{fig:imagenet}). This suggests the recovered intrinsic quality is correlated with the generative model's fidelity (see \cref{sec:control}). 
For evaluating generated images, we benchmarked against pseudo-ground truths derived from existing models, compensating for the lack of true ground truths. The performance, gauged through these comparisons, provides useful indicators but must be interpreted within the context of the selected pseudo-ground truths.

\begin{table*}[t!]
    \centering
    \caption{Quantiative analysis of scene intrinsics recovery performance by LoRA on generated images. We compare with pseudo ground truths from Omnidata-v2 for surface normals, ZoeDepth for depth, and Paradigms for albedo and shading. Metrics include mean angular error, median angular error, and L1 error for surface normals; RMS and $\delta < 1.25$ for depth; RMS for albedo and shading.
    }
    \resizebox{\linewidth}{!}{%
    \begin{tabular}{c  c  c c  c c c   c c  c  c  }
        \toprule
        Model & Pre-training Type & Domain & LoRA Param.  &  \multicolumn{3}{c}{Surface Normal} & \multicolumn{2}{c}{Depth} & \multicolumn{1}{c}{Albedo} & \multicolumn{1}{c}{Shading} \\
        \cmidrule(lr){5-7} \cmidrule(lr){8-9} \cmidrule(lr){10-10} \cmidrule(lr){11-11}
         &  &  & & Mean Error\textdegree $\downarrow$ & Median Error\textdegree $\downarrow$ & L1 Error{\tiny $\times$ 100} $\downarrow$  & RMS $\downarrow$ & $\delta < 1.25${\tiny $\times$100\%} $\uparrow$ & RMS $\downarrow$ & RMS $\downarrow$\\
        \midrule
        VQGAN & Autoregressive & FFHQ & 0.18$\%$ & 19.97 & 20.97 & 16.33  & 0.1819 & 62.33 & 0.0345 & 0.0106 \\
        StyleGAN-v2 & GAN & FFHQ & 0.57$\%$ & 16.93 & 19.60 & 13.87 & 0.1530 & 90.74 & 0.0283 & 0.0110 \\
        StyleGAN-XL & GAN & FFHQ & 0.29$\%$ & 15.28 & 18.07 & 12.63  & 0.1337 & 93.87 & 0.0287 & 0.0125 \\
        StyleGAN-v2 & GAN & LSUN Bedroom & 0.57$\%$ & 13.94 & 24.76 & 11.49 & 0.0897 & 66.88 & 0.0270 & 0.0074 \\
        StyleGAN-XL & GAN & ImageNet & 0.29$\%$ & 24.09 & 25.52 & 19.44 & 0.2175 & 38.38 & 0.1065 & 0.0119 \\      
        \hline
        \augunet (multi step) & Diffusion & Open & 0.17\% & 21.41  & 28.57 & 17.39 & 0.2042 & 41.21 & 0.0881 & 0.0099 \\
        SD-UNet (single step) & Diffusion & Open & 0.17\% & 16.63 & 23.64 & 13.69  & 0.1179 & 52.59 & 0.0487 & 0.0118 \\
        \bottomrule
    \end{tabular}
    }
    \label{tab:model_performance_generated_images}
     \vspace{-15pt}
\end{table*}

Thanks to their architecture as image-to-image translators, diffusion models are powerful image generators that easily apply to real images. Exploiting this, we use LoRA to directly retrieve intrinsic images from Stable Diffusion's UNet in a single step, bypassing the iterative reverse denoising process. The model takes a real image as input and outputs its intrinsic components, allowing for direct evaluation against actual ground truth. 
on DIODE dataset~\citep{vasiljevic2019diode}. We use the official training/evaluation split in all of our DIODE experiments. For training with fewer samples, we randomly chose samples from the official training partition. All the metrics we reported on DIODE are computed over the entire evaluation set.
In \cref{tab:model_performance_real}, we find that the LoRA adapters not only matches but, in several metrics (median error for surface normals, RMSE for depth), surpasses the performance of Omnidata and ZoeDepth -- the source of its training signal -- while using significantly less data, parameters, and training time (see Sec.\ref{sec:efficiency}).

\noindent\textbf{Extending to DINO.} 
LoRA intrinsic recovery extends beyond generative models to self-supervised, non-generative models like DINO~\citep{darcet2023vision}. 
We apply linear head and LoRA modules following ~\citet{oquab2023dinov2} to project DINO features into pixel space. Using DINOv2's `giant' model, we find quantitative results comparable to those from Stable Diffusion, with only a 0.26\% increase in parameters. But DINOv2 tends to recover intrinsics with visible discontinuities (\cref{fig:ablation_rank}d).

\begin{table*}[t!]
\vspace{-10pt}
    \centering
    \caption{Quantitative analysis of recovered scene intrinsicsacross different models on real images. 
    }
    \resizebox{\linewidth}{!}{%
    \begin{tabular}{c  c   c  c c   c c c }
        \toprule
        Model & Pre-training   & LoRA  & \multicolumn{3}{c}{Surface Normal} & \multicolumn{2}{c}{Depth} \\
        \cmidrule(lr){4-6} \cmidrule(lr){7-8}
          &Type  & Param & Mean Error\textdegree $\downarrow$ & Median Error\textdegree $\downarrow$ & L1 Error{\tiny $\times$ 100} $\downarrow$ & RMS $\downarrow$ & $\delta < 1.25${\tiny $\times$ 100} $\uparrow$  \\
       \midrule
        Omnidata-v2~\citep{kar20223d}/ZoeDepth~\citep{bhat2023zoedepth}  & Supervised  &-& {\bf 18.90} & 13.36 & {\bf 15.21}  & 0.2693 & {\bf 47.56}  \\
        \midrule
        DINOv2 & Non-Generative & 0.26\% &{19.74} & 13.72 & {16.00}  & 0.2094 & 44.32 \\
        \midrule
        \augunet (multi step) & Diffusion & 0.17\%  & 23.74 & 19.08 & 19.31  & 0.2651 & 43.19  \\
        SD-UNet (single step) & Diffusion & 0.17\%  & 20.31 & {\bf 12.54} & {16.53}  & {\bf 0.2046} & {44.90}  \\
        \bottomrule
    \end{tabular}
    }
    
    \label{tab:model_performance_real}
\vspace{-5pt}
\end{table*}

\subsection{Finding 2: Tiny new parameters \& data are enough for intrinsic recovery}\label{sec:efficiency}
\begin{figure*}[t!]
\centering
\tiny
  \setlength\tabcolsep{0pt}
  \renewcommand{\arraystretch}{0}
\begin{tabular}{ccccccccc}
\includegraphics[width = .11\textwidth]{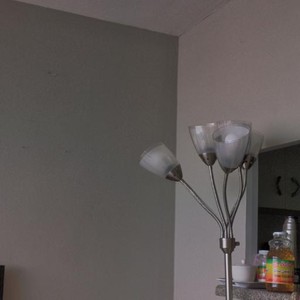}&
\includegraphics[width = .11\textwidth]{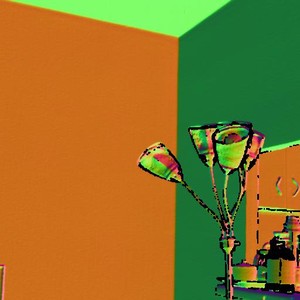}&
\includegraphics[width = .11\textwidth]{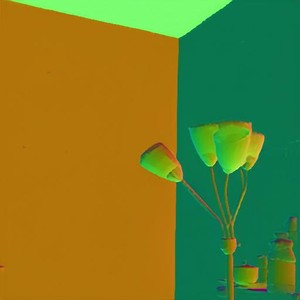}&
\includegraphics[width = .11\textwidth]{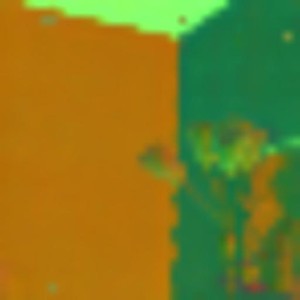}&
\includegraphics[width = .11\textwidth]{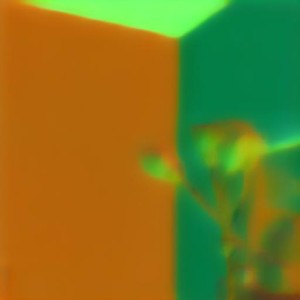}&
\includegraphics[width = .11\textwidth]{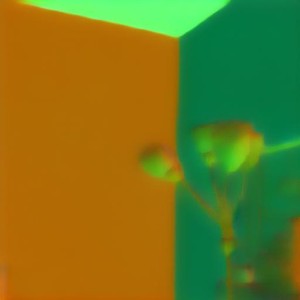}&
\includegraphics[width = .11\textwidth]{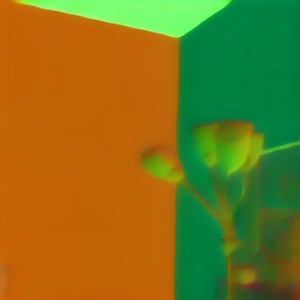}&
\includegraphics[width = .11\textwidth]{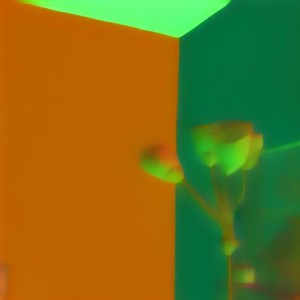}&
\includegraphics[width = .11\textwidth]{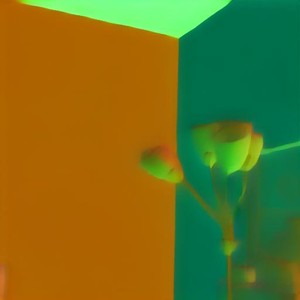}
\\
\includegraphics[width = .11\textwidth]{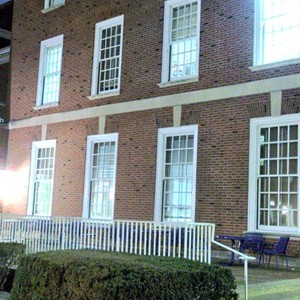}&
\includegraphics[width = .11\textwidth]{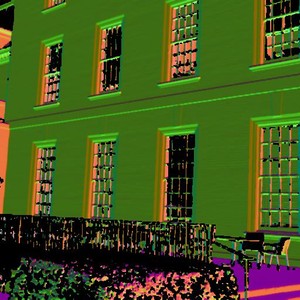}&
\includegraphics[width = .11\textwidth]{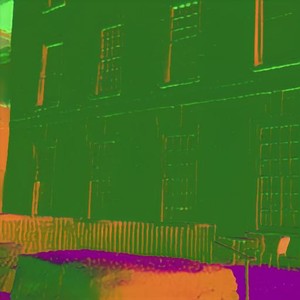}&
\includegraphics[width = .11\textwidth]{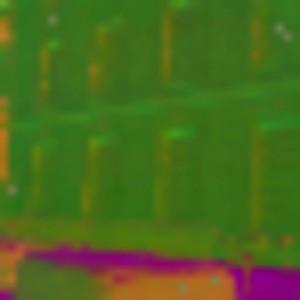}&
\includegraphics[width = .11\textwidth]{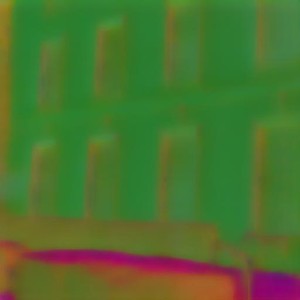}&
\includegraphics[width = .11\textwidth]{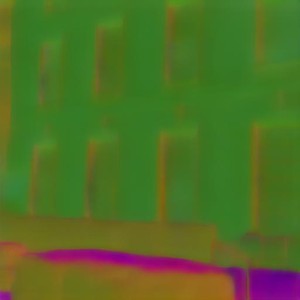}&
\includegraphics[width = .11\textwidth]{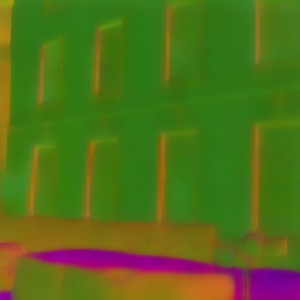}&
\includegraphics[width = .11\textwidth]{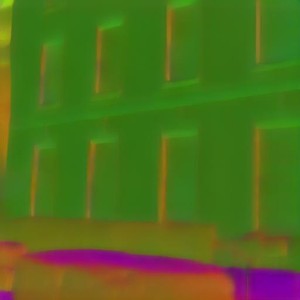}&
\includegraphics[width = .11\textwidth]{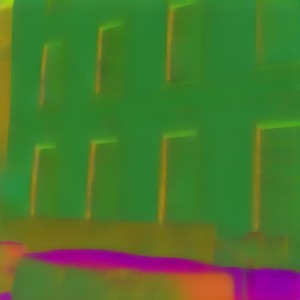}
\\
\vspace{2pt}
\\
 (a) Real &  (b) GT  &   (c)  OD-v2 &  (d)DINOv2&  (e) rank 2 &  (f) rank 4 &  (g) rank 8 &  (h) rank 16 &   (i) rank 32 \\
 \midrule
\multicolumn{2}{c}{  \begin{tabular}{c}   Mean Angular Error\textdegree $\downarrow$ \end{tabular}}&
  18.90 &   19.74&   22.28 &  22.57 &   20.31 &  21.17 &   21.84
\\
\multicolumn{2}{c}{  \begin{tabular}{c}   L1 Error ($\times$ 100 )$\downarrow$ \end{tabular}}&
   15.21 &  16.00 &   18.14 &   18.39 &   16.53 &  17.19 &   17.81 \\ 
 \multicolumn{2}{c}{  \begin{tabular}{c}   LoRA Param.  \end{tabular}}& - & 0.26\% & 0.04\% &0.08\% & 0.17\% & 0.34\% & 0.68\%
\\
\midrule
\end{tabular}
\vspace{-10pt}
\caption{Parameter Efficiency of LoRA. We evaluate various rank settings for normals recovery. Lower ranks such as 8 offer a balance between efficiency and effectiveness. All model variants are trained using SD's UNet (v1.5) with 4000 samples. Performance metrics, such as Mean Angular Error and L1 Error for normals, and additional parameter counts are detailed below each variant.
}
\vspace{-5pt}
\label{fig:ablation_rank}
\end{figure*}
\begin{figure*}[t!]
\centering
\tiny
  \setlength\tabcolsep{0pt}
  \renewcommand{\arraystretch}{0}
\begin{tabular}{cccccccc}
\includegraphics[width = .12\textwidth]{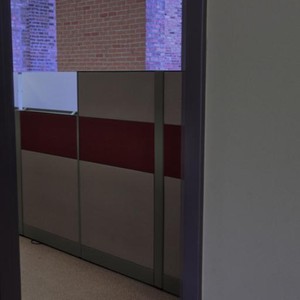}&
\includegraphics[width = .12\textwidth]{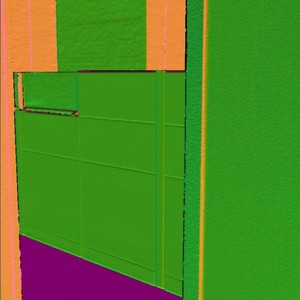}&
\includegraphics[width = .12\textwidth]{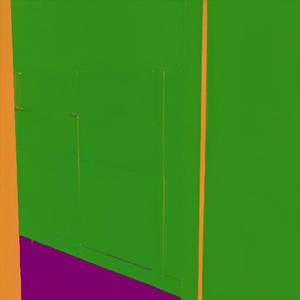}&
\includegraphics[width = .12\textwidth]{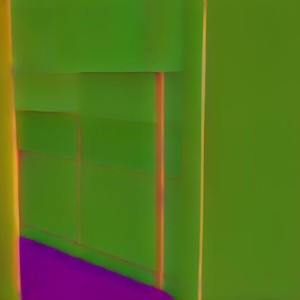}&
\includegraphics[width = .12\textwidth]{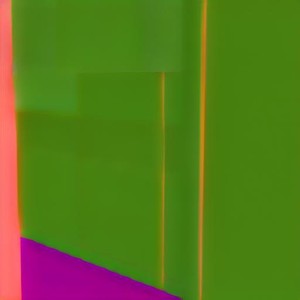}&
\includegraphics[width = .12\textwidth]{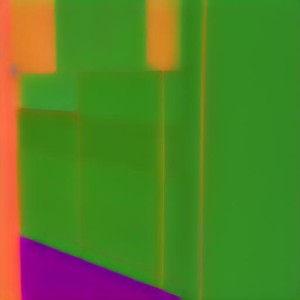}&
\includegraphics[width = .12\textwidth]{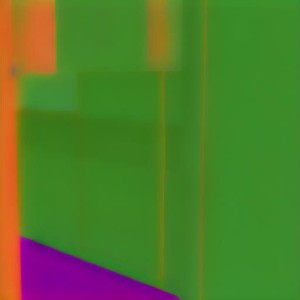}&
\includegraphics[width = .12\textwidth]{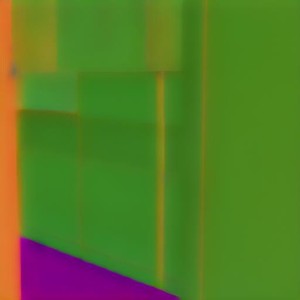}
\\
\vspace{3pt}
\\
 (a) Real &  (b) GT  &   (c)  OD-v2 &   
 (d) 250 &   (e) 1000 &  (f) 4000 &  (g) 16000 &  (h) 24895
\\
\midrule
\multicolumn{2}{c}{  \begin{tabular}{c} Mean Angular Error\textdegree $\downarrow$ \end{tabular}}&
  18.90&   
  27.73 &   22.22 &   20.31 &  21.26 &   21.64
\\
\multicolumn{2}{c}{  \begin{tabular}{c} L1 Error ($\times$ 100) $\downarrow$ \end{tabular}}&
  15.21 &   
  22.46 &   18.05 &   16.53 &   17.33 &   17.64
\\
\midrule
\end{tabular}
\vspace{-10pt}
\caption{Data efficiency. Note: SOTA supervised model (c), was trained using 12M+ labeled training samples. Even with 250 samples, LoRA captures surface normals. We observe the best performance with 4k samples. Models (d)-(h) all use the same SD UNet(v1-5) and rank 8 LoRA.}

\label{fig:ablation_num_samples}
\vspace{-10pt}
\end{figure*}
Our single-step SD-UNet model, distinguished by its high quantitative performance, serves as the basis for ablation studies that assess the influence of rank and labeled data quantity on intrinsic recovery efficiency. We verify that our requirements for compute, parameters, and data are minimal.

\noindent\textbf{Parameter efficiency.}
\cref{fig:ablation_rank} shows surface normal predictions across LoRA ranks. The highest accuracy is achieved with Rank 8, balancing accuracy and memory. Notably, a Rank 2 LoRA with only 0.4M additional parameters (a mere 0.04\% increase) still yields good performance. Note that across different models, Rank 8 adaptors adds only 0.17\% to 0.57\% additional parameters (\cref{tab:model_performance_generated_images}). \\
\noindent\textbf{Label efficiency.}
Ablations of labeled data size is included in \cref{fig:ablation_num_samples}. Peak performance is reached by using a modest 4000 training examples, with credible predictions visible from as few as 250 samples. 

\begin{figure*}[t!]
\centering
\setlength{\tabcolsep}{0pt}
\renewcommand{\arraystretch}{0}

\scriptsize
\begin{tabular}{cccccc}
\includegraphics[width = 0.11\textwidth]{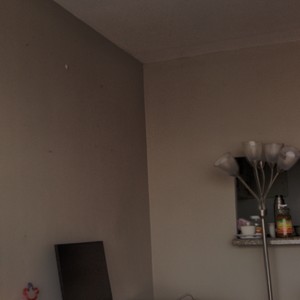}&
\includegraphics[width = 0.11\textwidth]{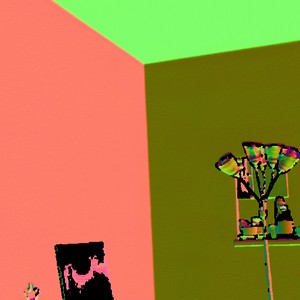}&
\includegraphics[width = 0.11\textwidth]{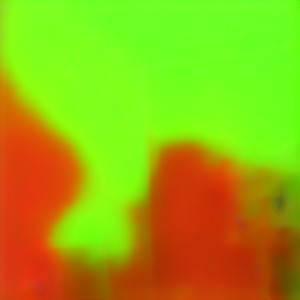}&
\includegraphics[width = 0.11\textwidth]{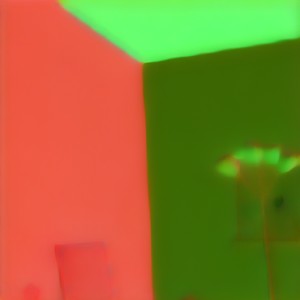}&
\includegraphics[width = 0.11\textwidth]{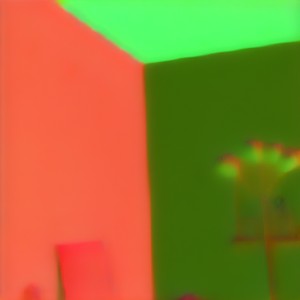}&
\includegraphics[width = 0.11\textwidth]{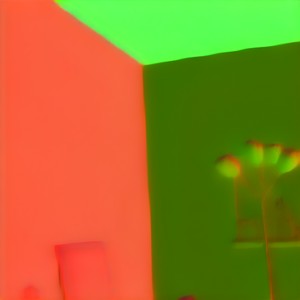}
\\
\includegraphics[width = 0.11\textwidth]{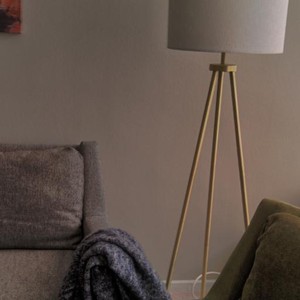}&
\includegraphics[width = 0.11\textwidth]{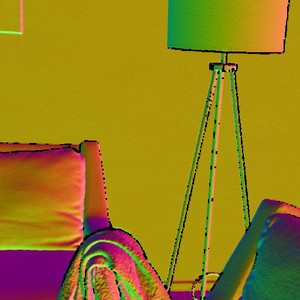}&
\includegraphics[width = 0.11\textwidth]{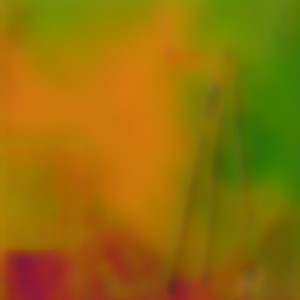}&
\includegraphics[width = 0.11\textwidth]{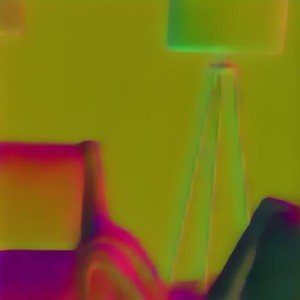}&
\includegraphics[width = 0.11\textwidth]{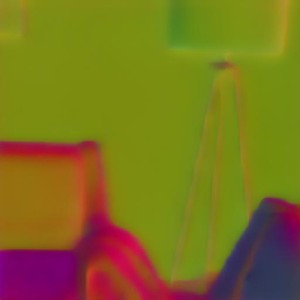}&
\includegraphics[width = 0.11\textwidth]{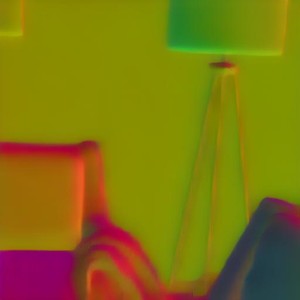}
\\
\vspace{2pt}\\
{ Real} & { GT} & {Random init.} & { SD-UNet v1-1} & { SD-UNet v1-2} & { SD-UNet v1-5}
\\
\midrule
\multicolumn{2}{c}{  \begin{tabular}{c} Mean Angular Error\textdegree $\downarrow$ \end{tabular}}& {36.18}&{ 21.84} & { 21.41} & { 20.31} 
\\
\multicolumn{2}{c}{  \begin{tabular}{c} L1 Error ($\times$ 100) $\downarrow$ \end{tabular}}& {29.28}&{ 17.78} 
 & { 17.38} & { 16.53}
\\
\midrule
\end{tabular}
\vspace{-10pt}
\caption{Better generators encode better intrinsics. We compare different versions of Stable Diffusion (v1-1, v1-2, v1-5). The progress from SD v1-1 to SD v1-5 shows improvements in recovered intrinsics paralleling improvements in image generation. Control experiments with a randomly initialized UNet fail to retrieve surface normals, emphasizing the reliance on learned priors from generative training for effective intrinsic representation recovery.
}
\label{fig:ablation_model_versions}
\vspace{-10pt}
\end{figure*}

\subsection{Finding 3: Better the generator better is intrinsic image recovery}\label{sec:control}

To assess if our method leverages pre-trained generative capabilities or primarily depends on LoRA layers, we performed a control experiment using a randomly initialized SD UNet, following the same training protocol of our SD-UNet model. The poor results from this model (see \cref{fig:ablation_model_versions}) corroborate that the learned features developed during generative pre-training are crucial for intrinsic retrieval, rather than the LoRA layers alone. Furthermore, analyzing different Stable Diffusion versions (v1-1, v1-2 and v1-5) under the same training protocol reveals that enhancements in image generation quality correlate positively with intrinsic recovery capabilities. This assertion is further reinforced by observing a correlation between lower FID scores (9.6 for VQGAN~\citep{esser2020taming}, 3.62 for StyleGAN-v2~\citep{Karras2020ada} and  2.19 for StyleGAN-XL~\citep{sauer2022stylegan}) and improved intrinsic predictions in our FFHQ experiments (\cref{fig:generators_comparison} and \cref{tab:model_performance_generated_images}: first three rows), confirming that superior generative models yield more accurate intrinsics.

\begin{wrapfigure}{r}{0.48\textwidth}
\vspace{-12pt}
  \centering
\scriptsize
  \setlength\tabcolsep{0.pt}
  \renewcommand{\arraystretch}{0.}

  \begin{tabular}{ccccc}

    \includegraphics[width=0.095\textwidth]{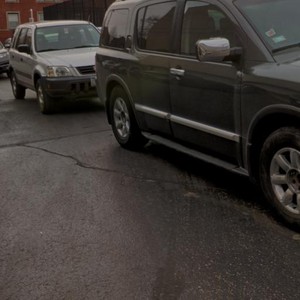}&
    \includegraphics[width=0.095\textwidth]{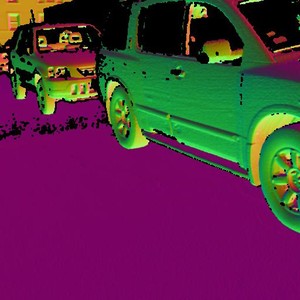}&
    \includegraphics[width=0.095\textwidth]{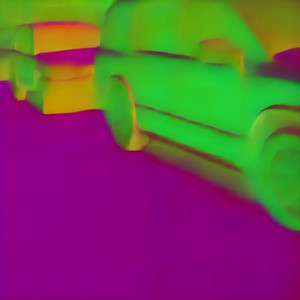}&
    \includegraphics[width=0.095\textwidth]{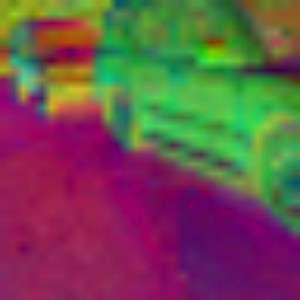}&
    \includegraphics[width=0.095\textwidth]{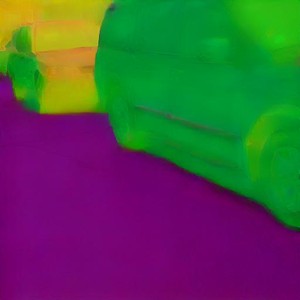}
    \\
    \includegraphics[width=0.095\textwidth]{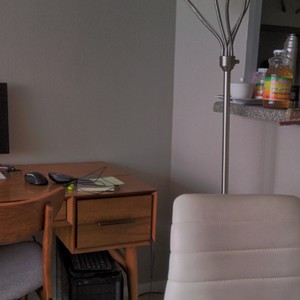}&
    \includegraphics[width=0.095\textwidth]{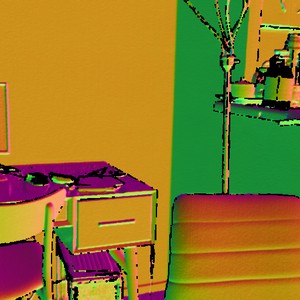}&
    \includegraphics[width=0.095\textwidth]{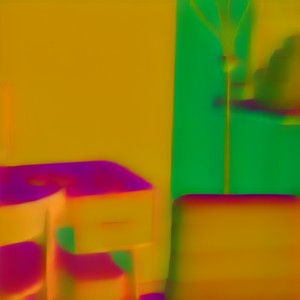}&
    \includegraphics[width=0.095\textwidth]{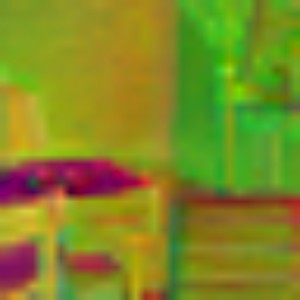}&
    \includegraphics[width=0.095\textwidth]{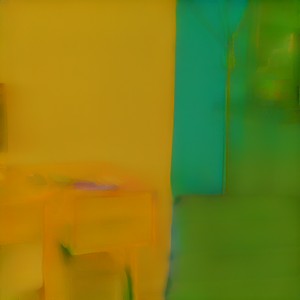}
    \\
        \vspace{2pt}
    \\
    Image & GT & \textbf{LoRA} & Linear Probe & Fine-tuning 
    \end{tabular}%
    \vspace{-5pt}
\caption{%
LoRA recovers better intrinsics. Here all approaches use 250 labeled data. 
}%
\label{fig:baseline_images}
\vspace{-2pt}
\end{wrapfigure}

\subsection{Finding 4: LoRA recovers better intrinsic images than other approaches}\label{sec:baselines}

We compare LoRA with two common approaches: linear probing and full model fine-tuning. Following 
\citet{chen2023beyond} for linear probing and using standard fine-tuning practices, we train all methods with a small dataset of 250 samples to 16000 samples. All three are trained with the same number of epochs and have converged at the end of the training. Our findings in \cref{tab:baseline_comparison} and \cref{fig:baseline_images} show that LoRA significantly outperforms the other two in low-data regimes, validating its preferable efficacy and data efficiency.

\begin{table}[htpb!]
\vspace{-10pt}
    \centering    \caption{We find LoRA to consistently outperform baselines for different training samples.}
    \resizebox{\linewidth}{!}{%
    \begin{tabular}{c c c c c c c c c c c}
        \toprule
         &Steps/s&Peak Train GPU Mem\%  &\multicolumn{2}{c}{250} & \multicolumn{2}{c}{1000}  &  \multicolumn{2}{c}{4000} & \multicolumn{2}{c}{16000} \\
        \cmidrule(lr){4-5} \cmidrule(lr){6-7} \cmidrule(lr){8-9} \cmidrule(lr){10-11}
         &&& Mean Error\textdegree $\downarrow$ & L1 {\tiny $\times$ 100} $\downarrow$ & Mean Error\textdegree $\downarrow$ & L1 {\tiny $\times$ 100} $\downarrow$ &Mean Error\textdegree $\downarrow$ & L1 {\tiny $\times$ 100} $\downarrow$ &Mean Error\textdegree $\downarrow$ & L1 {\tiny $\times$ 100} $\downarrow$ \\
        \midrule 

        Linear Probe &2.13&29.46\% & 29.10 & 23.74 & 28.45 & 23.25 & 28.52 & 23.26 & 28.22 & 23.11 \\
        Fine-tuning &0.77 & 86.78\% & 34.40 & 27.58 & 25.19 & 20.28 & 28.03 & 22.17 & 27.39 & 22.24\\
        \midrule
        LoRA & 0.94 & 63.48\% & {\bf 27.73} & {\bf 22.46} & {\bf 22.22} & {\bf 18.05} & {\bf 20.31} & {\bf 16.53} & {\bf 21.26} & {\bf 17.33} \\
        \bottomrule
    \end{tabular}
    }
    \label{tab:baseline_comparison}
\vspace{-10pt}
\end{table}

\section{Towards Improved Intrinsic Images Recovery}
\label{sec:p2i}
\begin{figure}[t!]
\centering
\setlength{\tabcolsep}{0pt}
\renewcommand{\arraystretch}{0}
\begin{tabular}{cccccc}
\scriptsize
\includegraphics[width=0.13\textwidth]{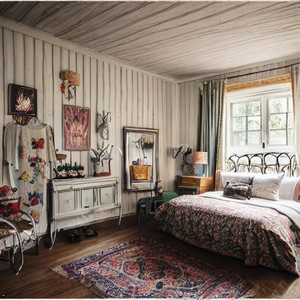}& 
\includegraphics[width=0.13\textwidth]{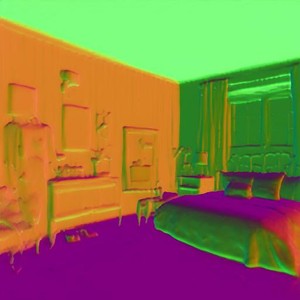}& 
\includegraphics[width=0.13\textwidth]{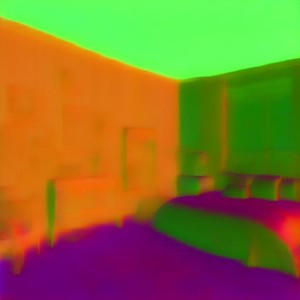}& 
\includegraphics[width=0.13\textwidth]{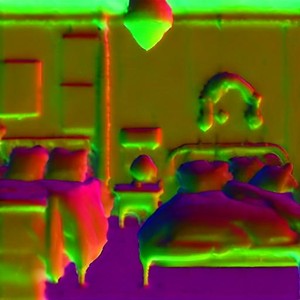}& 
\includegraphics[width=0.13\textwidth]{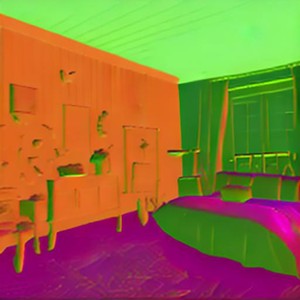}&
\includegraphics[width=0.13\textwidth]{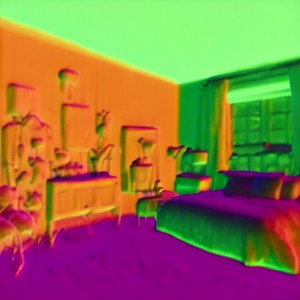}
\\
\includegraphics[width=0.13\textwidth]{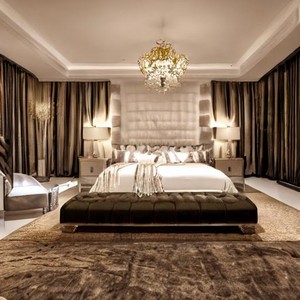}& 
\includegraphics[width=0.13\textwidth]{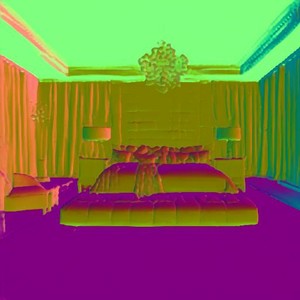}& 
\includegraphics[width=0.13\textwidth]{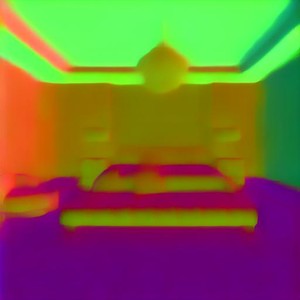}& 
\includegraphics[width=0.13\textwidth]{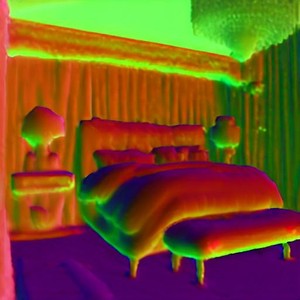}& 
\includegraphics[width=0.13\textwidth]{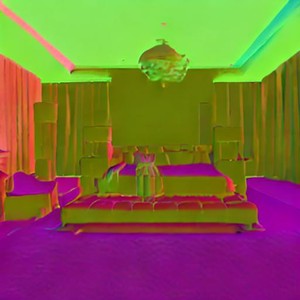}&
\includegraphics[width=0.13\textwidth]{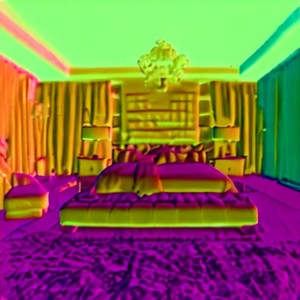}
\\
\vspace{2pt}
\\
\scriptsize
Image & \multicolumn{1}{m{0.13\textwidth}}{\scriptsize
\centering ~\citet{kar20223d}} & \multicolumn{1}{m{0.13\textwidth}}{\scriptsize
\centering SDv1-5 (single)} & \multicolumn{1}{m{0.13\textwidth}}{\scriptsize
\centering SDv1-5 (multi)}  & \multicolumn{1}{m{0.13\textwidth}}{\scriptsize
\centering \augunet (multi)} & \multicolumn{1}{m{0.13\textwidth}}{\scriptsize
\centering \augunet (multi) w/ curated data} 
\end{tabular}
\vspace{-5pt}
\caption{Naive multi-step diffusion leads to wrong intrinsics (fourth column). Our augmentation (\augunet), the fifth column, recovers with the correct layout. The last column further demonstrates highly detailed intrinsic recovery by training LoRA exclusively on domain-specific bedroom images.}
\label{fig:bedroom_img_prior}
\vspace{-2pt}
\end{figure}

\vspace{-4pt}
In the previous section, we showed that SD-UNet captures various intrinsic images like normals, depth, albedo, and shading, as evidenced by our evaluation. A natural question arises: can we improve these intrinsics using \emph{multi-step} diffusion inference? 
While multi-step diffusion improves sharpness, we find two challenges: (a) intrinsics misaligned with input, and (b) shift in the distribution of outputs relative to the ground truth (visually manifesting as a color shift) (see Fig.~\ref{fig:bedroom_img_prior}).

\begin{figure*}[t!]
\centering
\scriptsize
  \setlength\tabcolsep{0pt}
  \renewcommand{\arraystretch}{0}

  \begin{tabular}{ccccccc}

        \includegraphics[width=0.12\textwidth]{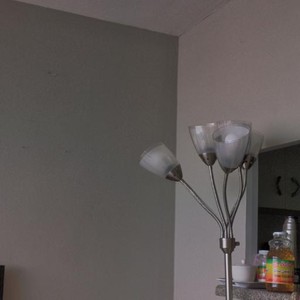}
        &
    \includegraphics[width=0.12\textwidth]{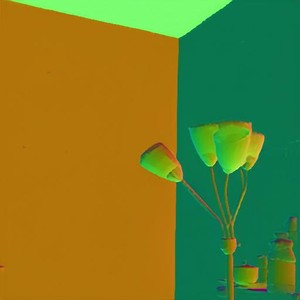} &
        \includegraphics[width=0.12\textwidth]{images/dinov2/00020_00185_indoors_020_020_normal.jpg} 
        &
        \includegraphics[width = 0.12\textwidth]{images/ablations/rank/ablation_rank_8_1.jpg}&
        \includegraphics[width=0.12\textwidth]{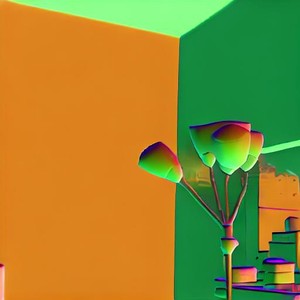} &
        \includegraphics[width=0.12\textwidth]{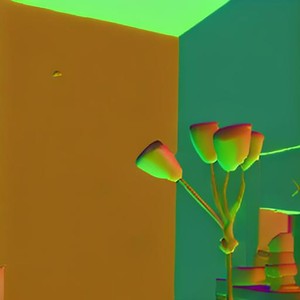} 
        &
        \includegraphics[width=0.12\textwidth]{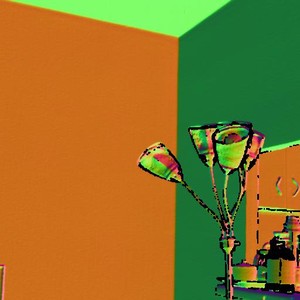}
\\
    &
    \includegraphics[width=0.12\textwidth]{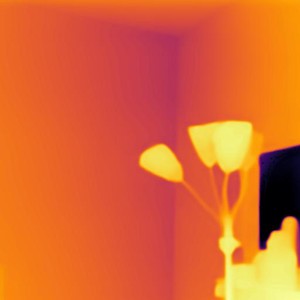}
    &
    \includegraphics[width=0.12\textwidth]{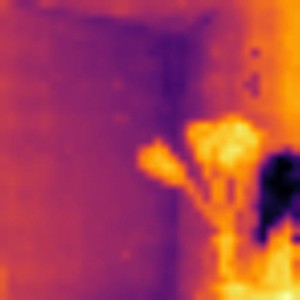}
    &
    \includegraphics[width=0.12\textwidth]{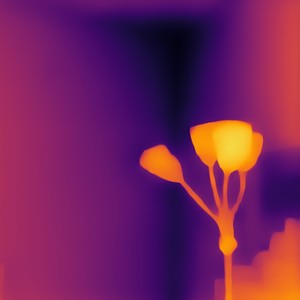}
    &
    \includegraphics[width=0.12\textwidth]{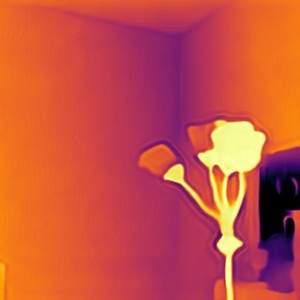}
    &
    \includegraphics[width=0.12\textwidth]{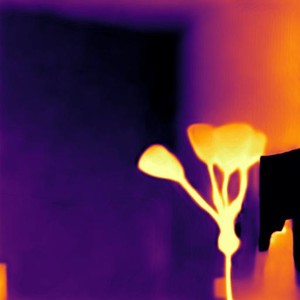}
    &
    \includegraphics[width=0.12\textwidth]{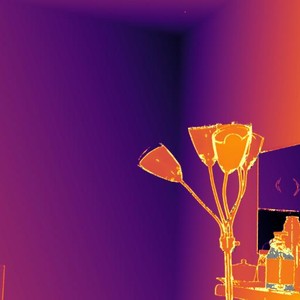}
\\

     \includegraphics[width=0.12\textwidth]{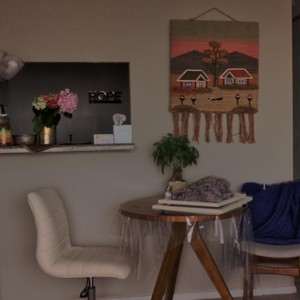} &
    \includegraphics[width=0.12\textwidth]{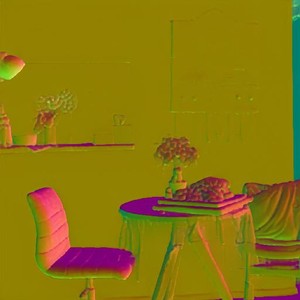} 
    &
    \includegraphics[width=0.12\textwidth]{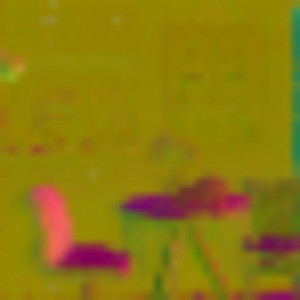} 
    &
    \includegraphics[width=0.12\textwidth]{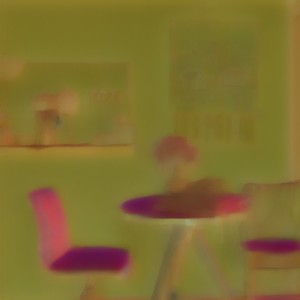} 
    &
    \includegraphics[width=0.12\textwidth]{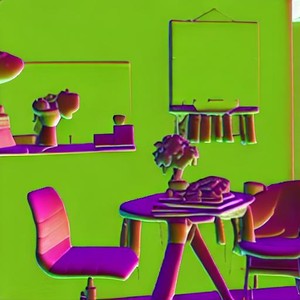} 
    &
    \includegraphics[width=0.12\textwidth]{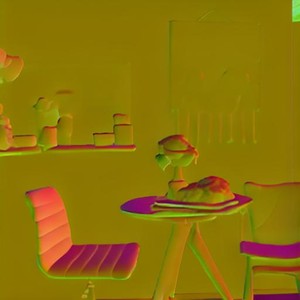} 
    &
        \includegraphics[width=0.12\textwidth]{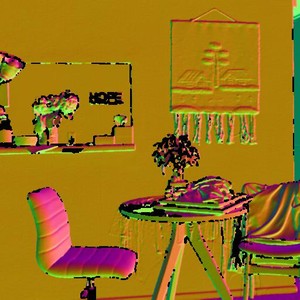} 
\\

& 

\includegraphics[width=0.12\textwidth]{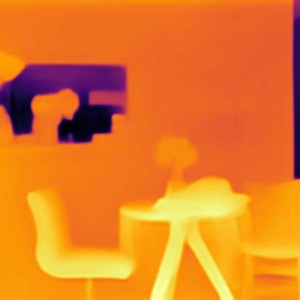}
&
\includegraphics[width=0.12\textwidth]{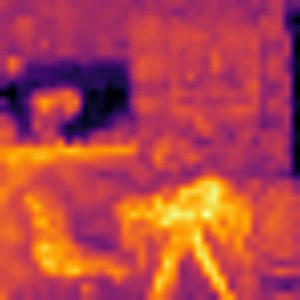}
&
\includegraphics[width=0.12\textwidth]{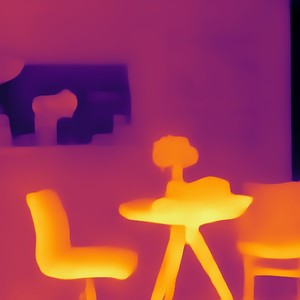}
&
\includegraphics[width=0.12\textwidth]{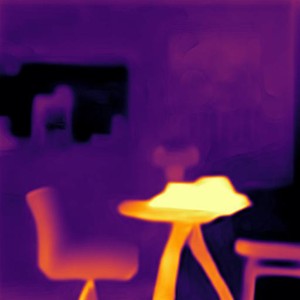}
&
 \includegraphics[width=0.12\textwidth]            {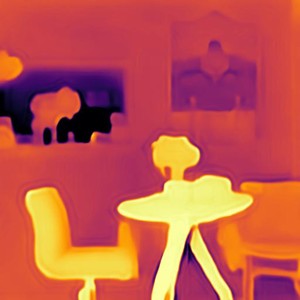}
&
\includegraphics[width=0.12\textwidth]{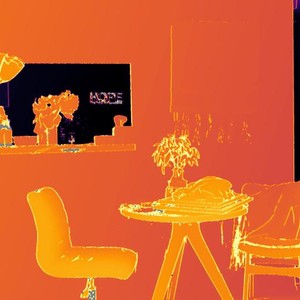}
\\
\vspace{2pt} 
\\
  Image & Pseudo GT & DINOv2 & SD-UNet & \augold & \augunet & GT
  \\
\end{tabular}
\vspace{-5pt}
\caption{ We show normals (top in each set) and depth (bottom in each set) derived from improved multi-step diffusion process from \augunet. \augold is similar to \augunet except it uses SDv1-5 and does not use Zero SNR strategy. 
\augold presents sharper details, especially in complex areas (lamp stand and chair). \augunetnospace, on the other hand, have a significant improvement in reducing color shifts while maintaining sharpness, as seen in the comparison with ground truth in the last column.
}
\vspace{-10pt}
\label{fig:image_conditioning}
\end{figure*}

To address (a), we augment the noise input to the UNet with the input image's latent encoding, as in InstructPix2Pix~\citep{brooks2023instructpix2pix}. These new parameters are frozen. 
(b) is a known artifact attributed to SD's difficulty generating images that are not with medium brightness~\citep{deck2023easing, lin2023common}. %
\citet{lin2023common} propose a Zero SNR strategy that improves color consistency but requires SD trained with a v-prediction objective, absent in SDv1-5. 
However, SD v2-1 employs a v-prediction objective. Therefore we replace SDv1-5 with SDv2-1 while maintaining our previously described learning protocol.
We name this multi-step augmented SDv2-1 model \augunetnospace. \augunet solves the misalignment issue and reduces the color shift significantly (Fig.~\ref{fig:image_conditioning}), resulting in the generation of high-quality, sharp scene intrinsics with improved quantitative accuracy. However, quantitatively, the results still fall short of our single-step SD-UNet result. 

\section{Discussions and Limitations%
}
We find consistent evidence that generative models implicitly learn intrinsic images, allowing tiny LoRA adapters to recover them with minimal fine-tuning on small labeled data. More powerful generative models produce more accurate intrinsic images, strengthening our hypothesis that learning this information is a natural byproduct of learning to generate images well. %

\noindent\textbf{Limitations}. Although we show that generative models carry a wealth of intrinsic information, it is still ambiguous how these models use this information when generating images. Secondly, even though our framework is both parameter and label-efficient, we believe there is still room for further reduction and perhaps the development of a parameter-free approach. Lastly, the \augunet generates sharper results but still lags behind its single-step counterpart in terms of quantitative analysis. Further work is needed to explore this question.

\clearpage
\bibliography{iclr2025_conference}

\begin{thebibliography}{71}
\providecommand{\natexlab}[1]{#1}
\providecommand{\url}[1]{\texttt{#1}}
\expandafter\ifx\csname urlstyle\endcsname\relax
  \providecommand{\doi}[1]{doi: #1}\else
  \providecommand{\doi}{doi: \begingroup \urlstyle{rm}\Url}\fi

\bibitem[Abdal et~al.(2021)Abdal, Zhu, Mitra, and Wonka]{abdal2021labels4free}
Rameen Abdal, Peihao Zhu, Niloy~J Mitra, and Peter Wonka.
\newblock Labels4free: Unsupervised segmentation using stylegan.
\newblock In \emph{Proceedings of the IEEE/CVF International Conference on Computer Vision}, 2021.

\bibitem[Bao et~al.(2022)Bao, Hebert, and Wang]{bao2022generative}
Zhipeng Bao, Martial Hebert, and Yu-Xiong Wang.
\newblock Generative modeling for multi-task visual learning.
\newblock In \emph{International Conference on Machine Learning}. PMLR, 2022.

\bibitem[Barrow \& Tenenbaum(1978)Barrow and Tenenbaum]{barrow1978recovering}
H~Barrow and J~Tenenbaum.
\newblock Recovering intrinsic scene characteristics.
\newblock \emph{Comput. vis. syst}, 1978.

\bibitem[Bau et~al.(2020)Bau, Zhu, Strobelt, Lapedriza, Zhou, and Torralba]{bau2018gan}
David Bau, Jun-Yan Zhu, Hendrik Strobelt, Agata Lapedriza, Bolei Zhou, and Antonio Torralba.
\newblock Understanding the role of individual units in a deep neural network.
\newblock \emph{Proceedings of the National Academy of Sciences}, 2020.

\bibitem[Bhat et~al.(2023)Bhat, Birkl, Wofk, Wonka, and M{\"u}ller]{bhat2023zoedepth}
Shariq~Farooq Bhat, Reiner Birkl, Diana Wofk, Peter Wonka, and Matthias M{\"u}ller.
\newblock Zoedepth: Zero-shot transfer by combining relative and metric depth.
\newblock \emph{arXiv preprint arXiv:2302.12288}, 2023.

\bibitem[Bhattad \& Forsyth(2022)Bhattad and Forsyth]{bhattad2022cut}
Anand Bhattad and David~A Forsyth.
\newblock Cut-and-paste object insertion by enabling deep image prior for reshading.
\newblock In \emph{2022 International Conference on 3D Vision (3DV)}. IEEE, 2022.

\bibitem[Bhattad et~al.(2023{\natexlab{a}})Bhattad, McKee, Hoiem, and Forsyth]{bhattad2023stylegan}
Anand Bhattad, Daniel McKee, Derek Hoiem, and DA~Forsyth.
\newblock Stylegan knows normal, depth, albedo, and more.
\newblock In \emph{Advances in Neural Information Processing Systems (NeurIPS)}, 2023{\natexlab{a}}.

\bibitem[Bhattad et~al.(2023{\natexlab{b}})Bhattad, Shah, Hoiem, and Forsyth]{bhattad2023make}
Anand Bhattad, Viraj Shah, Derek Hoiem, and DA~Forsyth.
\newblock Make it so: Steering stylegan for any image inversion and editing.
\newblock \emph{arXiv preprint arXiv:2304.14403}, 2023{\natexlab{b}}.

\bibitem[Bhattad et~al.(2024)Bhattad, Soole, and Forsyth]{bhattad2023StylitGAN}
Anand Bhattad, James Soole, and DA~Forsyth.
\newblock Stylitgan: Image-based relighting via latent control.
\newblock In \emph{Proceedings of the IEEE/CVF Conference on Computer Vision and Pattern Recognition}, pp.\  4231--4240, 2024.

\bibitem[Brooks et~al.(2023)Brooks, Holynski, and Efros]{brooks2023instructpix2pix}
Tim Brooks, Aleksander Holynski, and Alexei~A Efros.
\newblock Instructpix2pix: Learning to follow image editing instructions.
\newblock In \emph{Proceedings of the IEEE/CVF Conference on Computer Vision and Pattern Recognition}, 2023.

\bibitem[Chen et~al.(2023)Chen, Vi{\'e}gas, and Wattenberg]{chen2023beyond}
Yida Chen, Fernanda Vi{\'e}gas, and Martin Wattenberg.
\newblock Beyond surface statistics: Scene representations in a latent diffusion model.
\newblock \emph{arXiv preprint arXiv:2306.05720}, 2023.

\bibitem[Darcet et~al.(2023)Darcet, Oquab, Mairal, and Bojanowski]{darcet2023vision}
Timoth{\'e}e Darcet, Maxime Oquab, Julien Mairal, and Piotr Bojanowski.
\newblock Vision transformers need registers.
\newblock \emph{arXiv preprint arXiv:2309.16588}, 2023.

\bibitem[Deck \& Bischoff(2023)Deck and Bischoff]{deck2023easing}
Katherine Deck and Tobias Bischoff.
\newblock Easing color shifts in score-based diffusion models.
\newblock \emph{arXiv preprint arXiv:2306.15832}, 2023.

\bibitem[Deng et~al.(2009)Deng, Dong, Socher, Li, Li, and Fei-Fei]{deng2009imagenet}
Jia Deng, Wei Dong, Richard Socher, Li-Jia Li, Kai Li, and Li~Fei-Fei.
\newblock Imagenet: A large-scale hierarchical image database.
\newblock In \emph{2009 IEEE conference on computer vision and pattern recognition}. Ieee, 2009.

\bibitem[Eftekhar et~al.(2021)Eftekhar, Sax, Malik, and Zamir]{eftekhar2021omnidata}
Ainaz Eftekhar, Alexander Sax, Jitendra Malik, and Amir Zamir.
\newblock Omnidata: A scalable pipeline for making multi-task mid-level vision datasets from 3d scans.
\newblock In \emph{Proceedings of the IEEE/CVF International Conference on Computer Vision}, 2021.

\bibitem[Eigen et~al.(2014)Eigen, Puhrsch, and Fergus]{eigen2014depth}
David Eigen, Christian Puhrsch, and Rob Fergus.
\newblock Depth map prediction from a single image using a multi-scale deep network.
\newblock \emph{Advances in neural information processing systems}, 2014.

\bibitem[Esser et~al.(2020)Esser, Rombach, and Ommer]{esser2020taming}
Patrick Esser, Robin Rombach, and Björn Ommer.
\newblock Taming transformers for high-resolution image synthesis, 2020.

\bibitem[Forsyth \& Rock(2021)Forsyth and Rock]{forsyth2021intrinsic}
David Forsyth and Jason~J Rock.
\newblock Intrinsic image decomposition using paradigms.
\newblock \emph{IEEE transactions on pattern analysis and machine intelligence}, 2021.

\bibitem[Gal et~al.(2022)Gal, Alaluf, Atzmon, Patashnik, Bermano, Chechik, and Cohen-or]{gal2022image}
Rinon Gal, Yuval Alaluf, Yuval Atzmon, Or~Patashnik, Amit~Haim Bermano, Gal Chechik, and Daniel Cohen-or.
\newblock An image is worth one word: Personalizing text-to-image generation using textual inversion.
\newblock In \emph{The Eleventh International Conference on Learning Representations}, 2022.

\bibitem[Gan et~al.(2023)Gan, Park, Schubert, Philippakis, and Alaa]{gan2023instructcv}
Yulu Gan, Sungwoo Park, Alexander Schubert, Anthony Philippakis, and Ahmed Alaa.
\newblock Instructcv: Instruction-tuned text-to-image diffusion models as vision generalists.
\newblock \emph{arXiv preprint arXiv:2310.00390}, 2023.

\bibitem[Goodfellow et~al.(2014)Goodfellow, Pouget-Abadie, Mirza, Xu, Warde-Farley, Ozair, Courville, and Bengio]{goodfellow2014generative}
Ian Goodfellow, Jean Pouget-Abadie, Mehdi Mirza, Bing Xu, David Warde-Farley, Sherjil Ozair, Aaron Courville, and Yoshua Bengio.
\newblock Generative adversarial nets.
\newblock \emph{Advances in neural information processing systems}, 27, 2014.

\bibitem[Gutmann \& Hyv{\"a}rinen(2010)Gutmann and Hyv{\"a}rinen]{gutmann2010noise}
Michael Gutmann and Aapo Hyv{\"a}rinen.
\newblock Noise-contrastive estimation: A new estimation principle for unnormalized statistical models.
\newblock In \emph{Proceedings of the thirteenth international conference on artificial intelligence and statistics}. JMLR Workshop and Conference Proceedings, 2010.

\bibitem[Hedlin et~al.(2023)Hedlin, Sharma, Mahajan, Isack, Kar, Tagliasacchi, and Yi]{hedlin2023unsupervised}
Eric Hedlin, Gopal Sharma, Shweta Mahajan, Hossam Isack, Abhishek Kar, Andrea Tagliasacchi, and Kwang~Moo Yi.
\newblock Unsupervised semantic correspondence using stable diffusion.
\newblock \emph{arXiv preprint arXiv:2305.15581}, 2023.

\bibitem[Ho et~al.(2020)Ho, Jain, and Abbeel]{ho2020denoising}
Jonathan Ho, Ajay Jain, and Pieter Abbeel.
\newblock Denoising diffusion probabilistic models.
\newblock \emph{Advances in neural information processing systems}, 2020.

\bibitem[Hu et~al.(2022)Hu, Shen, Wallis, Allen-Zhu, Li, Wang, Wang, and Chen]{hu2021lora}
Edward~J Hu, Yelong Shen, Phillip Wallis, Zeyuan Allen-Zhu, Yuanzhi Li, Shean Wang, Lu~Wang, and Weizhu Chen.
\newblock Lo{RA}: Low-rank adaptation of large language models.
\newblock In \emph{International Conference on Learning Representations}, 2022.

\bibitem[Jahanian et~al.(2021)Jahanian, Puig, Tian, and Isola]{jahanian2021generative}
Ali Jahanian, Xavier Puig, Yonglong Tian, and Phillip Isola.
\newblock Generative models as a data source for multiview representation learning.
\newblock \emph{arXiv preprint arXiv:2106.05258}, 2021.

\bibitem[Kang et~al.(2023)Kang, Zhu, Zhang, Park, Shechtman, Paris, and Park]{kang2023gigagan}
Minguk Kang, Jun-Yan Zhu, Richard Zhang, Jaesik Park, Eli Shechtman, Sylvain Paris, and Taesung Park.
\newblock Scaling up gans for text-to-image synthesis.
\newblock In \emph{Proceedings of the IEEE Conference on Computer Vision and Pattern Recognition (CVPR)}, 2023.

\bibitem[Kar et~al.(2022)Kar, Yeo, Atanov, and Zamir]{kar20223d}
O{\u{g}}uzhan~Fatih Kar, Teresa Yeo, Andrei Atanov, and Amir Zamir.
\newblock 3d common corruptions and data augmentation.
\newblock In \emph{Proceedings of the IEEE/CVF Conference on Computer Vision and Pattern Recognition}, 2022.

\bibitem[Karras et~al.(2019)Karras, Laine, and Aila]{karras2019style}
Tero Karras, Samuli Laine, and Timo Aila.
\newblock A style-based generator architecture for generative adversarial networks.
\newblock In \emph{Proceedings of the IEEE/CVF Conference on Computer Vision and Pattern Recognition}, 2019.

\bibitem[Karras et~al.(2020{\natexlab{a}})Karras, Aittala, Hellsten, Laine, Lehtinen, and Aila]{Karras2020ada}
Tero Karras, Miika Aittala, Janne Hellsten, Samuli Laine, Jaakko Lehtinen, and Timo Aila.
\newblock Training generative adversarial networks with limited data.
\newblock In \emph{Proc. NeurIPS}, 2020{\natexlab{a}}.

\bibitem[Karras et~al.(2020{\natexlab{b}})Karras, Laine, Aittala, Hellsten, Lehtinen, and Aila]{karras2020analyzing}
Tero Karras, Samuli Laine, Miika Aittala, Janne Hellsten, Jaakko Lehtinen, and Timo Aila.
\newblock Analyzing and improving the image quality of stylegan.
\newblock In \emph{Proceedings of the IEEE/CVF Conference on Computer Vision and Pattern Recognition}, 2020{\natexlab{b}}.

\bibitem[Karras et~al.(2022)Karras, Aittala, Aila, and Laine]{karras2022elucidating}
Tero Karras, Miika Aittala, Timo Aila, and Samuli Laine.
\newblock Elucidating the design space of diffusion-based generative models.
\newblock \emph{Advances in Neural Information Processing Systems}, 2022.

\bibitem[Ke et~al.(2023)Ke, Obukhov, Huang, Metzger, Daudt, and Schindler]{ke2023repurposing}
Bingxin Ke, Anton Obukhov, Shengyu Huang, Nando Metzger, Rodrigo~Caye Daudt, and Konrad Schindler.
\newblock Repurposing diffusion-based image generators for monocular depth estimation.
\newblock \emph{arXiv preprint arXiv:2312.02145}, 2023.

\bibitem[Lee et~al.(2023)Lee, Tseng, Lee, and Yang]{lee2023dmp}
Hsin-Ying Lee, Hung-Yu Tseng, Hsin-Ying Lee, and Ming-Hsuan Yang.
\newblock Exploiting diffusion prior for generalizable pixel-level semantic prediction.
\newblock \emph{arXiv preprint arXiv:2311.18832}, 2023.

\bibitem[Li et~al.(2021)Li, Yang, Kreis, Torralba, and Fidler]{li2021semantic}
Daiqing Li, Junlin Yang, Karsten Kreis, Antonio Torralba, and Sanja Fidler.
\newblock Semantic segmentation with generative models: Semi-supervised learning and strong out-of-domain generalization.
\newblock In \emph{Proceedings of the IEEE/CVF Conference on Computer Vision and Pattern Recognition}, 2021.

\bibitem[Lin et~al.(2023)Lin, Liu, Li, and Yang]{lin2023common}
Shanchuan Lin, Bingchen Liu, Jiashi Li, and Xiao Yang.
\newblock Common diffusion noise schedules and sample steps are flawed.
\newblock \emph{arXiv preprint arXiv:2305.08891}, 2023.

\bibitem[Long et~al.(2015)Long, Shelhamer, and Darrell]{long2015fully}
Jonathan Long, Evan Shelhamer, and Trevor Darrell.
\newblock Fully convolutional networks for semantic segmentation.
\newblock In \emph{Proceedings of the IEEE conference on computer vision and pattern recognition}, 2015.

\bibitem[Lu et~al.(2022)Lu, Zhou, Bao, Chen, Li, and Zhu]{lu2022dpm}
Cheng Lu, Yuhao Zhou, Fan Bao, Jianfei Chen, Chongxuan Li, and Jun Zhu.
\newblock Dpm-solver++: Fast solver for guided sampling of diffusion probabilistic models.
\newblock \emph{arXiv preprint arXiv:2211.01095}, 2022.

\bibitem[Luo et~al.(2023{\natexlab{a}})Luo, Darrell, Wang, Goldman, and Holynski]{luo2023readoutguidance}
Grace Luo, Trevor Darrell, Oliver Wang, Dan~B Goldman, and Aleksander Holynski.
\newblock Readout guidance: Learning control from diffusion features.
\newblock \emph{arXiv preprint arXiv:2312.02150}, 2023{\natexlab{a}}.

\bibitem[Luo et~al.(2023{\natexlab{b}})Luo, Dunlap, Park, Holynski, and Darrell]{luo2023diffusion}
Grace Luo, Lisa Dunlap, Dong~Huk Park, Aleksander Holynski, and Trevor Darrell.
\newblock Diffusion hyperfeatures: Searching through time and space for semantic correspondence.
\newblock In \emph{Advances in Neural Information Processing Systems}, 2023{\natexlab{b}}.

\bibitem[Meng et~al.(2021)Meng, He, Song, Song, Wu, Zhu, and Ermon]{meng2021sdedit}
Chenlin Meng, Yutong He, Yang Song, Jiaming Song, Jiajun Wu, Jun-Yan Zhu, and Stefano Ermon.
\newblock Sdedit: Guided image synthesis and editing with stochastic differential equations.
\newblock In \emph{International Conference on Learning Representations}, 2021.

\bibitem[Nguyen et~al.(2023)Nguyen, Li, Ojha, and Lee]{nguyen2023visual}
Thao Nguyen, Yuheng Li, Utkarsh Ojha, and Yong~Jae Lee.
\newblock Visual instruction inversion: Image editing via visual prompting.
\newblock \emph{arXiv preprint arXiv:2307.14331}, 2023.

\bibitem[Noguchi \& Harada(2020)Noguchi and Harada]{RGBDGAN}
Atsuhiro Noguchi and Tatsuya Harada.
\newblock Rgbd-gan: Unsupervised 3d representation learning from natural image datasets via rgbd image synthesis.
\newblock In \emph{International Conference on Learning Representations}, 2020.

\bibitem[Oquab et~al.(2023)Oquab, Darcet, Moutakanni, Vo, Szafraniec, Khalidov, Fernandez, Haziza, Massa, El-Nouby, et~al.]{oquab2023dinov2}
Maxime Oquab, Timoth{\'e}e Darcet, Th{\'e}o Moutakanni, Huy Vo, Marc Szafraniec, Vasil Khalidov, Pierre Fernandez, Daniel Haziza, Francisco Massa, Alaaeldin El-Nouby, et~al.
\newblock Dinov2: Learning robust visual features without supervision.
\newblock \emph{arXiv preprint arXiv:2304.07193}, 2023.

\bibitem[Pan et~al.(2021)Pan, Dai, Liu, Loy, and Luo]{pan20202d}
Xingang Pan, Bo~Dai, Ziwei Liu, Chen~Change Loy, and Ping Luo.
\newblock Do 2d gans know 3d shape? unsupervised 3d shape reconstruction from 2d image gans.
\newblock In \emph{International Conference on Learning Representations}, 2021.

\bibitem[Podell et~al.(2023)Podell, English, Lacey, Blattmann, Dockhorn, M{\"u}ller, Penna, and Rombach]{podell2023sdxl}
Dustin Podell, Zion English, Kyle Lacey, Andreas Blattmann, Tim Dockhorn, Jonas M{\"u}ller, Joe Penna, and Robin Rombach.
\newblock Sdxl: Improving latent diffusion models for high-resolution image synthesis.
\newblock \emph{arXiv preprint arXiv:2307.01952}, 2023.

\bibitem[Ranftl et~al.(2021)Ranftl, Bochkovskiy, and Koltun]{ranftl2021vision}
Ren{\'e} Ranftl, Alexey Bochkovskiy, and Vladlen Koltun.
\newblock Vision transformers for dense prediction.
\newblock In \emph{Proceedings of the IEEE/CVF International Conference on Computer Vision}, 2021.

\bibitem[Razavi et~al.(2019)Razavi, Van~den Oord, and Vinyals]{razavi2019generating}
Ali Razavi, Aaron Van~den Oord, and Oriol Vinyals.
\newblock Generating diverse high-fidelity images with vq-vae-2.
\newblock \emph{Advances in neural information processing systems}, 32, 2019.

\bibitem[Rombach et~al.(2022)Rombach, Blattmann, Lorenz, Esser, and Ommer]{rombach2022high}
Robin Rombach, Andreas Blattmann, Dominik Lorenz, Patrick Esser, and Bj{\"o}rn Ommer.
\newblock High-resolution image synthesis with latent diffusion models.
\newblock In \emph{Proceedings of the IEEE/CVF Conference on Computer Vision and Pattern Recognition}, 2022.

\bibitem[Saharia et~al.(2022)Saharia, Chan, Saxena, Li, Whang, Denton, Ghasemipour, Gontijo~Lopes, Karagol~Ayan, Salimans, et~al.]{saharia2022photorealistic}
Chitwan Saharia, William Chan, Saurabh Saxena, Lala Li, Jay Whang, Emily~L Denton, Kamyar Ghasemipour, Raphael Gontijo~Lopes, Burcu Karagol~Ayan, Tim Salimans, et~al.
\newblock Photorealistic text-to-image diffusion models with deep language understanding.
\newblock \emph{Advances in Neural Information Processing Systems}, 2022.

\bibitem[Sariyildiz et~al.(2023)Sariyildiz, Alahari, Larlus, and Kalantidis]{sariyildiz2023fake}
Mert~Bulent Sariyildiz, Karteek Alahari, Diane Larlus, and Yannis Kalantidis.
\newblock Fake it till you make it: Learning transferable representations from synthetic imagenet clones.
\newblock In \emph{CVPR 2023--IEEE/CVF Conference on Computer Vision and Pattern Recognition}, 2023.

\bibitem[Sarkar et~al.(2023)Sarkar, Mai, Mahapatra, Lazebnik, and Bhattad]{sarkar2023shadows}
Ayush Sarkar, Hanlin Mai, Amitabh Mahapatra, Svetlana Lazebnik, and Anand Bhattad.
\newblock Shadows don't lie and lines can't bend! generative models don't know projective geometry... for now.
\newblock \emph{arXiv preprint arXiv:2311.17138}, 2023.

\bibitem[Sauer et~al.(2022)Sauer, Schwarz, and Geiger]{sauer2022stylegan}
Axel Sauer, Katja Schwarz, and Andreas Geiger.
\newblock Stylegan-xl: Scaling stylegan to large diverse datasets.
\newblock In \emph{ACM SIGGRAPH 2022 conference proceedings}, 2022.

\bibitem[Tang et~al.(2023)Tang, Jia, Wang, Phoo, and Hariharan]{tang2023emergent}
Luming Tang, Menglin Jia, Qianqian Wang, Cheng~Perng Phoo, and Bharath Hariharan.
\newblock Emergent correspondence from image diffusion.
\newblock \emph{arXiv preprint arXiv:2306.03881}, 2023.

\bibitem[Van~den Oord et~al.(2016)Van~den Oord, Kalchbrenner, Espeholt, Vinyals, Graves, et~al.]{van2016conditional}
Aaron Van~den Oord, Nal Kalchbrenner, Lasse Espeholt, Oriol Vinyals, Alex Graves, et~al.
\newblock Conditional image generation with pixelcnn decoders.
\newblock \emph{Advances in neural information processing systems}, 29, 2016.

\bibitem[Van Den~Oord et~al.(2016)Van Den~Oord, Kalchbrenner, and Kavukcuoglu]{van2016pixel}
A{\"a}ron Van Den~Oord, Nal Kalchbrenner, and Koray Kavukcuoglu.
\newblock Pixel recurrent neural networks.
\newblock In \emph{International conference on machine learning}. PMLR, 2016.

\bibitem[Vasiljevic et~al.(2019)Vasiljevic, Kolkin, Zhang, Luo, Wang, Dai, Daniele, Mostajabi, Basart, Walter, et~al.]{vasiljevic2019diode}
Igor Vasiljevic, Nick Kolkin, Shanyi Zhang, Ruotian Luo, Haochen Wang, Falcon~Z Dai, Andrea~F Daniele, Mohammadreza Mostajabi, Steven Basart, Matthew~R Walter, et~al.
\newblock Diode: A dense indoor and outdoor depth dataset.
\newblock \emph{arXiv preprint arXiv:1908.00463}, 2019.

\bibitem[Vincent(2011)]{vincent2011connection}
Pascal Vincent.
\newblock A connection between score matching and denoising autoencoders.
\newblock \emph{Neural computation}, 2011.

\bibitem[Wu et~al.(2023)Wu, Zhao, Chen, Gu, Zhao, He, Zhou, Shou, and Shen]{wu2023datasetdm}
Weijia Wu, Yuzhong Zhao, Hao Chen, Yuchao Gu, Rui Zhao, Yefei He, Hong Zhou, Mike~Zheng Shou, and Chunhua Shen.
\newblock Datasetdm: Synthesizing data with perception annotations using diffusion models.
\newblock \emph{Thirty-seventh Conference on Neural Information Processing Systems (NeurIPS 2023)}, 2023.

\bibitem[Wu et~al.(2021)Wu, Lischinski, and Shechtman]{wu2021stylespace}
Zongze Wu, Dani Lischinski, and Eli Shechtman.
\newblock Stylespace analysis: Disentangled controls for stylegan image generation.
\newblock In \emph{Proceedings of the IEEE/CVF Conference on Computer Vision and Pattern Recognition}, 2021.

\bibitem[Xu et~al.(2023)Xu, Liu, Vahdat, Byeon, Wang, and De~Mello]{xu2023odise}
Jiarui Xu, Sifei Liu, Arash Vahdat, Wonmin Byeon, Xiaolong Wang, and Shalini De~Mello.
\newblock {Open-Vocabulary Panoptic Segmentation with Text-to-Image Diffusion Models}.
\newblock \emph{arXiv preprint arXiv:2303.04803}, 2023.

\bibitem[Yang et~al.(2021)Yang, Shen, and Zhou]{yang2021semantic}
Ceyuan Yang, Yujun Shen, and Bolei Zhou.
\newblock Semantic hierarchy emerges in deep generative representations for scene synthesis.
\newblock \emph{International Journal of Computer Vision}, 2021.

\bibitem[Yu et~al.(2015)Yu, Seff, Zhang, Song, Funkhouser, and Xiao]{yu2015lsun}
Fisher Yu, Ari Seff, Yinda Zhang, Shuran Song, Thomas Funkhouser, and Jianxiong Xiao.
\newblock Lsun: Construction of a large-scale image dataset using deep learning with humans in the loop.
\newblock \emph{arXiv preprint arXiv:1506.03365}, 2015.

\bibitem[Yu et~al.(2022)Yu, Xu, Koh, Luong, Baid, Wang, Vasudevan, Ku, Yang, Ayan, Hutchinson, Han, Parekh, Li, Zhang, Baldridge, and Wu]{yu2022scaling}
Jiahui Yu, Yuanzhong Xu, Jing~Yu Koh, Thang Luong, Gunjan Baid, Zirui Wang, Vijay Vasudevan, Alexander Ku, Yinfei Yang, Burcu~Karagol Ayan, Ben Hutchinson, Wei Han, Zarana Parekh, Xin Li, Han Zhang, Jason Baldridge, and Yonghui Wu.
\newblock Scaling autoregressive models for content-rich text-to-image generation.
\newblock \emph{Transactions on Machine Learning Research}, 2022.
\newblock ISSN 2835-8856.

\bibitem[Yu et~al.(2023)Yu, Shi, Pasunuru, Muller, Golovneva, Wang, Babu, Tang, Karrer, Sheynin, et~al.]{yu2023scaling}
Lili Yu, Bowen Shi, Ramakanth Pasunuru, Benjamin Muller, Olga Golovneva, Tianlu Wang, Arun Babu, Binh Tang, Brian Karrer, Shelly Sheynin, et~al.
\newblock Scaling autoregressive multi-modal models: Pretraining and instruction tuning.
\newblock \emph{arXiv preprint arXiv:2309.02591}, 2023.

\bibitem[Yu et~al.(2021)Yu, Liu, Dundar, Tao, Catanzaro, Davis, and Fritz]{yu2021dual}
Ning Yu, Guilin Liu, Aysegul Dundar, Andrew Tao, Bryan Catanzaro, Larry~S Davis, and Mario Fritz.
\newblock Dual contrastive loss and attention for gans.
\newblock In \emph{Proceedings of the IEEE/CVF International Conference on Computer Vision}, 2021.

\bibitem[Yu \& Smith(2019)Yu and Smith]{yu2019inverserendernet}
Ye~Yu and William~AP Smith.
\newblock Inverserendernet: Learning single image inverse rendering.
\newblock In \emph{Proceedings of the IEEE Conference on Computer Vision and Pattern Recognition}, 2019.

\bibitem[Zhan et~al.(2023)Zhan, Zheng, Xie, and Zisserman]{zhan2023does}
Guanqi Zhan, Chuanxia Zheng, Weidi Xie, and Andrew Zisserman.
\newblock What does stable diffusion know about the 3d scene?
\newblock \emph{arXiv preprint arXiv:2310.06836}, 2023.

\bibitem[Zhang et~al.(2021{\natexlab{a}})Zhang, Chen, Ling, Gao, Zhang, Torralba, and Fidler]{zhang2020image}
Yuxuan Zhang, Wenzheng Chen, Huan Ling, Jun Gao, Yinan Zhang, Antonio Torralba, and Sanja Fidler.
\newblock Image gans meet differentiable rendering for inverse graphics and interpretable 3d neural rendering.
\newblock In \emph{International Conference on Learning Representations}, 2021{\natexlab{a}}.

\bibitem[Zhang et~al.(2021{\natexlab{b}})Zhang, Ling, Gao, Yin, Lafleche, Barriuso, Torralba, and Fidler]{zhang2021datasetgan}
Yuxuan Zhang, Huan Ling, Jun Gao, Kangxue Yin, Jean-Francois Lafleche, Adela Barriuso, Antonio Torralba, and Sanja Fidler.
\newblock Datasetgan: Efficient labeled data factory with minimal human effort.
\newblock In \emph{Proceedings of the IEEE/CVF Conference on Computer Vision and Pattern Recognition}, 2021{\natexlab{b}}.

\bibitem[Zhao et~al.(2023)Zhao, Rao, Liu, Liu, Zhou, and Lu]{zhao2023unleashing}
Wenliang Zhao, Yongming Rao, Zuyan Liu, Benlin Liu, Jie Zhou, and Jiwen Lu.
\newblock Unleashing text-to-image diffusion models for visual perception.
\newblock \emph{ICCV}, 2023.

\end{thebibliography}
\bibliographystyle{iclr2025_conference}

\clearpage
\appendix
\section{Additional Ablation Studies}

\subsection{Number of Diffusion Steps}
\begin{figure*}[ht]
\centering
\tiny
\setlength\tabcolsep{0pt}
\begin{tabular}{*{10}{>{\centering\arraybackslash}p{0.1\linewidth}}}
\multicolumn{3}{c}{Mean Angular Error\textdegree $\downarrow$ } & 25.83& 23.79&23.48&23.86&23.79&23.74&23.67\\
\hline
\multicolumn{3}{c}{L1 Error ($\times$ 100) $\downarrow$}&21.08&19.39&19.10&19.40&19.35&19.31&19.25\\
\multicolumn{10}{c}{\includegraphics[width=1\linewidth]{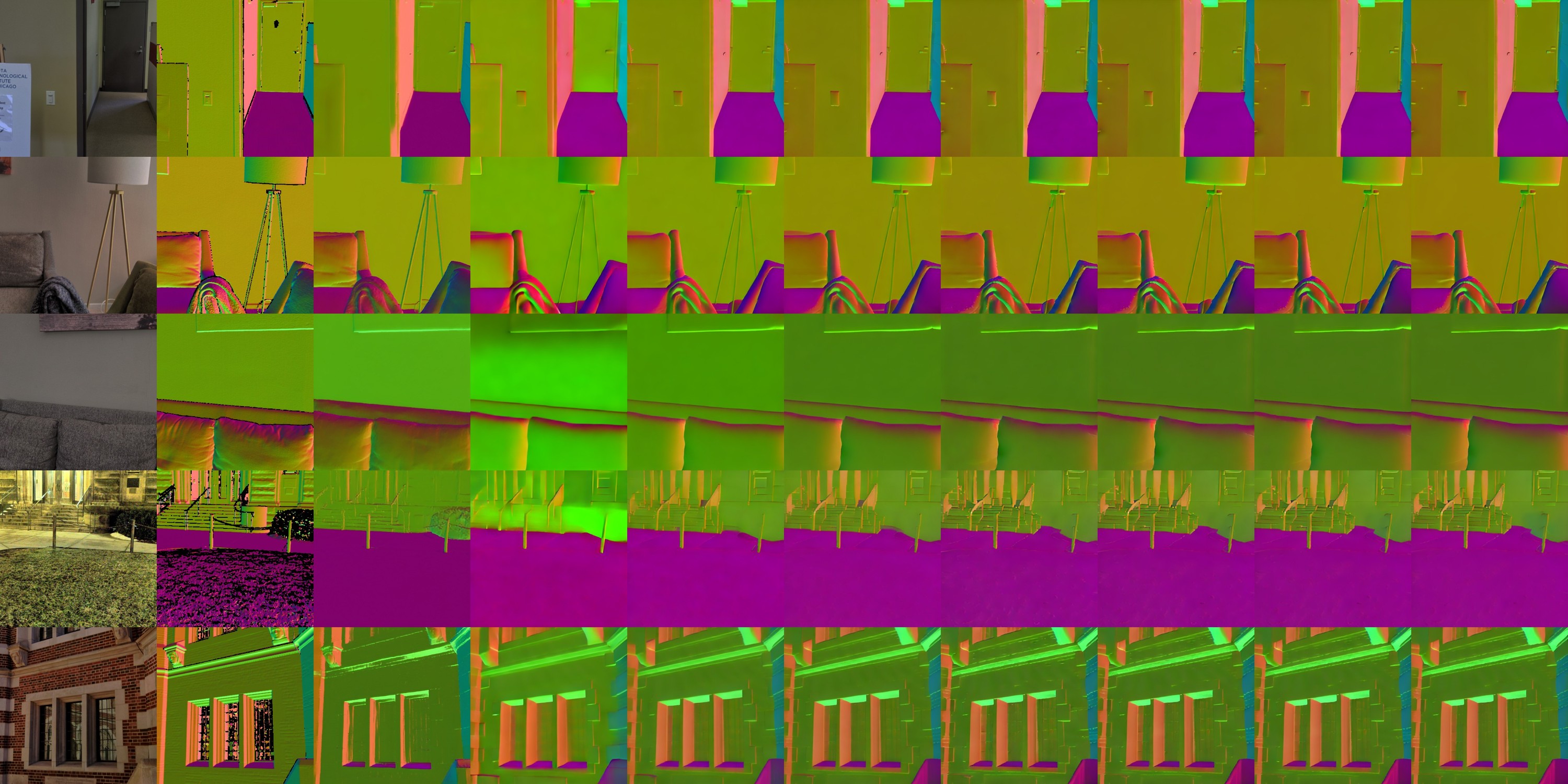}} \\

\centering Image & \centering GT & \centering Omni-v2~\cite{kar20223d} & \centering Steps=2 & \centering Steps=5 & \centering Steps=10 & \centering Steps=15 & \centering Steps=20 & \centering Steps=25 & \centering\arraybackslash Steps=50

\end{tabular}
\caption{Ablation study to determine the effect of varying numbers of diffusion steps while keeping CFG fixed at 3.0. Our findings show that there are very small differences, both in terms of quantity and quality, after 10 steps. For our main paper, we report results for 25 steps as it is more stable across different intrinsics.}
\label{fig:steps}
\end{figure*}

To assess the impact of the number of diffusion steps on the performance of the multi-step \augunet model, we conducted an ablation study. The results are presented in Fig.~\ref{fig:steps}. For all our experiments in the main text, we used DPMSolver++~\citep{lu2022dpm}. Interestingly, the quality of results did not vary significantly with an increased number of steps, indicating that 10 steps are sufficient for extracting better surface normals from the Stable Diffusion. Nevertheless, we use 25 steps for all our experiments because it is more stable across different image intrinsics.

\subsection{CFG scales}\label{sec:cfg}
\begin{figure*}[ht]
\centering
\tiny
\setlength\tabcolsep{0pt}
\begin{tabular}{*{10}{>{\centering\arraybackslash}p{0.1\linewidth}}}
\multicolumn{3}{c}{Mean Angular Error\textdegree $\downarrow$ } & 24.28& 23.48&25.72&27.80&29.85&31.93&34.12\\
\hline
\multicolumn{3}{c}{L1 Error ($\times$ 100) $\downarrow$}&19.48&19.10&21.01&22.72&24.36&26.03&27.78\\
\multicolumn{10}{c}{\includegraphics[width=1\linewidth]{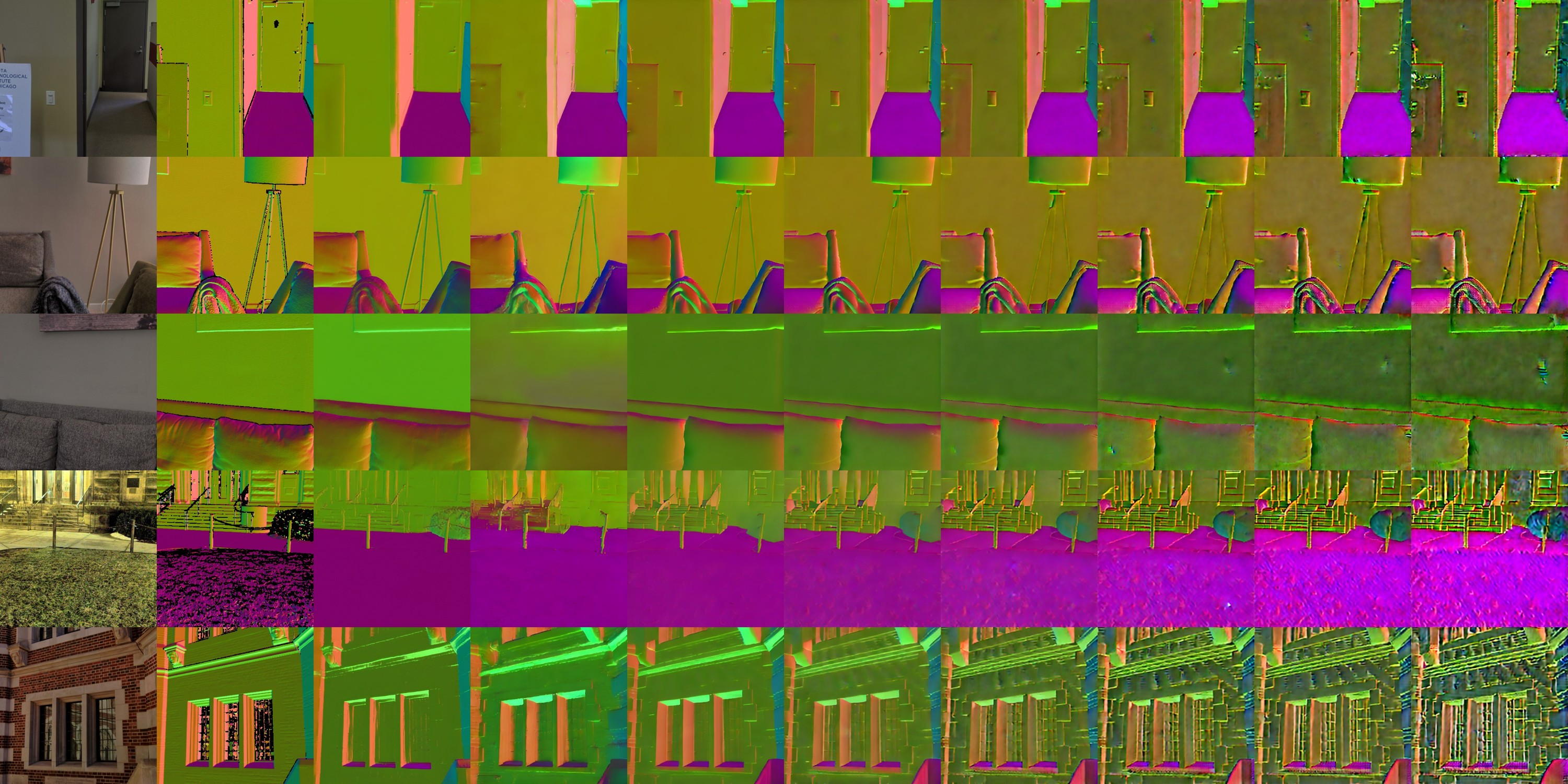}} \\
\centering Image & \centering GT & \centering Omni-v2~\cite{kar20223d} & \centering CFG=1 & \centering CFG=3 & \centering CFG=5 & \centering CFG=7 & \centering CFG=9 & \centering CFG=11 & \centering\arraybackslash CFG=13
\\
\end{tabular}

\caption{Ablation study analyzing the impact of different classifier-free guidance (CFG) on \augunet surface normal prediction. For efficiency, we experimented with a step of 10. We observed that CFG=1 sometimes led to incorrect semantic predictions, particularly in the case of stairs in row 4. On the other hand, using large CFGs (5 and beyond) results in more severe color shift problems.}
\label{fig:cfg}
\end{figure*}

When working with the multi-step \augunet, the quality of the final output is influenced by the choice of classifier-free guidance (CFG) scales during the inference process. In Fig.~\ref{fig:cfg}, we present a comparison of the effects of using different CFG scales. Based on our experiments, we found that using CFG=3.0 results in the best overall quality and minimizes color-shift artifacts.

\begin{figure*}[t!]
\centering
\tiny
  \setlength\tabcolsep{0pt}
  \renewcommand{\arraystretch}{0}
\begin{tabular}{cccccccc}
\includegraphics[width = 0.125\textwidth]{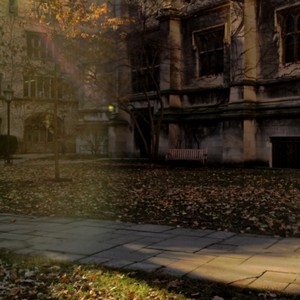}&
\includegraphics[width = 0.125\textwidth]{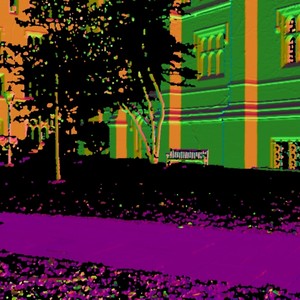}&
\includegraphics[width = 0.125\textwidth]{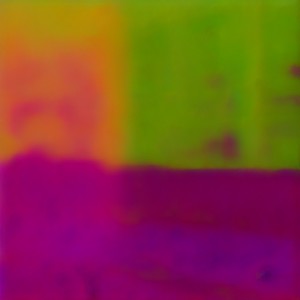}&
\includegraphics[width = 0.125\textwidth]{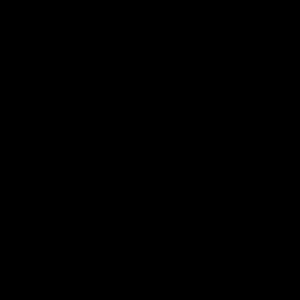}&
\includegraphics[width = 0.125\textwidth]{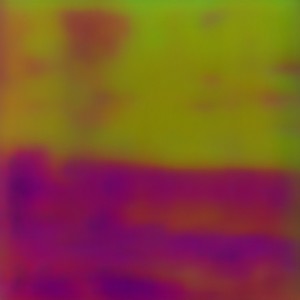}&
\includegraphics[width = 0.125\textwidth]{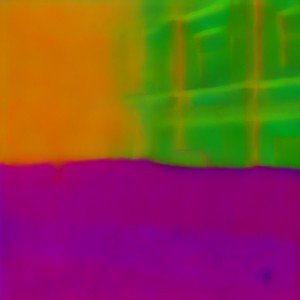}&
\includegraphics[width = 0.125\textwidth]{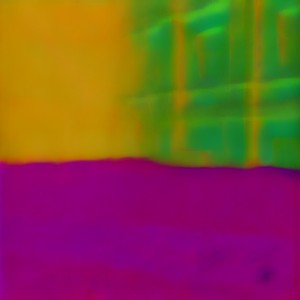}&
\includegraphics[width = 0.125\textwidth]{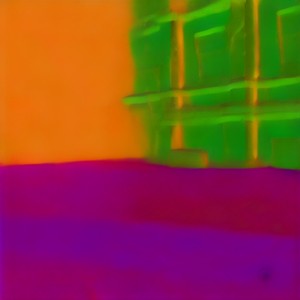}

\\
\vspace{2pt}
\\
Image & GT & Up blocks & Mid block & Down blocks & Cross-attn & Self-attn & All
\\
\midrule
\multicolumn{2}{c}{ \tiny \begin{tabular}{c} Mean Angular Error\textdegree $\downarrow$ \end{tabular}}&
32.25&-&36.71&23.72&21.70&20.31
\\
\multicolumn{2}{c}{ \tiny \begin{tabular}{c} L1 Error ($\times$ 100) $\downarrow$ \end{tabular}}&
26.10&-&29.95&19.27&17.69&16.53 \\
\midrule
\end{tabular}
\vspace{-10pt}
\caption{Ablation study on the effect of applying LoRA on different types of attention layers. We started all models with SD v1-5, 4000 training samples and LoRA rank=8. Training with LoRA only on the mid block never converges.
}
\label{fig:ablation_attns}
\end{figure*}
\begin{figure*}[t!]
\centering
\tiny
  \setlength\tabcolsep{0pt}
  \renewcommand{\arraystretch}{0}

\begin{tabular}{cccccc}
\includegraphics[width=0.16\textwidth]{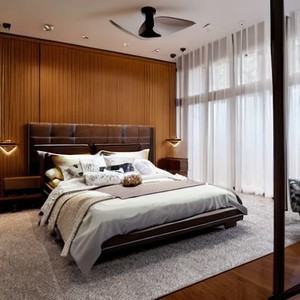}& 
\includegraphics[width=0.16\textwidth]{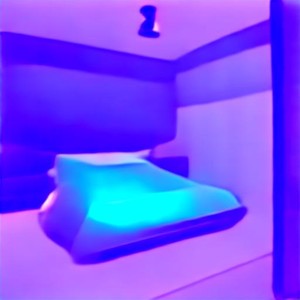}& 
\includegraphics[width=0.16\textwidth]{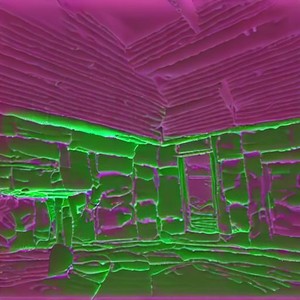}&
\includegraphics[width=0.16\textwidth]{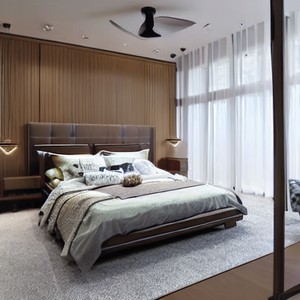}&
\includegraphics[width=0.16\textwidth]{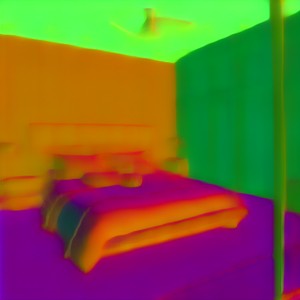}&
\includegraphics[width=0.16\textwidth]{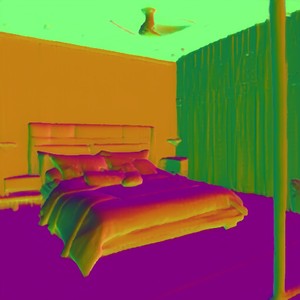}
\\
\vspace{2pt}
\\
Image & \multicolumn{1}{m{0.16\textwidth}}{\centering VISII~\citep{nguyen2023visual}} & \multicolumn{1}{m{0.16\textwidth}}{\centering Text. Invers.~\citep{gal2022image}} & \multicolumn{1}{m{0.16\textwidth}}{\centering IP2P~\citep{brooks2023instructpix2pix}}  & \textbf{LoRA (Ours)} & Pseudo GT
\end{tabular}
\caption{Comparison of image editing techniques for surface normal mapping. VISII and Textual Inversion yield unsatisfactory results, while InstructPix2Pix fails to interpret the task, resulting in near-original output.}
\label{fig:visii}
\end{figure*}

\section{Other Ablations and Baselines} \label{sec:ablations}
We extensively study the effect of applying LoRA to different attention layers within Stable Diffusion models. Specifically, we investigate the outcomes of targeting up-blocks, mid-block, down-blocks, cross-attention, and self-attention layers individually. We find (\cref{fig:ablation_attns}) that isolating LoRA to up or down blocks or the mid-block alone is less effective or diverges, and applying to either cross- or self-attention layers yields decent results, though combining them is best.

Additionally, we evaluated other image editing methods such as Textual Inversion~\citep{gal2022image} and VISII~\citep{nguyen2023visual}, alongside InstructPix2Pix's response to ``Turn it into a surface normal map'' instruction~\citep{brooks2023instructpix2pix}. As shown in \cref{fig:visii}, these methods perform poorly for intrinsic image extraction, demonstrating the effectiveness of the LoRA approach in extracting scene intrinsics. 

\begin{table*}[t]
    \centering
    \scriptsize
    \caption{Comparison of quality of normals extracted from StyleGAN \citet{bhattad2023stylegan}.}
    \resizebox{.9\linewidth}{!}{%
    \begin{tabular}{c c c c}
        \toprule
         & Mean Error\textdegree $\downarrow$ & Median Error\textdegree $\downarrow$ & L1 {\tiny $\times$ 100} $\downarrow$  \\
        \midrule 
        
       ``StyleGAN knows"~\citep{bhattad2023stylegan} & 19.92 & 46.65 & 16.64\\
        LoRA-StyleGAN (Ours) & {\bf 13.24} & {\bf 23.55} & {\bf 10.92} \\
        \bottomrule
    \end{tabular}
    
    }
    
    \label{tab:stylegan_knows_comparison}
\end{table*}

\begin{figure*}[t]
\centering
	\scriptsize
  \setlength\tabcolsep{0.pt}
  \renewcommand{\arraystretch}{0.}

  \begin{tabular}{cccc}

    \includegraphics[width=0.175\linewidth]{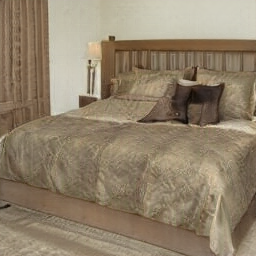}&
    \includegraphics[width=0.175\linewidth]{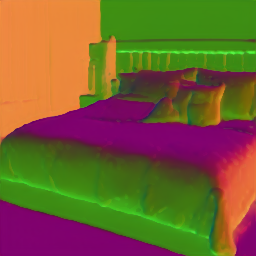}&
    \includegraphics[width=0.175\linewidth]{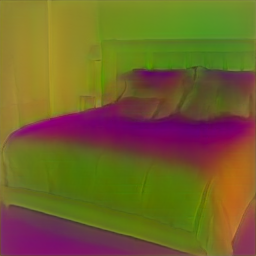}
    &
    \includegraphics[width=0.175\linewidth]{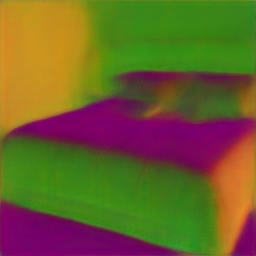}
    \\ \vspace{2pt}\\
    Image & Pseudo GT & \citet{bhattad2023stylegan} & Ours
\\
        \end{tabular}
      \caption{
       Qualitative results of normals extracted from StyleGAN by \citet{bhattad2023stylegan} and Ours.  
}
        \label{fig:rebuttal_baseline_images}
\end{figure*}

We also provide a comparison with \citet{bhattad2023stylegan} in \cref{tab:stylegan_knows_comparison} and \cref{fig:rebuttal_baseline_images}. This comparison is for the same 500 randomly generated images. Ours outperforms \citet{bhattad2023stylegan} significantly.

\begin{figure*}[t!]
\centering
\scriptsize
\setlength{\tabcolsep}{0pt}
\renewcommand{\arraystretch}{0}

\begin{tabular}{ccccc}
\includegraphics[width=0.15\linewidth]{images/motivation_for_imagePrior/00000_RGB.jpg}& 
\includegraphics[width=0.15\linewidth]{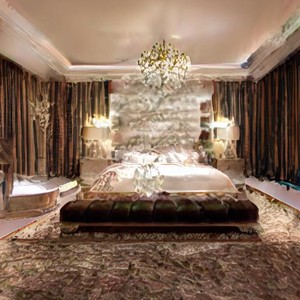}& 
\includegraphics[width=0.15\linewidth]{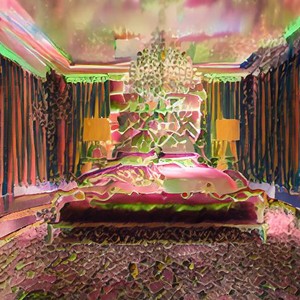}& 
\includegraphics[width=0.15\linewidth]{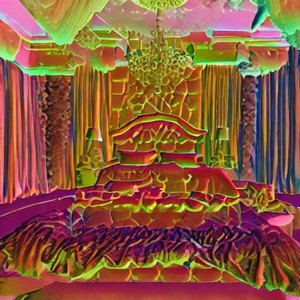}& 
\includegraphics[width=0.15\linewidth]{images/motivation_for_imagePrior/00000_NORMAL_sd.jpg}
\\
\vspace{2pt}
\\
Image & s=0.2 & s=0.5 & s=0.7 & s=1.0
\end{tabular}
\caption{We observe applying SDEdit method on the SDv1-5 model alone, without incorporating the additional input image latent encoding, fails to produce satisfactorily aligned and high-quality scene intrinsics. The reason for this might be the considerable domain shift that exists between RGB images and surface normal maps, which results in severe artifacts when using SDEdit. The variable ``s'' represents the strength of SDEdit.}
\label{fig:sdedit}
\end{figure*}

In addition, 
we show that directly applying SDEdit~\citep{meng2021sdedit} will also fail to extract reasonable image intrinsics. We take the model from the SDv1-5 column in Fig.13 of the main paper and apply SDEdit. In Fig.~\ref{fig:sdedit}, we show directly applying SDEdit results in severe artifacts, regardless of strength. 

\section{Hyper-parameters}\label{sec:hyperparameters}
In Table~\ref{tab:hyperparams}, we show the hyperparameters we use for each model. 

\begin{table*}[ht]
    \centering
    \caption{Hyper-parameters for each model. LR refers to the learning rate and BS refers to the batch size. Please note that the number of steps required to reach convergence reported above is for normal/depth. However, it is worth noting that albedo and shading tend to require significantly fewer steps to converge (usually half of normal/depth). Additionally, \augunet (multi-step) and SD-UNet (single-step) are trained on real-world DIODE dataset, while the other models are trained on synthetic images within a specific domain. (Num. of params of VQGAN counts transformer + first stage models; Num. of params of \augunet and SD-UNet counts VAE+UNet)}
    \resizebox{\linewidth}{!}{%
    \begin{tabular}{c  c c c c c c c c}
        \toprule
        Model & Dataset & Resolution & Rank & LR & BS 
        & LoRA Params 
        & Generator Params 
        & Convergence Steps \\
        \midrule
        VQGAN       & FFHQ         & 256 & 8 & 1e-03 & 1 
        & 0.13M
        & 873.9M
        & $\sim$ 4000\\
        StyleGAN-v2 & FFHQ         & 256 & 8 & 1e-03 & 1 
        & 0.14M
        & 24.8M
        & $\sim$ 4000\\
        StyleGAN-v2 & LSUN Bedroom & 256 & 8 & 1e-03 & 1 
        & 0.14M
        & 24.8M
        & $\sim$ 4000\\
        StyleGAN-XL & FFHQ         & 256 & 8 & 1e-03 & 1 
        & 0.19M
        & 67.9M
        &$\sim$ 4000\\
        StyleGAN-XL & ImageNet     & 256 & 8 & 1e-03 & 1 
        & 0.19M
        & 67.9M
        &$\sim$ 4000\\      
        \augunet (multi step)    & Open  & 512 & 8 & 1e-04 & 4 
        & 1.59M
        & 943.2M
        & $\sim$ 30000\\
        SD-UNet (single step)    & Open  &  512  & 8 & 1e-04 & 4 
        & 1.59M
        & 943.2M
        &$\sim$ 15000\\
        \bottomrule
    \end{tabular}
    }
    \label{tab:hyperparams}
\end{table*}

\section{Generated Images Used for Quantitative Analysis}
In Tab. 2 of the main paper, we report quantitative results on synthetic images. For Autoregressive models and GANs, we first randomly sample 500 noises and use them to generate 500 RGB images. The same 500 noises will then be used to generate intrinsics with our learned LoRAs loaded. For Stable Diffusion experiments (both single-step and multi-step), we use a single dataset with 1000 synthetic images with various prompts.

The pseudo GT are obtained by applying SOTA off-the-shelf models on the RGB images.

\section{Additional Qualitative Results}
\begin{figure*}[t!]
\centering
	\tiny
  \setlength\tabcolsep{0pt}
  \renewcommand{\arraystretch}{0}

\begin{tabular}{cccccc}
    \includegraphics[width=0.15\linewidth]{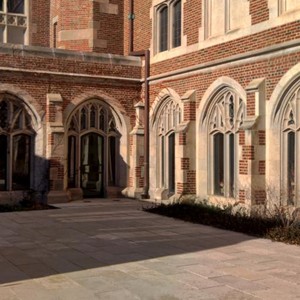}
    &
    \includegraphics[width=0.15\linewidth]{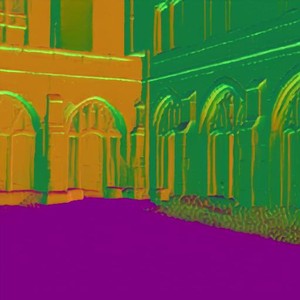} 
    &
    \includegraphics[width=0.15\linewidth]{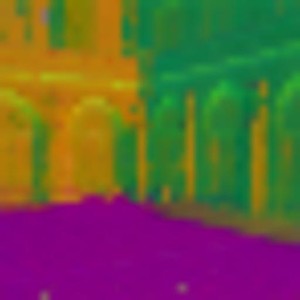}
    &
    \includegraphics[width=0.15\linewidth]{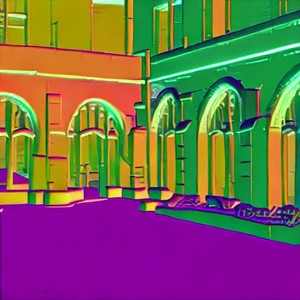} 
    &
    \includegraphics[width=0.15\linewidth]{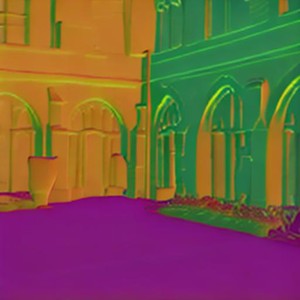}
    &
    \includegraphics[width=0.15\linewidth]{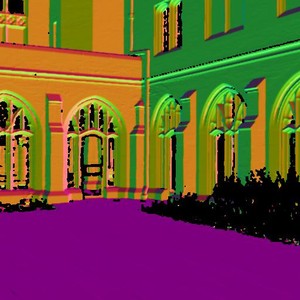}  
\\
    &\includegraphics[width=0.15\linewidth]{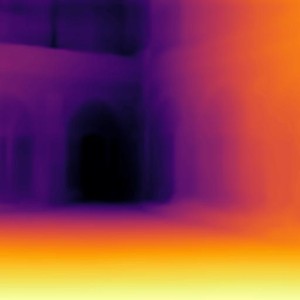}
    &
    \includegraphics[width=0.15\linewidth]{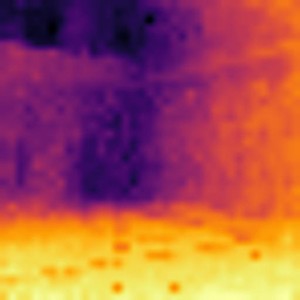}
    &
    \includegraphics[width=0.15\linewidth]{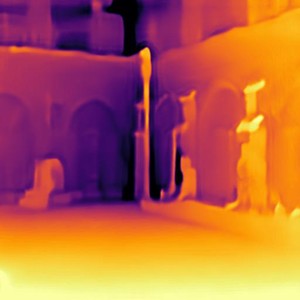}
    &
    \includegraphics[width=0.15\linewidth]{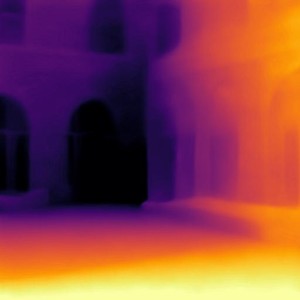}
    &
   \includegraphics[width=0.15\linewidth]{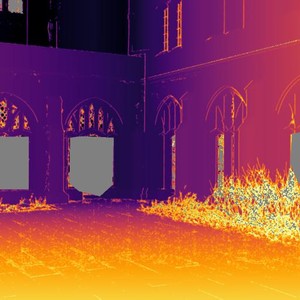}
\\
    \includegraphics[width=0.15\linewidth]{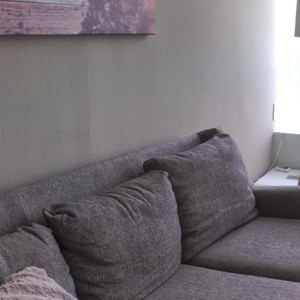}
    &
    \includegraphics[width=0.15\linewidth]{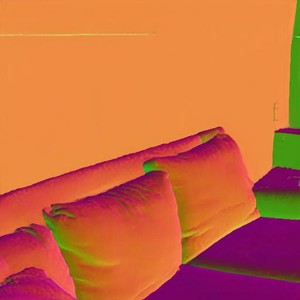} 
    &
    \includegraphics[width=0.15\linewidth]{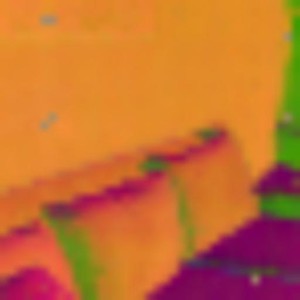}
    &
    \includegraphics[width=0.15\linewidth]{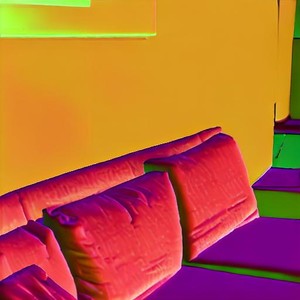} 
    &
    \includegraphics[width=0.15\linewidth]{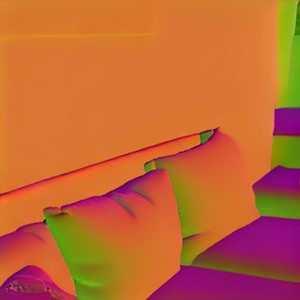}
    &
    \includegraphics[width=0.15\linewidth]{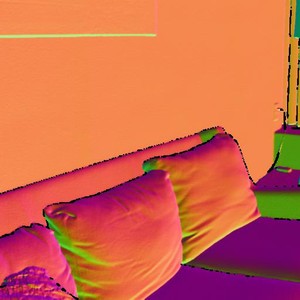}  
\\
    & \includegraphics[width=0.15\linewidth]{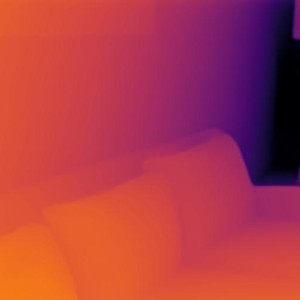}
    &
    \includegraphics[width=0.15\linewidth]{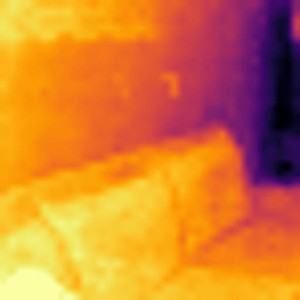}
    &
    \includegraphics[width=0.15\linewidth]{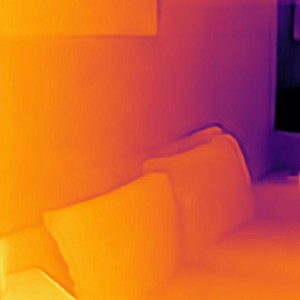}
    &
    \includegraphics[width=0.15\linewidth]{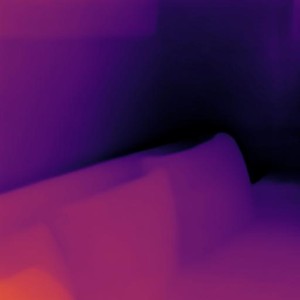}
    &
   \includegraphics[width=0.15\linewidth]{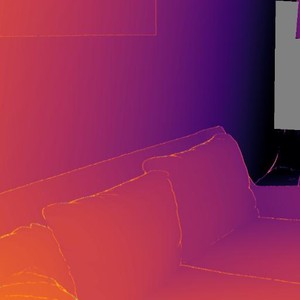}
\\
    \includegraphics[width=0.15\linewidth]{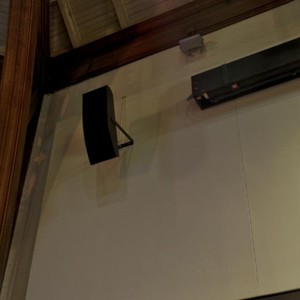}
    &
    \includegraphics[width=0.15\linewidth]{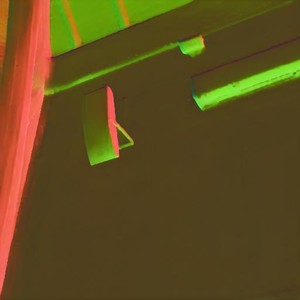} 
    &
    \includegraphics[width=0.15\linewidth]{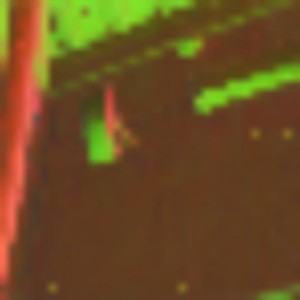}
    &
    \includegraphics[width=0.15\linewidth]{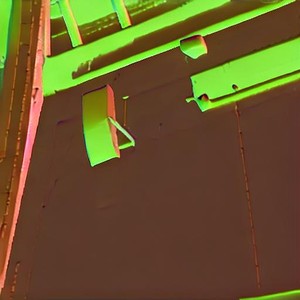} 
    &
    \includegraphics[width=0.15\linewidth]{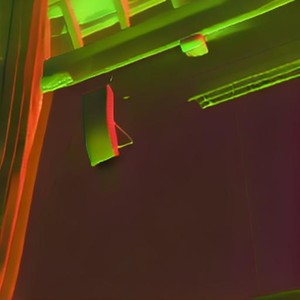}
    &
    \includegraphics[width=0.15\linewidth]{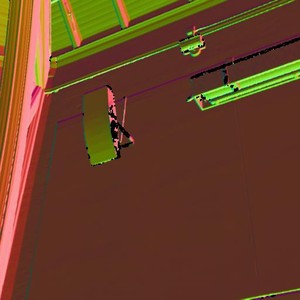} 
\\
    & \includegraphics[width=0.15\linewidth]{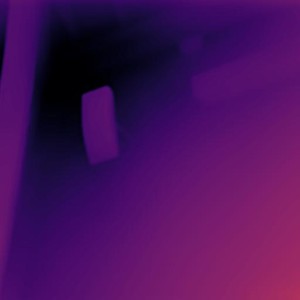}
    &
    \includegraphics[width=0.15\linewidth]{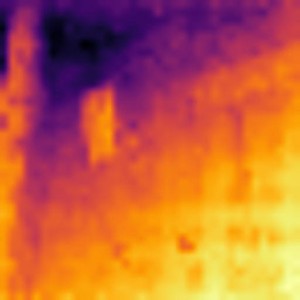}
    &
    \includegraphics[width=0.15\linewidth]{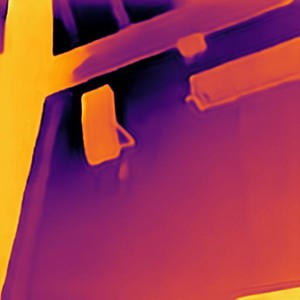}
    &
    \includegraphics[width=0.15\linewidth]{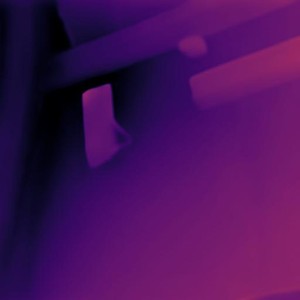}
    &
   \includegraphics[width=0.15\linewidth]{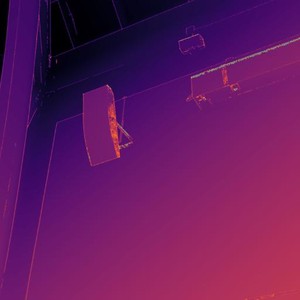}
\\
    \includegraphics[width=0.15\linewidth]{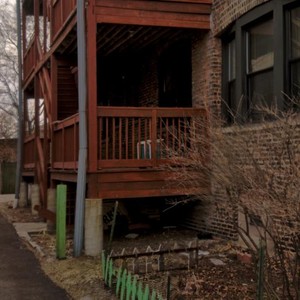}
    &
    \includegraphics[width=0.15\linewidth]{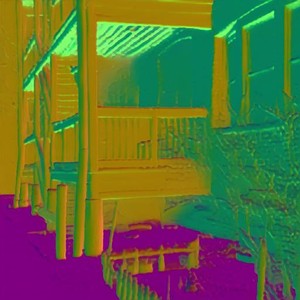} 
    &
    \includegraphics[width=0.15\linewidth]{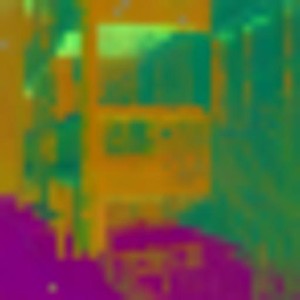}
    &
    \includegraphics[width=0.15\linewidth]{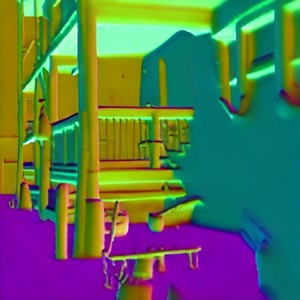} 
    &
    \includegraphics[width=0.15\linewidth]{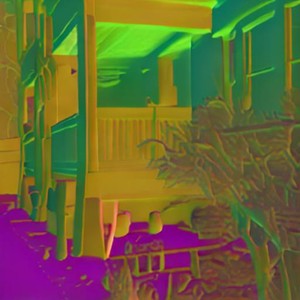}
    &
    \includegraphics[width=0.15\linewidth]{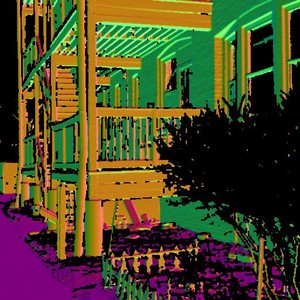}  
\\
    & \includegraphics[width=0.15\linewidth]{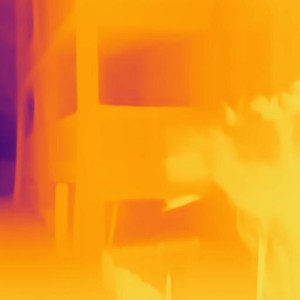}
    &
    \includegraphics[width=0.15\linewidth]{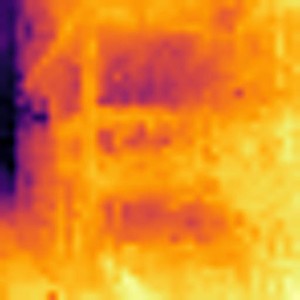}
    &
    \includegraphics[width=0.15\linewidth]{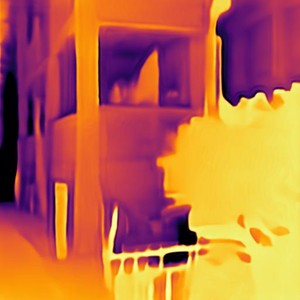}
    &
    \includegraphics[width=0.15\linewidth]{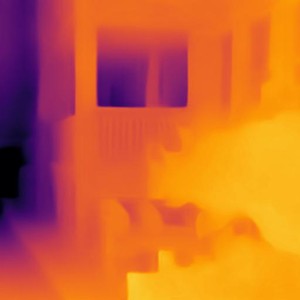}
    &
   \includegraphics[width=0.15\linewidth]{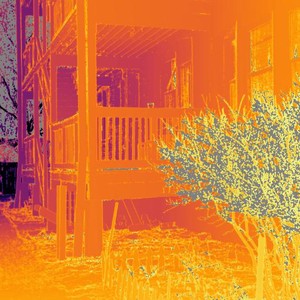}
\\
    \vspace{5pt} 
\\
  Real & Pseudo GT & DINOv2 & \augold & \augunet & GT
\\
        \end{tabular}
      \caption{
       Additional results after applying improved diffusion techniques with \augunetnospace. \augunet was found to significantly reduce color shift artifacts observed in \augold during the extraction of detailed scene intrinsic results.
}
        \label{fig:color_shift_supp}
	\end{figure*}
\begin{figure*}[!tbh]
\centering
\scriptsize
\setlength\tabcolsep{0pt}
  \renewcommand{\arraystretch}{0}
\begin{tabular}{ccccccccccc}
&&\multicolumn{2}{c}{Surface Normals} & \multicolumn{2}{c}{Depth} & \multicolumn{2}{c}{Albedo} & \multicolumn{2}{c}{Shading} \\
\multirow{2}{*}{\rotatebox[origin=c]{90}{\parbox[c]{.5cm}{\centering VQGAN}}} & 
\includegraphics[width=0.107\linewidth]{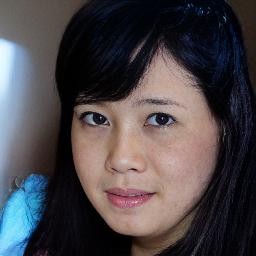} & 
\includegraphics[width=0.107\linewidth]{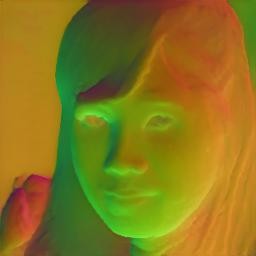} & 
\includegraphics[width=0.107\linewidth]{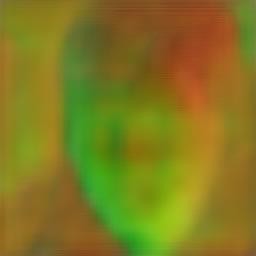} & 
\includegraphics[width=0.107\linewidth]{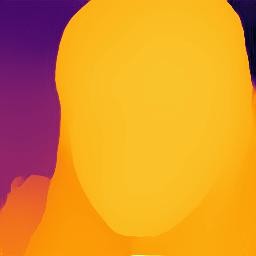} & 
\includegraphics[width=0.107\linewidth]{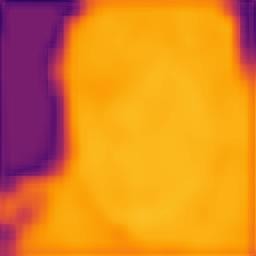} & 
\includegraphics[width=0.107\linewidth]{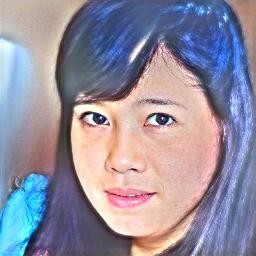} & 
\includegraphics[width=0.107\linewidth]{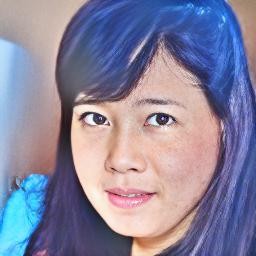} & 
\includegraphics[width=0.107\linewidth]{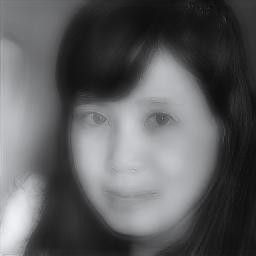} & 
\includegraphics[width=0.107\linewidth]{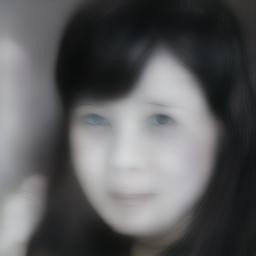} \\
& \includegraphics[width=0.107\linewidth]{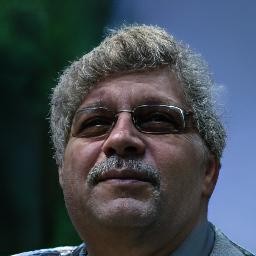} & 
\includegraphics[width=0.107\linewidth]{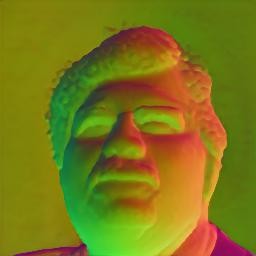} & 
\includegraphics[width=0.107\linewidth]{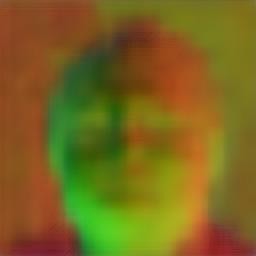} & 
\includegraphics[width=0.107\linewidth]{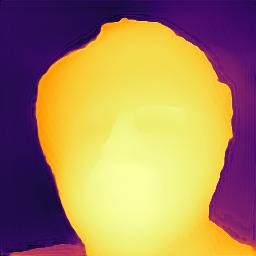} & 
\includegraphics[width=0.107\linewidth]{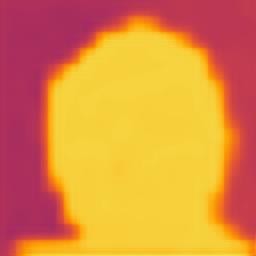} & 
\includegraphics[width=0.107\linewidth]{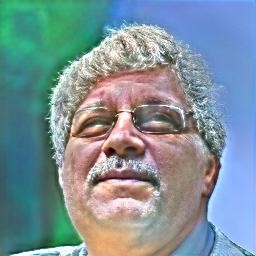} & 
\includegraphics[width=0.107\linewidth]{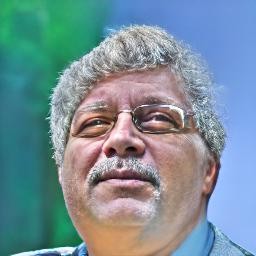} & 
\includegraphics[width=0.107\linewidth]{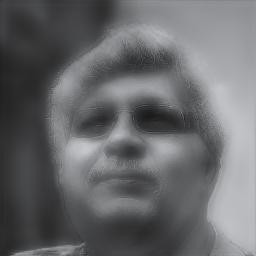} & 
\includegraphics[width=0.107\linewidth]{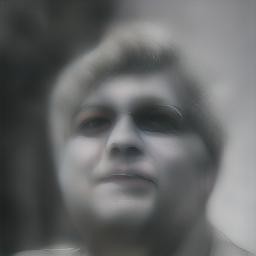} 
\\
\midrule
\multirow{2}{*}{\rotatebox[origin=c]{90}{\parbox[c]{.8cm}{\centering StyleGANv2}}} &
\includegraphics[width=0.107\linewidth]{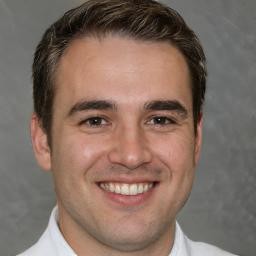} & 
\includegraphics[width=0.107\linewidth]{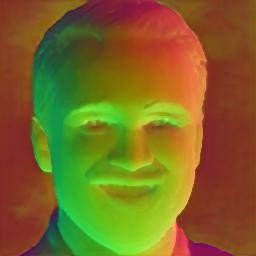} & 
\includegraphics[width=0.107\linewidth]{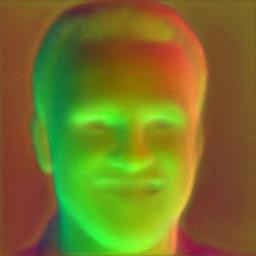} & 
\includegraphics[width=0.107\linewidth]{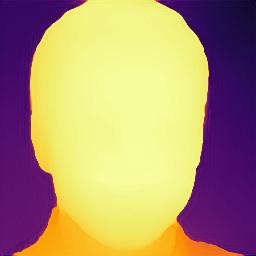} & 
\includegraphics[width=0.107\linewidth]{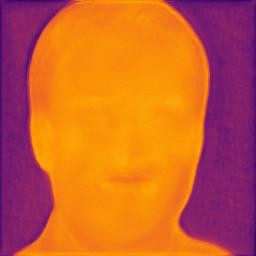} & 
\includegraphics[width=0.107\linewidth]{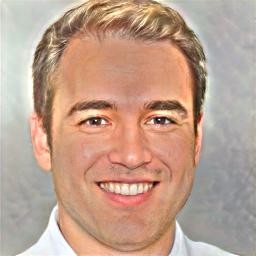} & 
\includegraphics[width=0.107\linewidth]{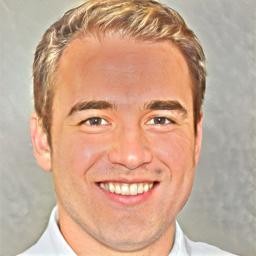} & 
\includegraphics[width=0.107\linewidth]{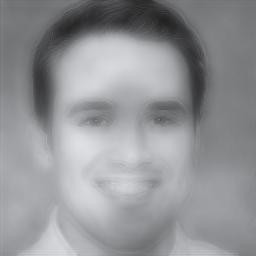} & 
\includegraphics[width=0.107\linewidth]{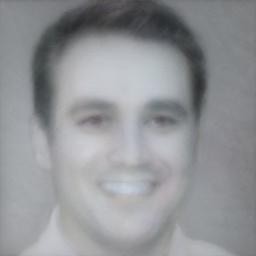}
\\
& \includegraphics[width=0.107\linewidth]{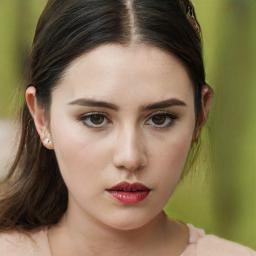} & 
\includegraphics[width=0.107\linewidth]{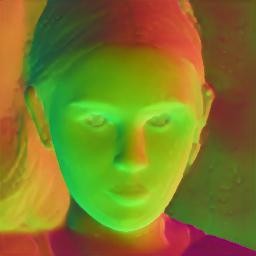} & 
\includegraphics[width=0.107\linewidth]{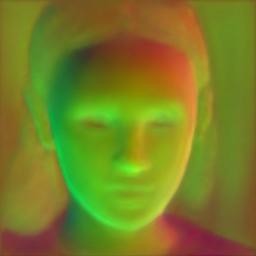} & 
\includegraphics[width=0.107\linewidth]{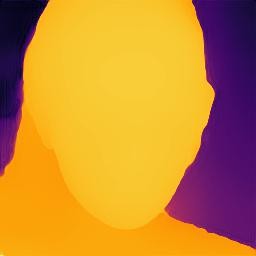} & 
\includegraphics[width=0.107\linewidth]{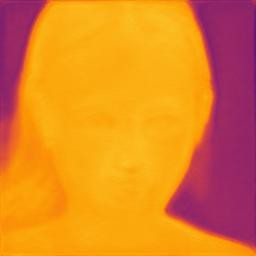} & 
\includegraphics[width=0.107\linewidth]{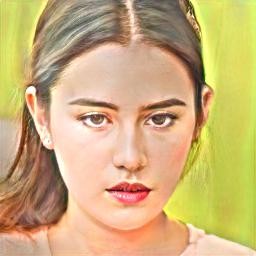} & 
\includegraphics[width=0.107\linewidth]{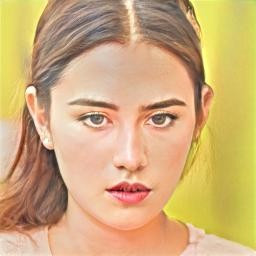} & 
\includegraphics[width=0.107\linewidth]{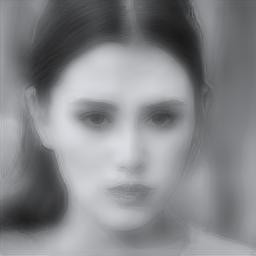} & 
\includegraphics[width=0.107\linewidth]{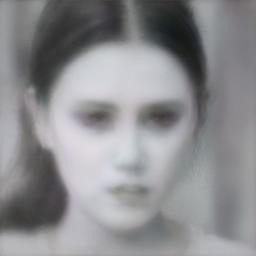}
\\
\midrule
\multirow{2}{*}{\rotatebox[origin=c]{90}{\parbox[c]{.8cm}{\centering StyleGANXL}}} & \includegraphics[width=0.107\linewidth]{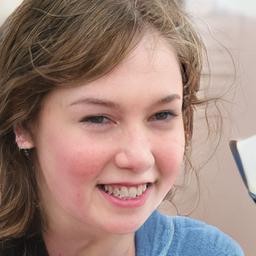} & 
\includegraphics[width=0.107\linewidth]{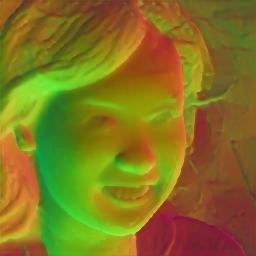} & 
\includegraphics[width=0.107\linewidth]{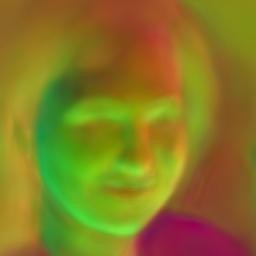} & 
\includegraphics[width=0.107\linewidth]{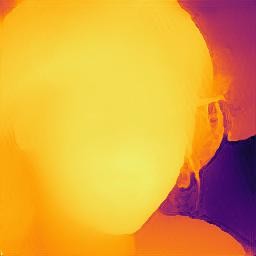} & 
\includegraphics[width=0.107\linewidth]{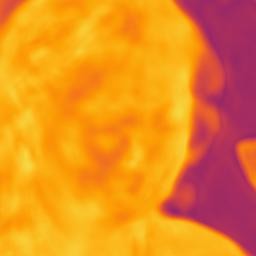} & 
\includegraphics[width=0.107\linewidth]{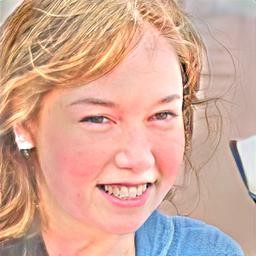} & 
\includegraphics[width=0.107\linewidth]{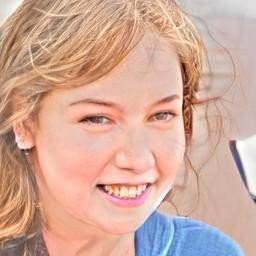} & 
\includegraphics[width=0.107\linewidth]{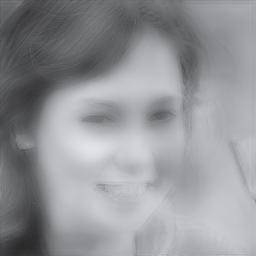} & 
\includegraphics[width=0.107\linewidth]{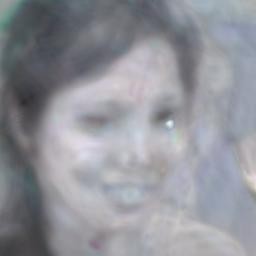} 
\\
& \includegraphics[width=0.107\linewidth]{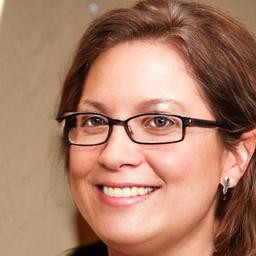} & 
\includegraphics[width=0.107\linewidth]{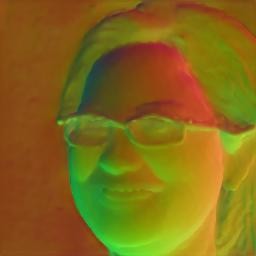} & 
\includegraphics[width=0.107\linewidth]{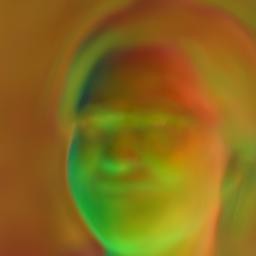} & 
\includegraphics[width=0.107\linewidth]{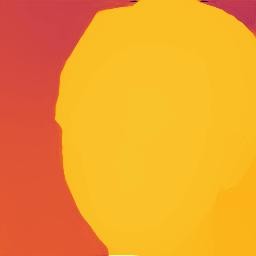} & 
\includegraphics[width=0.107\linewidth]{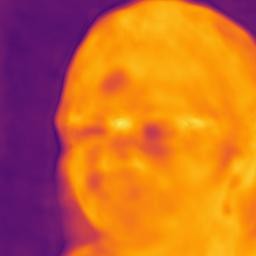} & 
\includegraphics[width=0.107\linewidth]{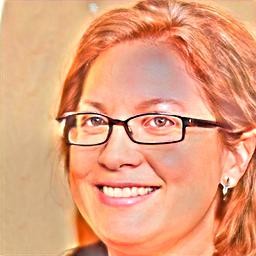} & 
\includegraphics[width=0.107\linewidth]{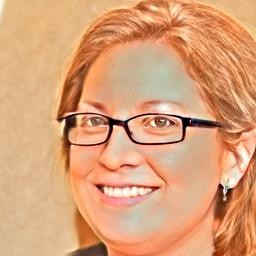} & 
\includegraphics[width=0.107\linewidth]{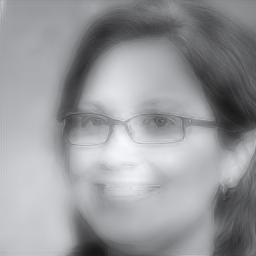} & 
\includegraphics[width=0.107\linewidth]{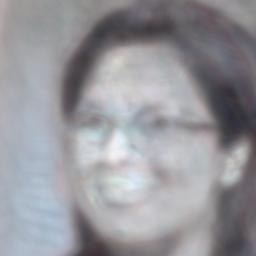} 
\\
\vspace{2pt}
\\
& \tiny Image & \multicolumn{1}{m{0.107\textwidth}}{\centering \tiny OD-v2~\citep{kar20223d}} & \tiny  \textbf{Recovered} & \multicolumn{1}{m{0.107\textwidth}}{\centering \tiny ZoeD~\citep{bhat2023zoedepth}} & \tiny \textbf{Recovered} & \multicolumn{1}{m{0.107\textwidth}}{\centering \tiny  PD~\citep{bhattad2022cut}} & \tiny  \textbf{Recovered} & \multicolumn{1}{m{0.107\textwidth}}{\centering \tiny  PD~\citep{bhattad2022cut}} & \tiny  \textbf{Recovered} 
\end{tabular}
\caption{Additional results of scene intrinsics from different generators -- VQGAN, StyleGAN-v2, and StyleGAN-XL -- trained on FFHQ dataset.}
\label{fig:generators_comparison_supp}
\end{figure*}

\begin{figure*}[ht]
\centering
\scriptsize
\setlength\tabcolsep{0pt}
\renewcommand{\arraystretch}{0}
\begin{tabular}{ccccccccc}
&\multicolumn{2}{c}{Surface Normals} & \multicolumn{2}{c}{Depth} & \multicolumn{2}{c}{Albedo} & \multicolumn{2}{c}{Shading} \\
\includegraphics[width=0.11\linewidth]{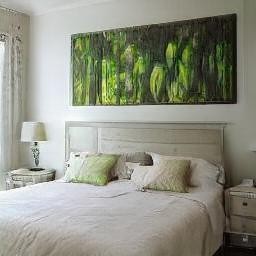} & 
\includegraphics[width=0.11\linewidth]{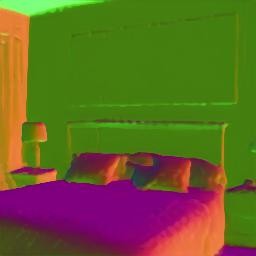} & 
\includegraphics[width=0.11\linewidth]{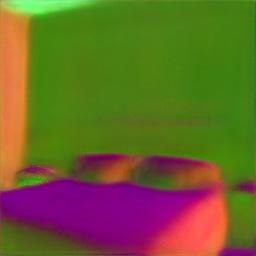} & 
\includegraphics[width=0.11\linewidth]{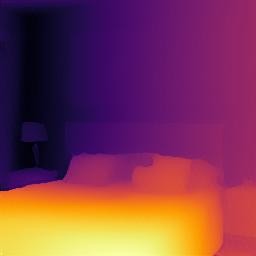} & 
\includegraphics[width=0.11\linewidth]{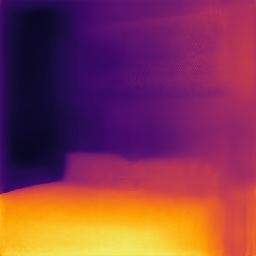} & 
\includegraphics[width=0.11\linewidth]{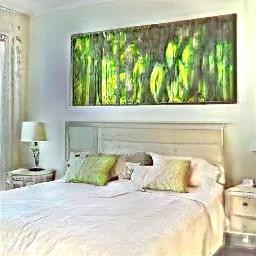} & 
\includegraphics[width=0.11\linewidth]{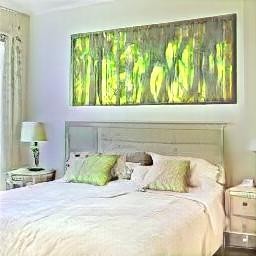} & 
\includegraphics[width=0.11\linewidth]{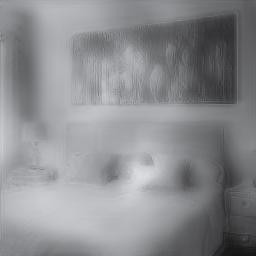} & 
\includegraphics[width=0.11\linewidth]{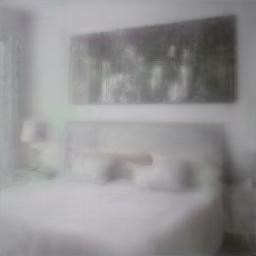} 
\\
\includegraphics[width=0.11\linewidth]{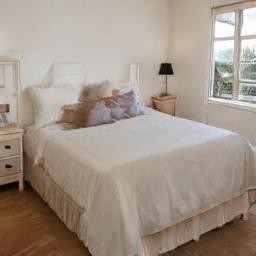} & 
\includegraphics[width=0.11\linewidth]{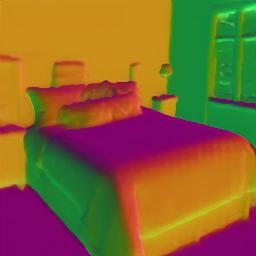} & 
\includegraphics[width=0.11\linewidth]{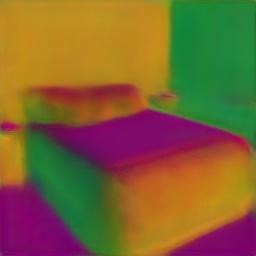} & 
\includegraphics[width=0.11\linewidth]{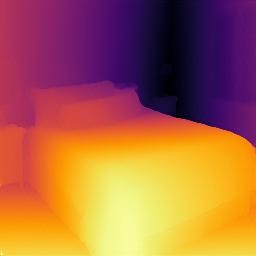} & 
\includegraphics[width=0.11\linewidth]{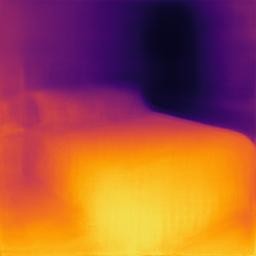} & 
\includegraphics[width=0.11\linewidth]{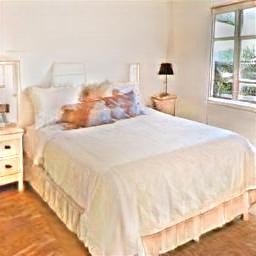} & 
\includegraphics[width=0.11\linewidth]{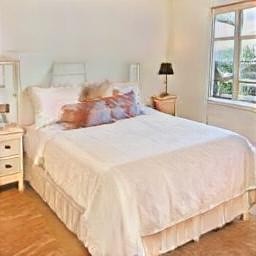} & 
\includegraphics[width=0.11\linewidth]{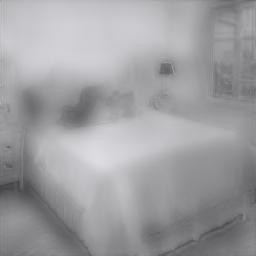} & 
\includegraphics[width=0.11\linewidth]{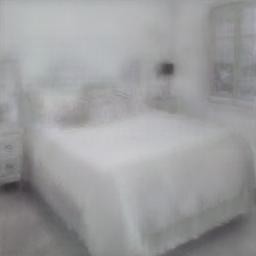} 
\\
\includegraphics[width=0.11\linewidth]{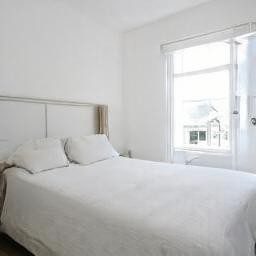} & 
\includegraphics[width=0.11\linewidth]{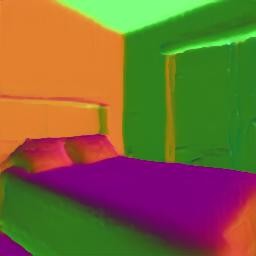} & 
\includegraphics[width=0.11\linewidth]{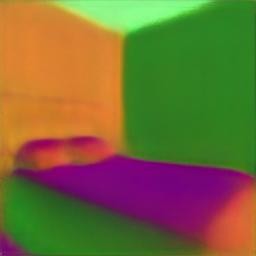} & 
\includegraphics[width=0.11\linewidth]{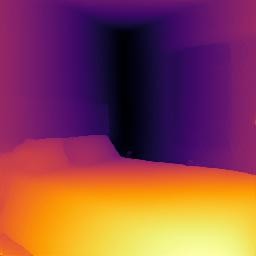} & 
\includegraphics[width=0.11\linewidth]{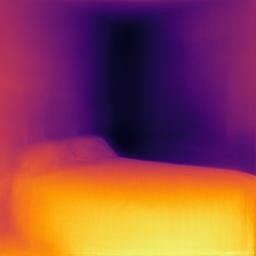} & 
\includegraphics[width=0.11\linewidth]{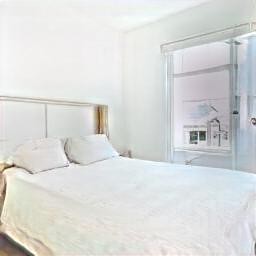} & 
\includegraphics[width=0.11\linewidth]{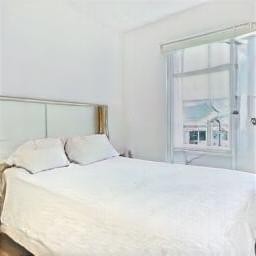} & 
\includegraphics[width=0.11\linewidth]{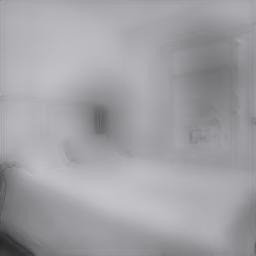} & 
\includegraphics[width=0.11\linewidth]{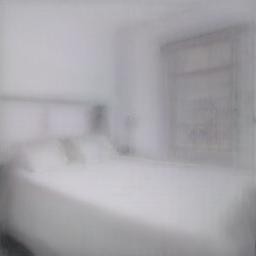} 
\\
\includegraphics[width=0.11\linewidth]{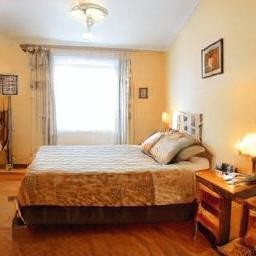} & 
\includegraphics[width=0.11\linewidth]{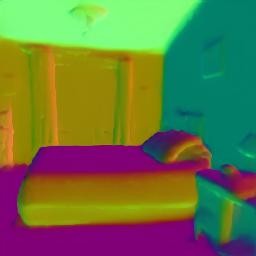} & 
\includegraphics[width=0.11\linewidth]{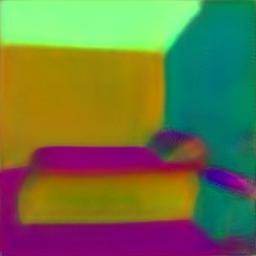} & 
\includegraphics[width=0.11\linewidth]{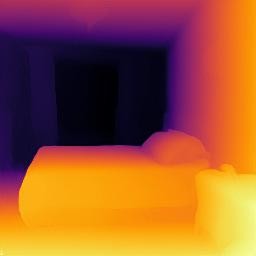} & 
\includegraphics[width=0.11\linewidth]{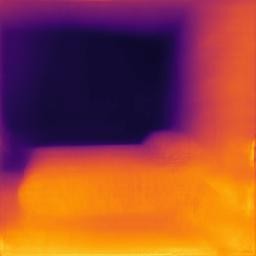} & 
\includegraphics[width=0.11\linewidth]{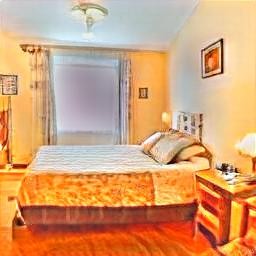} & 
\includegraphics[width=0.11\linewidth]{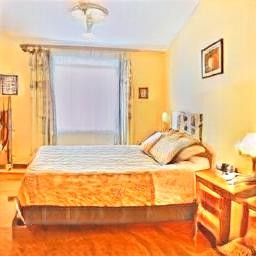} & 
\includegraphics[width=0.11\linewidth]{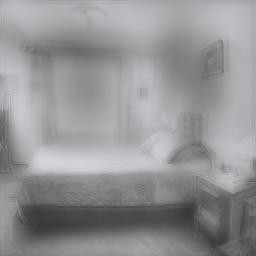} & 
\includegraphics[width=0.11\linewidth]{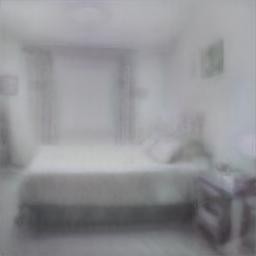} 
\\
\includegraphics[width=0.11\linewidth]{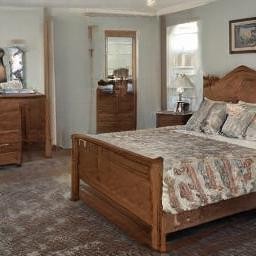} & 
\includegraphics[width=0.11\linewidth]{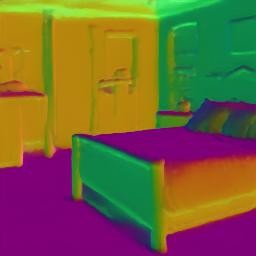} & 
\includegraphics[width=0.11\linewidth]{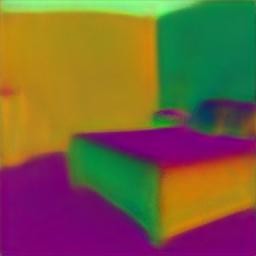} & 
\includegraphics[width=0.11\linewidth]{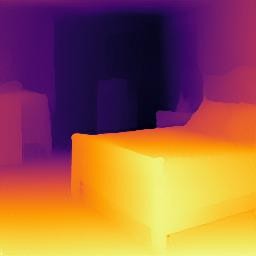} & 
\includegraphics[width=0.11\linewidth]{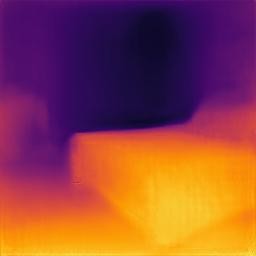} & 
\includegraphics[width=0.11\linewidth]{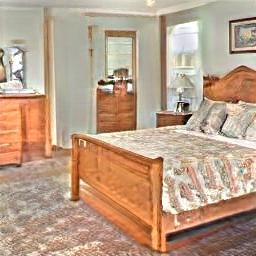} & 
\includegraphics[width=0.11\linewidth]{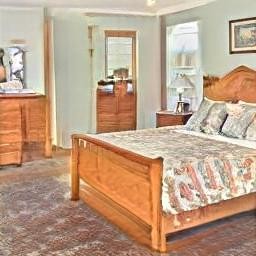} & 
\includegraphics[width=0.11\linewidth]{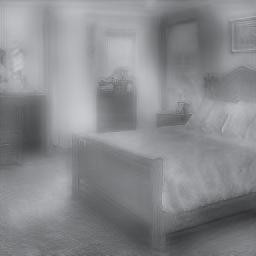} & 
\includegraphics[width=0.11\linewidth]{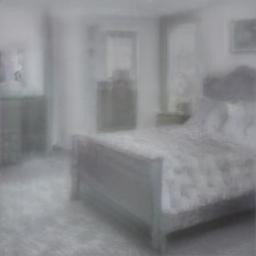} \\
\vspace{2pt}
\\
\tiny Image & \multicolumn{1}{m{0.11\textwidth}}{\centering \tiny OD-v2~\citep{kar20223d}} & \tiny  \textbf{Recovered} & \multicolumn{1}{m{0.11\textwidth}}{\centering \tiny ZoeD~\citep{bhat2023zoedepth}} & \tiny \textbf{Recovered} & \multicolumn{1}{m{0.11\textwidth}}{\centering \tiny  PD~\citep{bhattad2022cut}} & \tiny  \textbf{Recovered} & \multicolumn{1}{m{0.11\textwidth}}{\centering \tiny  PD~\citep{bhattad2022cut}} & \tiny  \textbf{Recovered} 

\end{tabular}
\caption{Additional results of scene intrinsics extraction from Stylegan-v2 trained on LSUN bedroom images.}
\label{fig:bedroom_supp}
\end{figure*}

\begin{figure*}[ht]
\centering
\scriptsize
  \setlength\tabcolsep{0pt}
  \renewcommand{\arraystretch}{0}
\begin{tabular}{ccccccccc}
&\multicolumn{2}{c}{Surface Normals} & \multicolumn{2}{c}{Depth} & \multicolumn{2}{c}{Albedo} & \multicolumn{2}{c}{Shading} \\
\includegraphics[width=0.11\textwidth]{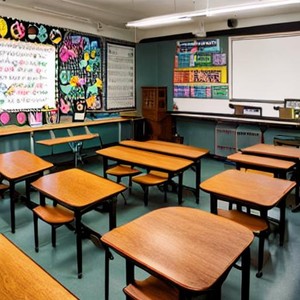} & 
\includegraphics[width=0.11\textwidth]{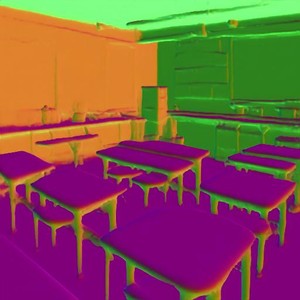} & 
\includegraphics[width=0.11\textwidth]{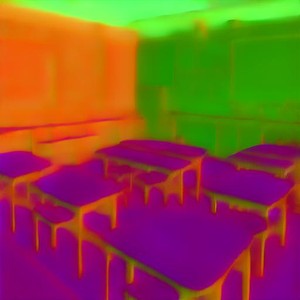} & 
\includegraphics[width=0.11\textwidth]{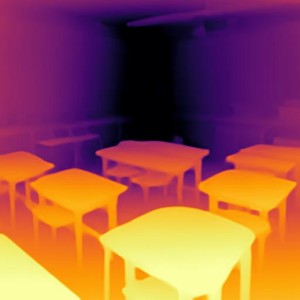} & 
\includegraphics[width=0.11\textwidth]{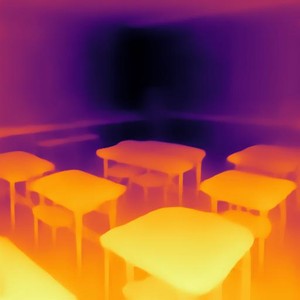} & 
\includegraphics[width=0.11\textwidth]{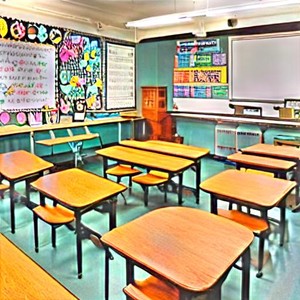} & 
\includegraphics[width=0.11\textwidth]{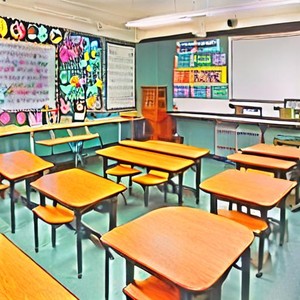} & 
\includegraphics[width=0.11\textwidth]{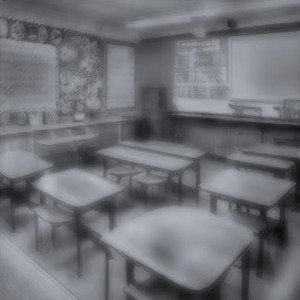} & 
\includegraphics[width=0.11\textwidth]{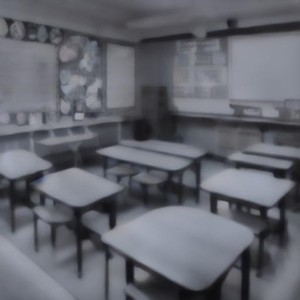} 
\\
\includegraphics[width=0.11\textwidth]{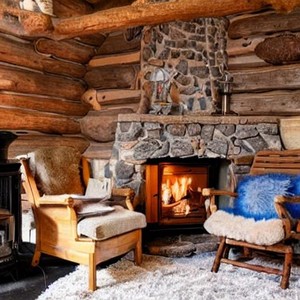} & 
\includegraphics[width=0.11\textwidth]{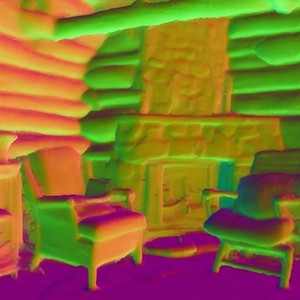} & 
\includegraphics[width=0.11\textwidth]{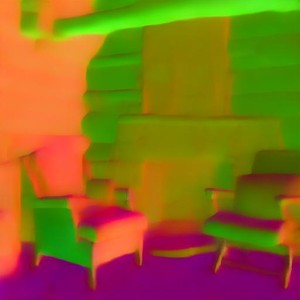} & 
\includegraphics[width=0.11\textwidth]{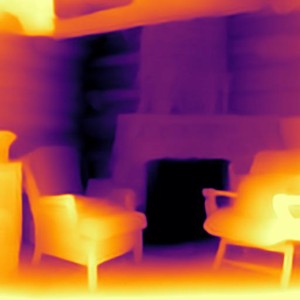} & 
\includegraphics[width=0.11\textwidth]{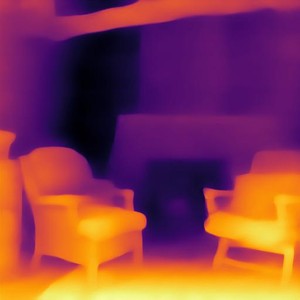} & 
\includegraphics[width=0.11\textwidth]{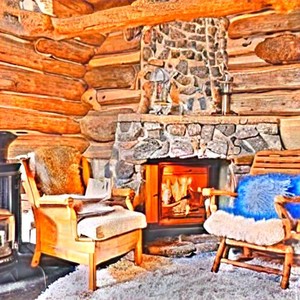} & 
\includegraphics[width=0.11\textwidth]{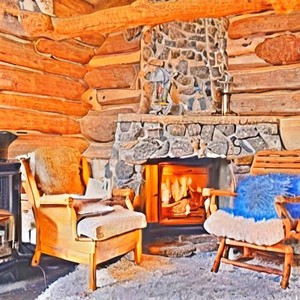} & 
\includegraphics[width=0.11\textwidth]{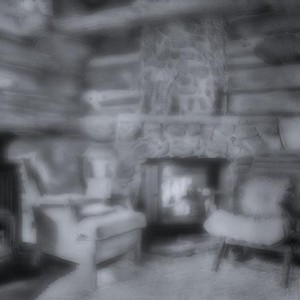} & 
\includegraphics[width=0.11\textwidth]{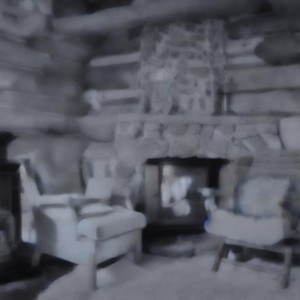} 
\\
\includegraphics[width=0.11\textwidth]{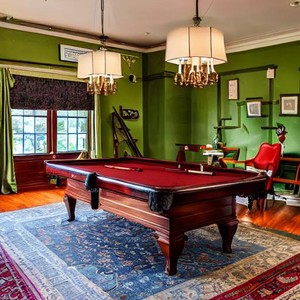} & 
\includegraphics[width=0.11\textwidth]{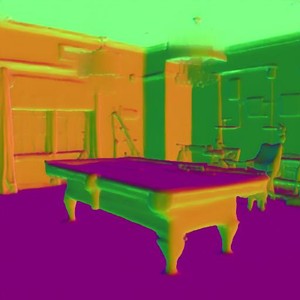} & 
\includegraphics[width=0.11\textwidth]{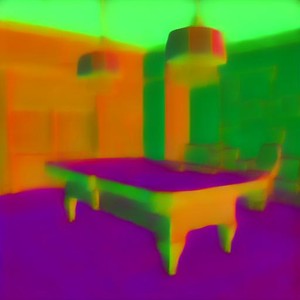} & 
\includegraphics[width=0.11\textwidth]{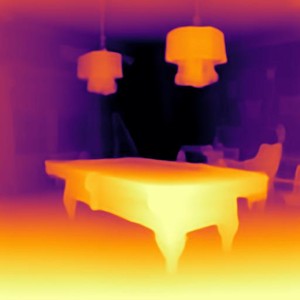} & 
\includegraphics[width=0.11\textwidth]{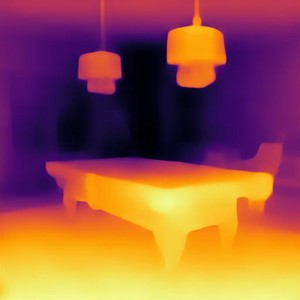} & 
\includegraphics[width=0.11\textwidth]{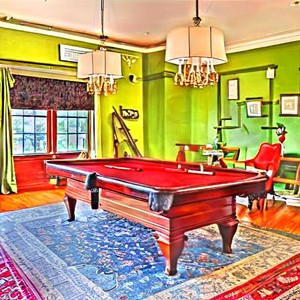} & 
\includegraphics[width=0.11\textwidth]{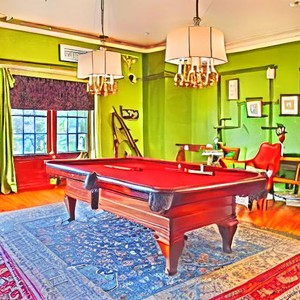} & 
\includegraphics[width=0.11\textwidth]{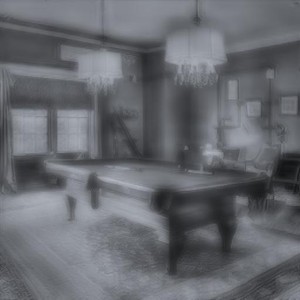} & 
\includegraphics[width=0.11\textwidth]{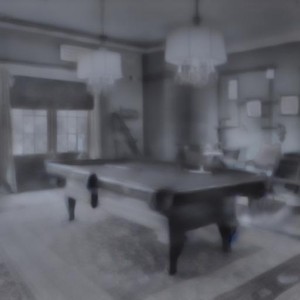} 
\\
\includegraphics[width=0.11\textwidth]{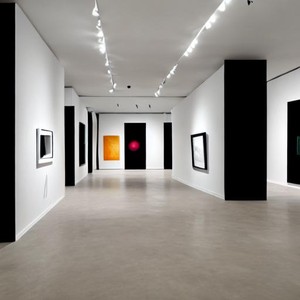} & 
\includegraphics[width=0.11\textwidth]{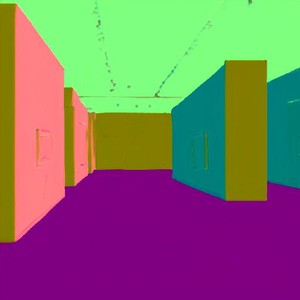} & 
\includegraphics[width=0.11\textwidth]{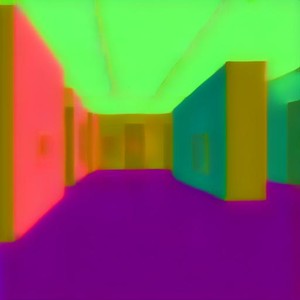} & 
\includegraphics[width=0.11\textwidth]{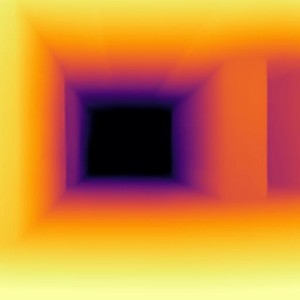} & 
\includegraphics[width=0.11\textwidth]{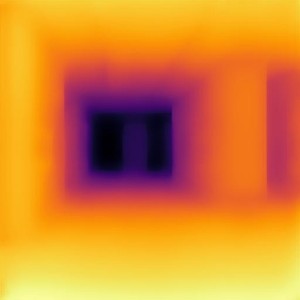} & 
\includegraphics[width=0.11\textwidth]{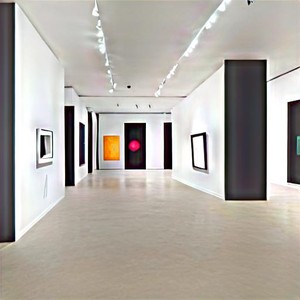} & 
\includegraphics[width=0.11\textwidth]{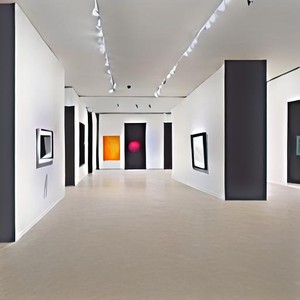} & 
\includegraphics[width=0.11\textwidth]{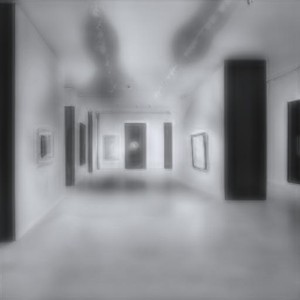} & 
\includegraphics[width=0.11\textwidth]{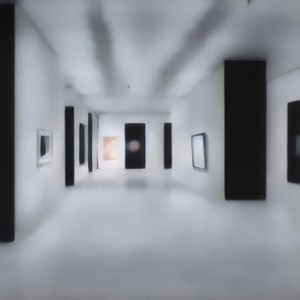} 
\\
\includegraphics[width=0.11\textwidth]{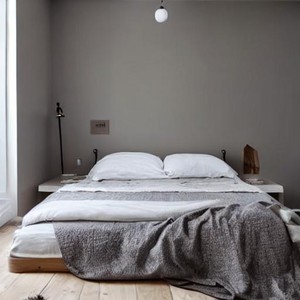} & 
\includegraphics[width=0.11\textwidth]{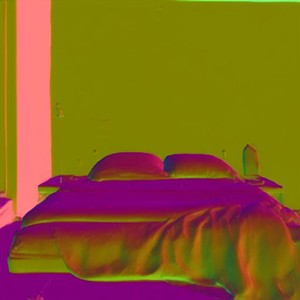} & 
\includegraphics[width=0.11\textwidth]{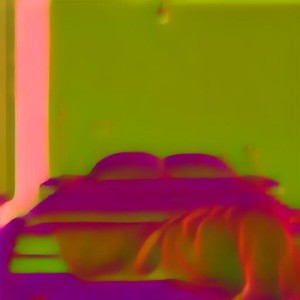} & 
\includegraphics[width=0.11\textwidth]{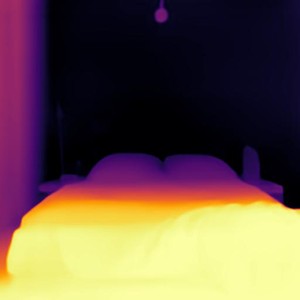} & 
\includegraphics[width=0.11\textwidth]{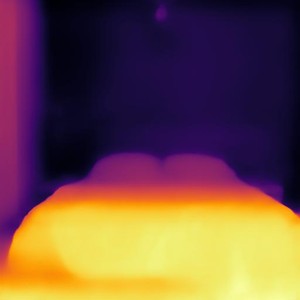} & 
\includegraphics[width=0.11\textwidth]{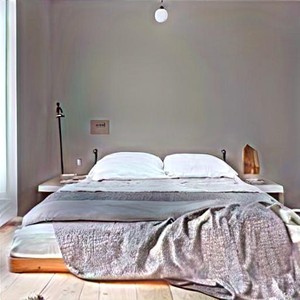} & 
\includegraphics[width=0.11\textwidth]{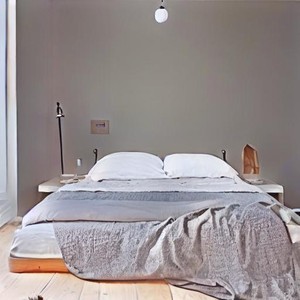} & 
\includegraphics[width=0.11\textwidth]{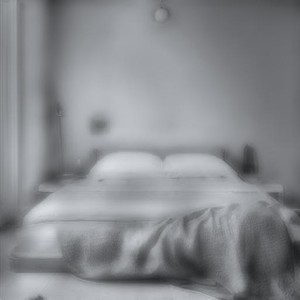} & 
\includegraphics[width=0.11\textwidth]{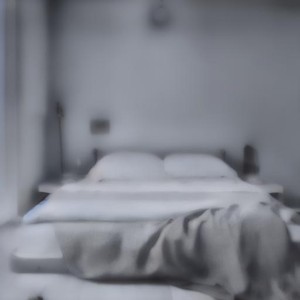} 
\\
\vspace{2pt}
\\
\tiny Image & \multicolumn{1}{m{0.11\textwidth}}{\centering \tiny OD-v2~\citep{kar20223d}} & \tiny  \textbf{Recovered} & \multicolumn{1}{m{0.11\textwidth}}{\centering \tiny ZoeD~\citep{bhat2023zoedepth}} & \tiny \textbf{Recovered} & \multicolumn{1}{m{0.11\textwidth}}{\centering \tiny  PD~\citep{bhattad2022cut}} & \tiny  \textbf{Recovered} & \multicolumn{1}{m{0.11\textwidth}}{\centering \tiny  PD~\citep{bhattad2022cut}} & \tiny  \textbf{Recovered} 
\end{tabular}
\caption{Additional results of scene intrinsics extraction from Stable Diffusion UNet (single-step). }
\label{fig:sdsingle_supp}
\end{figure*}

\begin{figure*}[ht]
\centering
\scriptsize
  \setlength\tabcolsep{0pt}
  \renewcommand{\arraystretch}{0}
\begin{tabular}{ccccccccc}
&\multicolumn{2}{c}{Surface Normals} & \multicolumn{2}{c}{Depth} & \multicolumn{2}{c}{Albedo} & \multicolumn{2}{c}{Shading} \\

\includegraphics[width=0.11\linewidth]{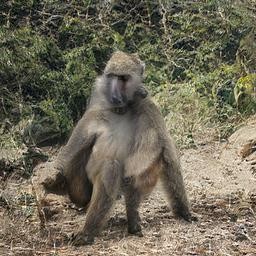} & 
\includegraphics[width=0.11\linewidth]{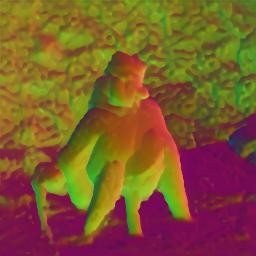} & 
\includegraphics[width=0.11\linewidth]{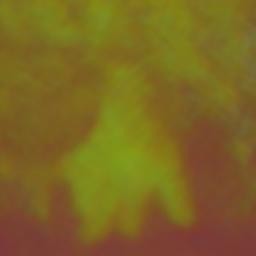} & 
\includegraphics[width=0.11\linewidth]{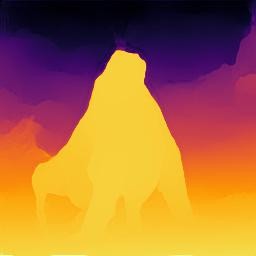} & 
\includegraphics[width=0.11\linewidth]{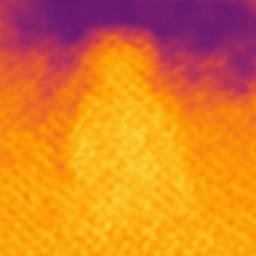} & 
\includegraphics[width=0.11\linewidth]{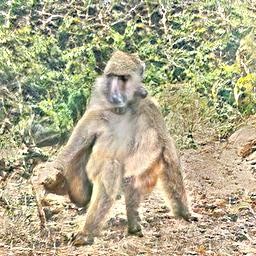} & 
\includegraphics[width=0.11\linewidth]{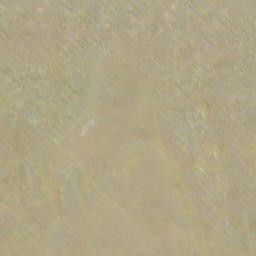} & 
\includegraphics[width=0.11\linewidth]{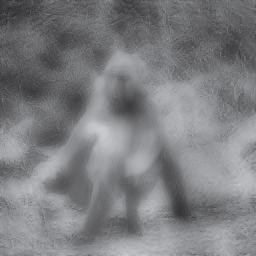} & 
\includegraphics[width=0.11\linewidth]{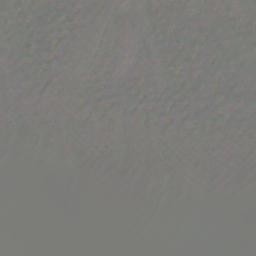} \\
\includegraphics[width=0.11\linewidth]{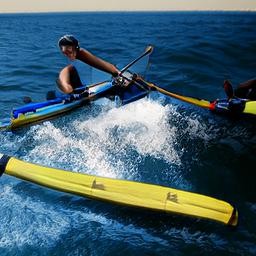} & 
\includegraphics[width=0.11\linewidth]{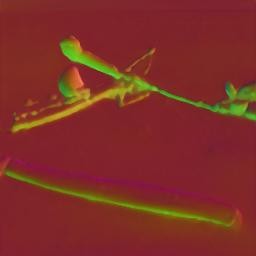} & 
\includegraphics[width=0.11\linewidth]{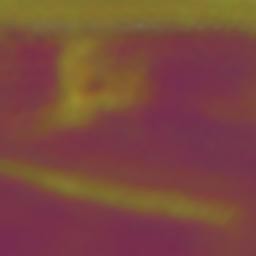} & 
\includegraphics[width=0.11\linewidth]{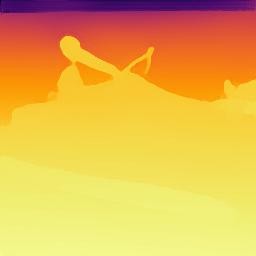} & 
\includegraphics[width=0.11\linewidth]{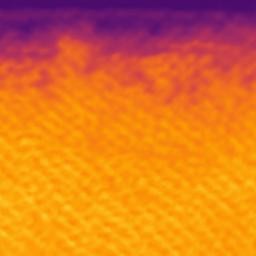} & 
\includegraphics[width=0.11\linewidth]{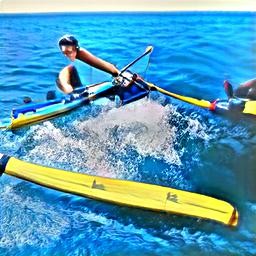} & 
\includegraphics[width=0.11\linewidth]{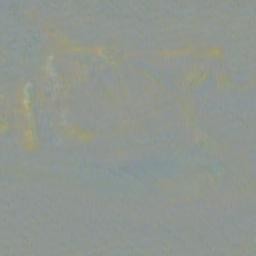} & 
\includegraphics[width=0.11\linewidth]{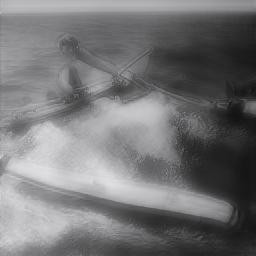} & 
\includegraphics[width=0.11\linewidth]{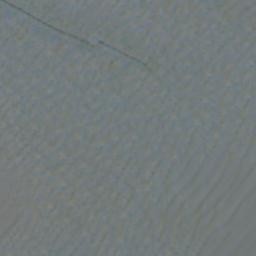} \\
\includegraphics[width=0.11\linewidth]{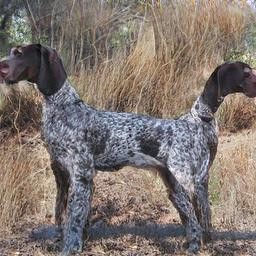} & 
\includegraphics[width=0.11\linewidth]{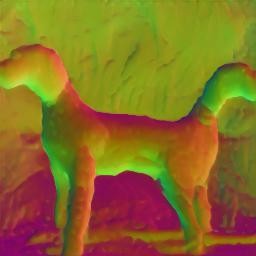} & 
\includegraphics[width=0.11\linewidth]{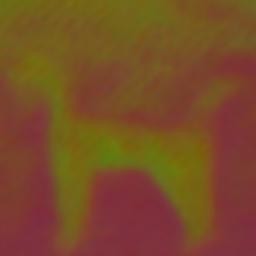} & 
\includegraphics[width=0.11\linewidth]{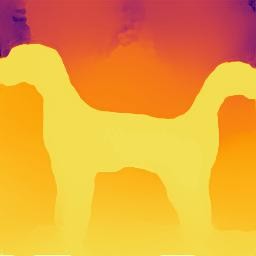} & 
\includegraphics[width=0.11\linewidth]{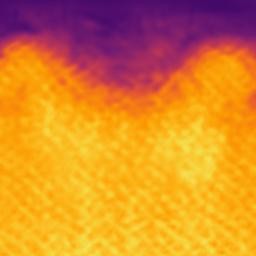} & 
\includegraphics[width=0.11\linewidth]{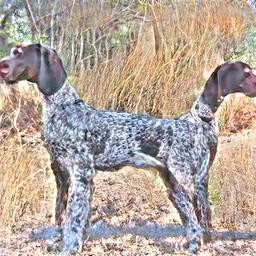} & 
\includegraphics[width=0.11\linewidth]{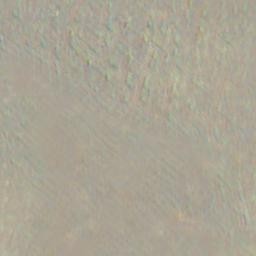} & 
\includegraphics[width=0.11\linewidth]{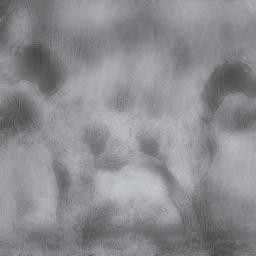} & 
\includegraphics[width=0.11\linewidth]{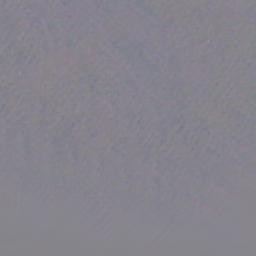} \\
\includegraphics[width=0.11\linewidth]{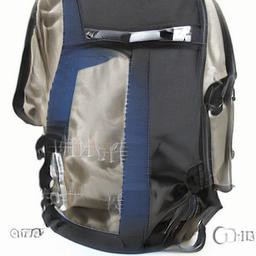} & 
\includegraphics[width=0.11\linewidth]{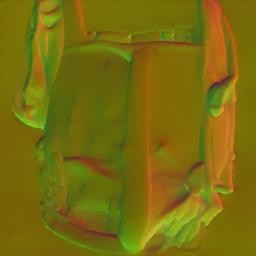} & 
\includegraphics[width=0.11\linewidth]{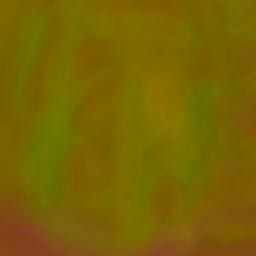} & 
\includegraphics[width=0.11\linewidth]{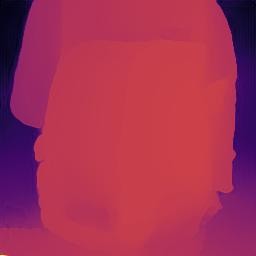} & 
\includegraphics[width=0.11\linewidth]{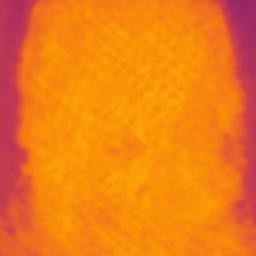} & 
\includegraphics[width=0.11\linewidth]{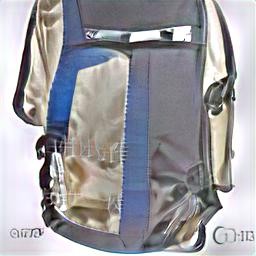} & 
\includegraphics[width=0.11\linewidth]{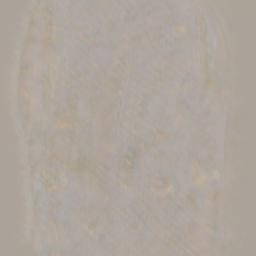} & 
\includegraphics[width=0.11\linewidth]{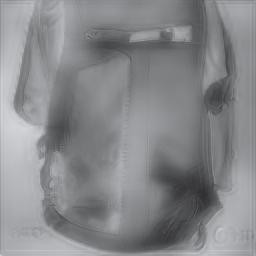} & 
\includegraphics[width=0.11\linewidth]{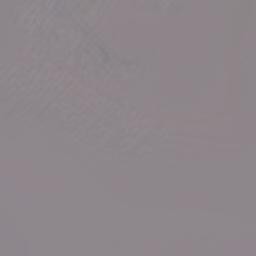} \\
\includegraphics[width=0.11\linewidth]{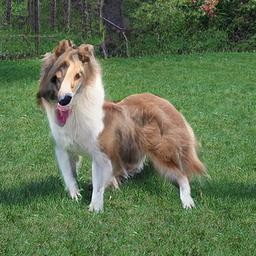} & 
\includegraphics[width=0.11\linewidth]{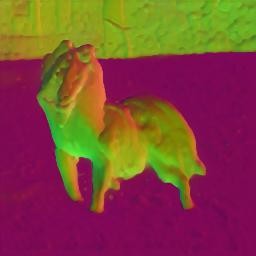} & 
\includegraphics[width=0.11\linewidth]{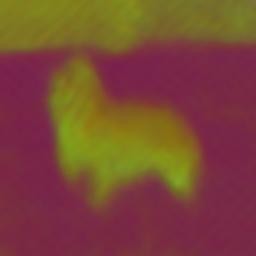} & 
\includegraphics[width=0.11\linewidth]{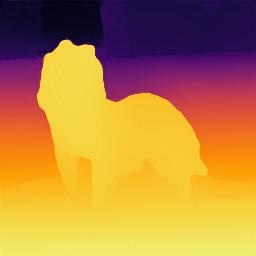} & 
\includegraphics[width=0.11\linewidth]{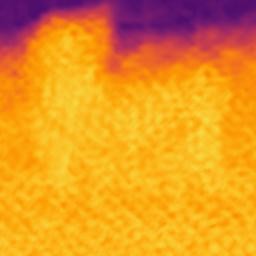} & 
\includegraphics[width=0.11\linewidth]{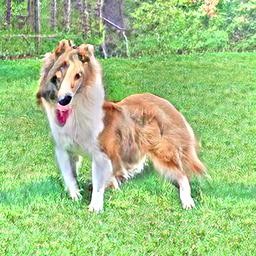} & 
\includegraphics[width=0.11\linewidth]{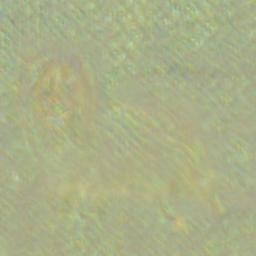} & 
\includegraphics[width=0.11\linewidth]{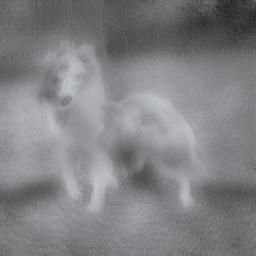} & 
\includegraphics[width=0.11\linewidth]{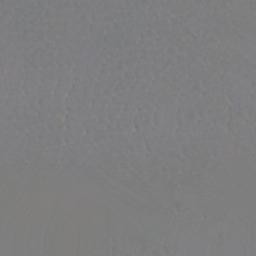}
\\
\vspace{2pt}
\\
\tiny Image & \multicolumn{1}{m{0.11\textwidth}}{\centering \tiny OD-v2~\citep{kar20223d}} & \tiny  \textbf{Recovered} & \multicolumn{1}{m{0.11\textwidth}}{\centering \tiny ZoeD~\citep{bhat2023zoedepth}} & \tiny \textbf{Recovered} & \multicolumn{1}{m{0.11\textwidth}}{\centering \tiny  PD~\citep{bhattad2022cut}} & \tiny  \textbf{Recovered} & \multicolumn{1}{m{0.11\textwidth}}{\centering \tiny  PD~\citep{bhattad2022cut}} & \tiny  \textbf{Recovered} 
\end{tabular}
\caption{Additional results for StyleGAN-XL trained on ImageNet. StyleGAN-XL's inability to produce image intrinsics may be due to its inability to create high-quality plausible images.}
\label{fig:imagenet_supp}
\end{figure*}

In Fig.~\ref{fig:color_shift_supp}, we present more results for \augunet and \augold. Fig.~\ref{fig:generators_comparison_supp} shows extra results for models trained on FFHQ dataset. More examples of scene intrinsics extracted from StyleGAN-v2 trained on LSUN bedroom can be found in Fig.~\ref{fig:bedroom_supp}. In Fig.~\ref{fig:sdsingle_supp}, we show results for SD-UNet (single-step) on generated images. Shown in Fig.~\ref{fig:imagenet_supp} are extra results for StyleGAN-XL trained on ImageNet.

\section{Results on $1024^2$ synthetic images}\label{sec:1024}
\begin{figure*}[ht]
\centering
\scriptsize
\setlength\tabcolsep{0.pt}
\renewcommand{\arraystretch}{0.}
\begin{tabular}{ccc}
\includegraphics[width=0.31\linewidth]{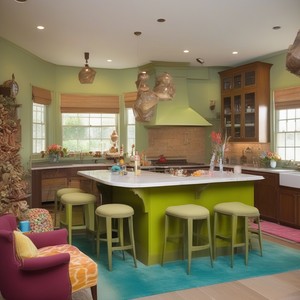} &
  \includegraphics[width=0.31\linewidth]{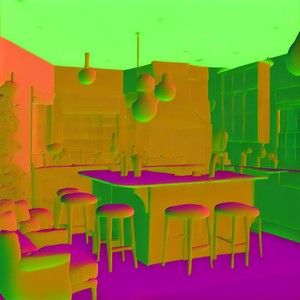} &
  \includegraphics[width=0.31\linewidth]{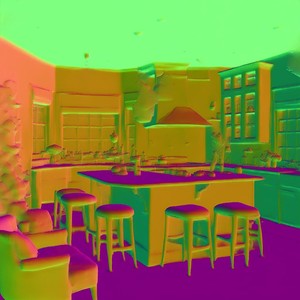} \\
& \includegraphics[width=0.31\linewidth]{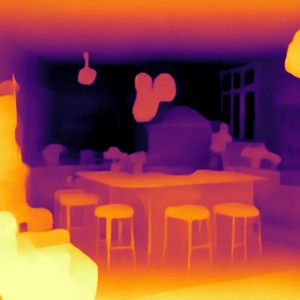} &
  \includegraphics[width=0.31\linewidth]{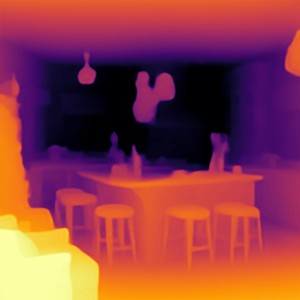} \\
& \includegraphics[width=0.31\linewidth]{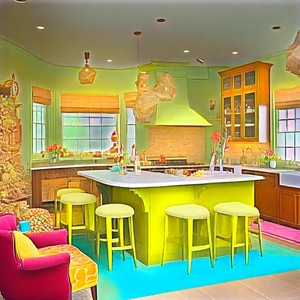} &
  \includegraphics[width=0.31\linewidth]{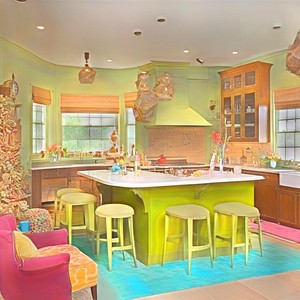} \\
& \includegraphics[width=0.31\linewidth]{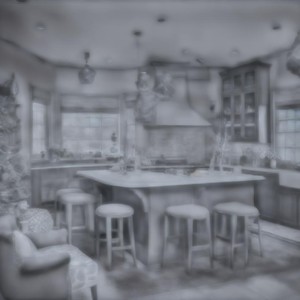} &
  \includegraphics[width=0.31\linewidth]{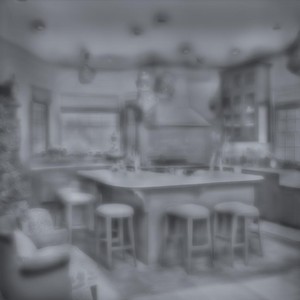}

\end{tabular}
\caption{Results of \augunet models applied on unseen $1024^2$ synthetic images. Left: original image; middle: ours; right: pseudo ground truth.}
\label{fig:1024_results1}
\end{figure*}

\begin{figure*}[ht]
\centering
\scriptsize
\setlength\tabcolsep{0.pt}
\renewcommand{\arraystretch}{0.}
\begin{tabular}{ccc}
\includegraphics[width=0.31\linewidth]{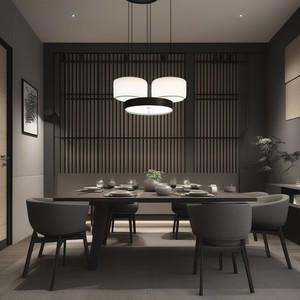} &
  \includegraphics[width=0.31\linewidth]{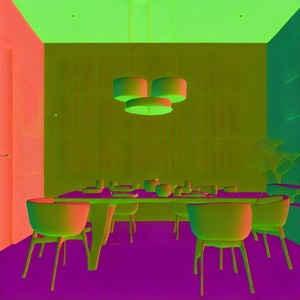} &
  \includegraphics[width=0.31\linewidth]{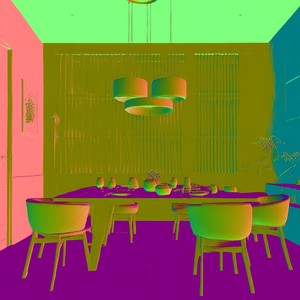} \\
& \includegraphics[width=0.31\linewidth]{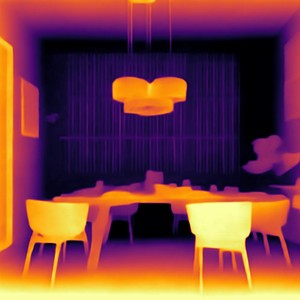} &
  \includegraphics[width=0.31\linewidth]{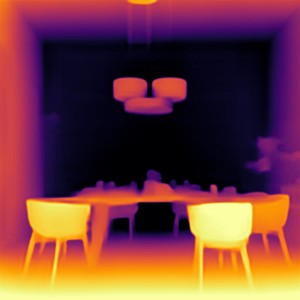} \\
& \includegraphics[width=0.31\linewidth]{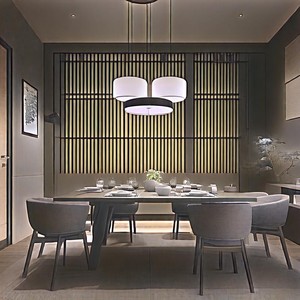} &
  \includegraphics[width=0.31\linewidth]{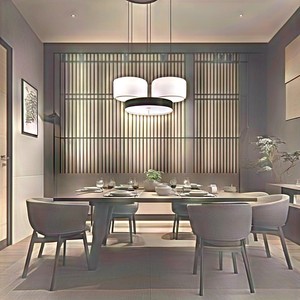} \\
& \includegraphics[width=0.31\linewidth]{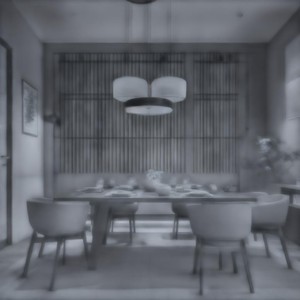} &
  \includegraphics[width=0.31\linewidth]{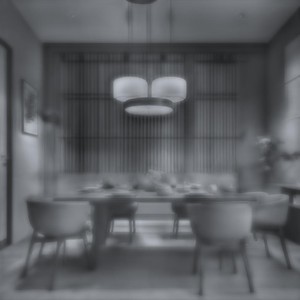}

\end{tabular}
\caption{Cont. results of \augunet models applied on unseen $1024^2$ synthetic images. Left: original image; middle: ours; right: pseudo ground truth.}
\label{fig:1024_results2}
\end{figure*}

\begin{figure*}[ht]
\centering
\scriptsize
\setlength\tabcolsep{0.pt}
\renewcommand{\arraystretch}{0.}
\begin{tabular}{ccc}
\includegraphics[width=0.31\linewidth]{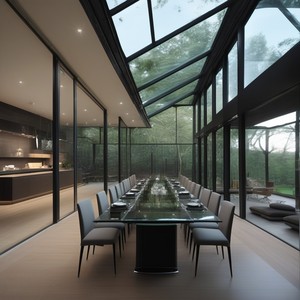} &
  \includegraphics[width=0.31\linewidth]{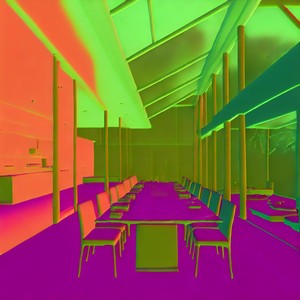} &
  \includegraphics[width=0.31\linewidth]{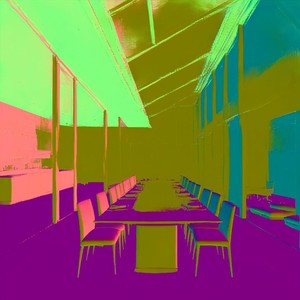} \\
& \includegraphics[width=0.31\linewidth]{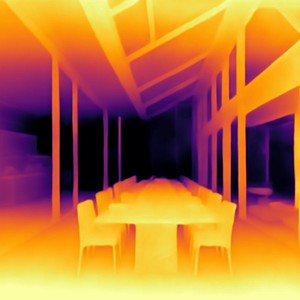} &
  \includegraphics[width=0.31\linewidth]{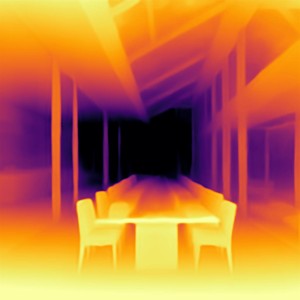} \\
& \includegraphics[width=0.31\linewidth]{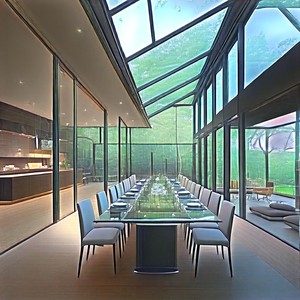} &
  \includegraphics[width=0.31\linewidth]{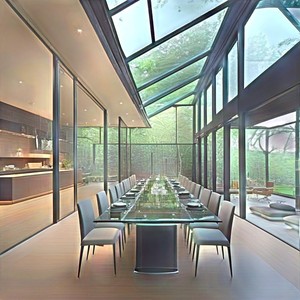} \\
& \includegraphics[width=0.31\linewidth]{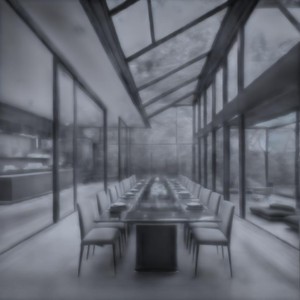} &
  \includegraphics[width=0.31\linewidth]{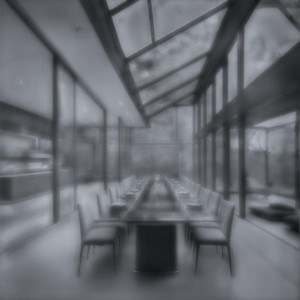}

\end{tabular}
\caption{Cont. results of \augunet models applied on unseen $1024^2$ synthetic images. Left: original image; middle: ours; right: pseudo ground truth.}
\label{fig:1024_results3}
\end{figure*}

\begin{figure*}[ht]
\centering
\scriptsize
\setlength\tabcolsep{0.pt}
\renewcommand{\arraystretch}{0.}
\begin{tabular}{ccc}
\includegraphics[width=0.31\linewidth]{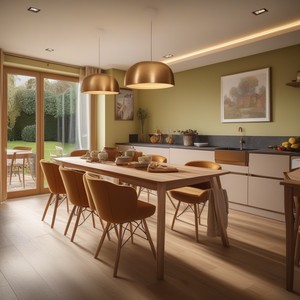} &
  \includegraphics[width=0.31\linewidth]{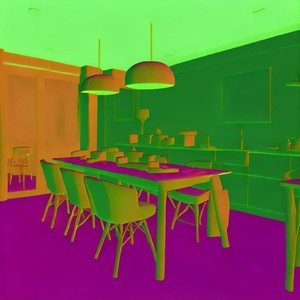} &
  \includegraphics[width=0.31\linewidth]{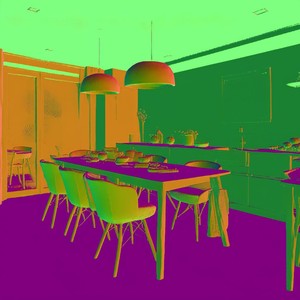} \\
& \includegraphics[width=0.31\linewidth]{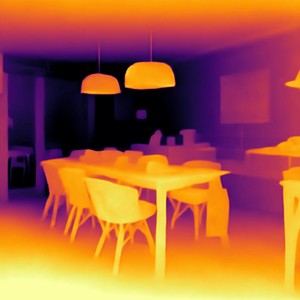} &
  \includegraphics[width=0.31\linewidth]{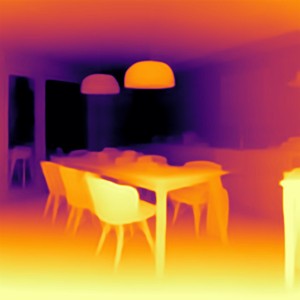} \\
& \includegraphics[width=0.31\linewidth]{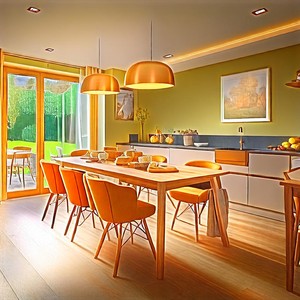} &
  \includegraphics[width=0.31\linewidth]{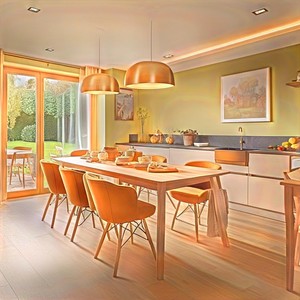} \\
& \includegraphics[width=0.31\linewidth]{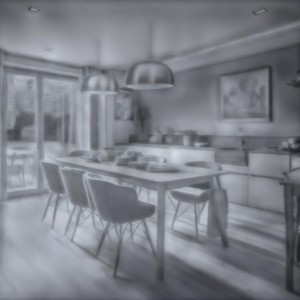} &
  \includegraphics[width=0.31\linewidth]{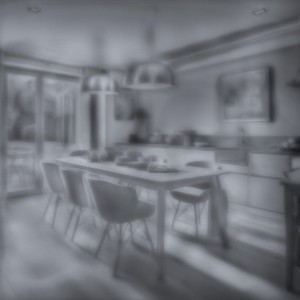}

\end{tabular}
\caption{Cont. results of \augunet models applied on unseen $1024^2$ synthetic images. Left: original image; middle: ours; right: pseudo ground truth.}
\label{fig:1024_results4}
\end{figure*}

\begin{figure*}[ht]
\centering
\scriptsize
\setlength\tabcolsep{0.pt}
\renewcommand{\arraystretch}{0.}
\begin{tabular}{ccc}
\includegraphics[width=0.31\linewidth]{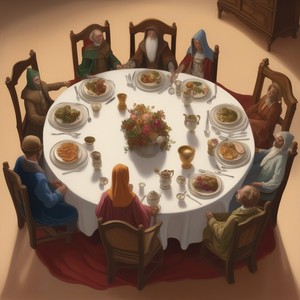} &
  \includegraphics[width=0.31\linewidth]{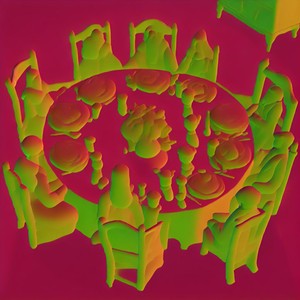} &
  \includegraphics[width=0.31\linewidth]{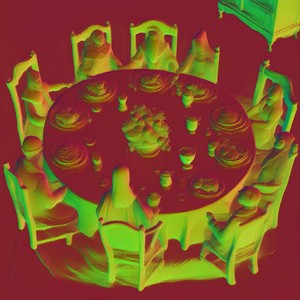} \\
& \includegraphics[width=0.31\linewidth]{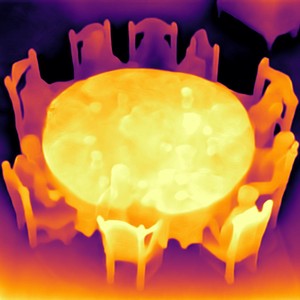} &
  \includegraphics[width=0.31\linewidth]{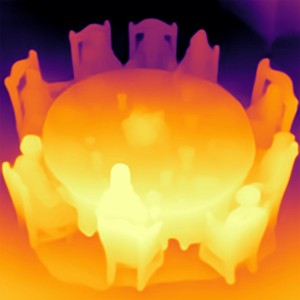} \\
& \includegraphics[width=0.31\linewidth]{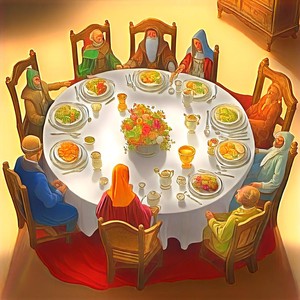} &
  \includegraphics[width=0.31\linewidth]{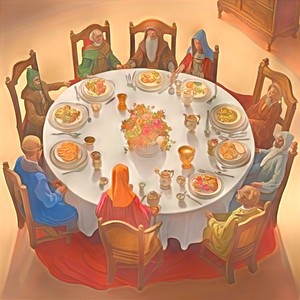} \\
& \includegraphics[width=0.31\linewidth]{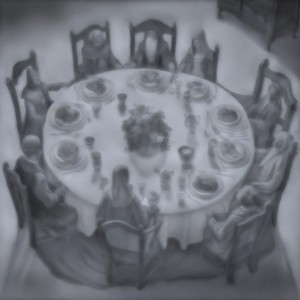} &
  \includegraphics[width=0.31\linewidth]{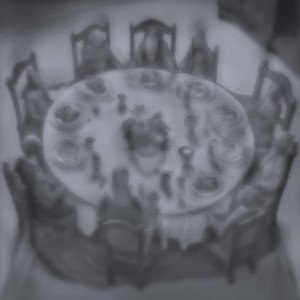}

\end{tabular}
\caption{Cont. results of \augunet models applied on unseen $1024^2$ synthetic images. Left: original image; middle: ours; right: pseudo ground truth.}
\label{fig:1024_results5}
\end{figure*}

\begin{figure*}[ht]
\centering
\scriptsize
\setlength\tabcolsep{0.pt}
\renewcommand{\arraystretch}{0.}
\begin{tabular}{ccc}
\includegraphics[width=0.31\linewidth]{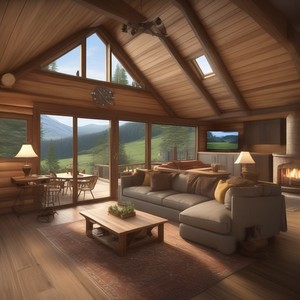} &
  \includegraphics[width=0.31\linewidth]{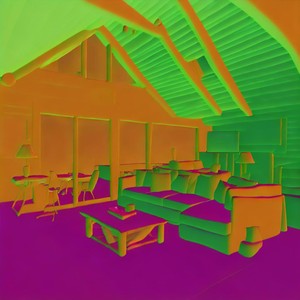} &
  \includegraphics[width=0.31\linewidth]{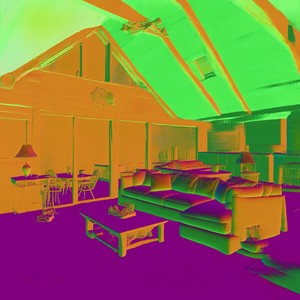} \\
& \includegraphics[width=0.31\linewidth]{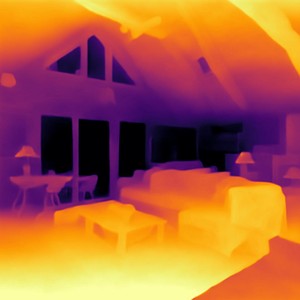} &
  \includegraphics[width=0.31\linewidth]{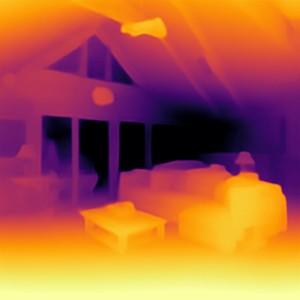} \\
& \includegraphics[width=0.31\linewidth]{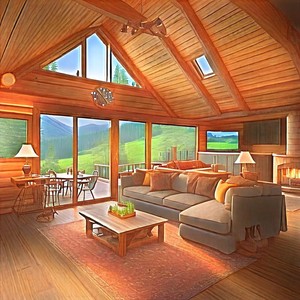} &
  \includegraphics[width=0.31\linewidth]{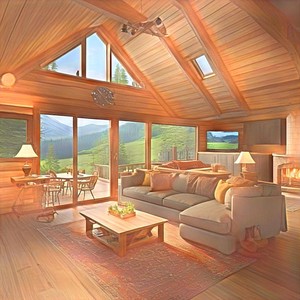} \\
& \includegraphics[width=0.31\linewidth]{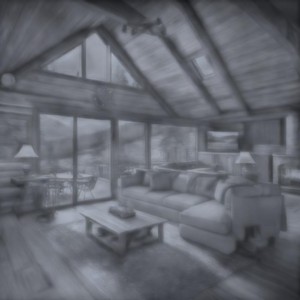} &
  \includegraphics[width=0.31\linewidth]{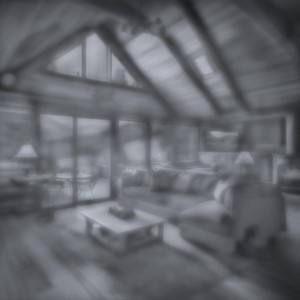}

\end{tabular}
\caption{Cont. results of \augunet models applied on unseen $1024^2$ synthetic images. Left: original image; middle: ours; right: pseudo ground truth.}
\label{fig:1024_results6}
\end{figure*}

\begin{figure*}[ht]
\centering
\scriptsize
\setlength\tabcolsep{0.pt}
\renewcommand{\arraystretch}{0.}
\begin{tabular}{ccc}
\includegraphics[width=0.31\linewidth]{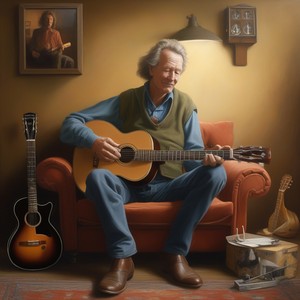} &
  \includegraphics[width=0.31\linewidth]{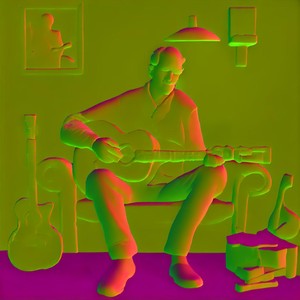} &
  \includegraphics[width=0.31\linewidth]{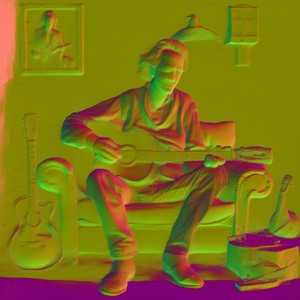} \\
& \includegraphics[width=0.31\linewidth]{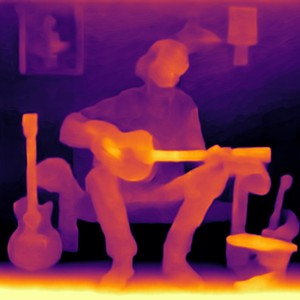} &
  \includegraphics[width=0.31\linewidth]{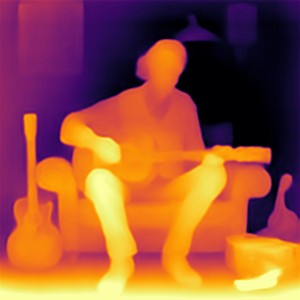} \\
& \includegraphics[width=0.31\linewidth]{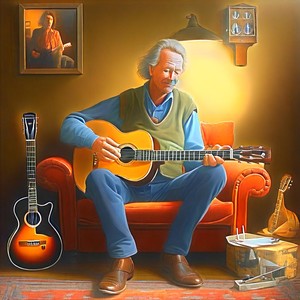} &
  \includegraphics[width=0.31\linewidth]{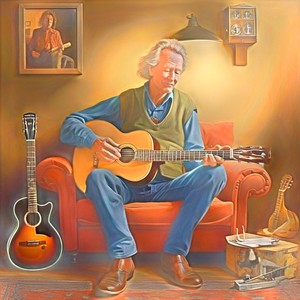} \\
& \includegraphics[width=0.31\linewidth]{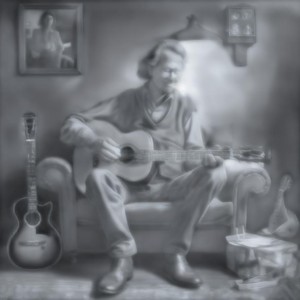} &
  \includegraphics[width=0.31\linewidth]{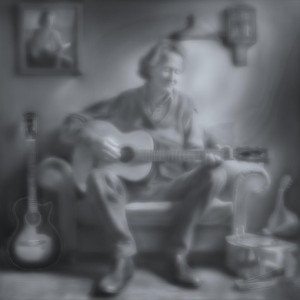}

\end{tabular}
\caption{Cont. results of \augunet models applied on unseen $1024^2$ synthetic images. Left: original image; middle: ours; right: pseudo ground truth.}
\label{fig:1024_results7}
\end{figure*}

\begin{figure*}[ht]
\centering
\scriptsize
\setlength\tabcolsep{0.pt}
\renewcommand{\arraystretch}{0.}
\begin{tabular}{ccc}
\includegraphics[width=0.31\linewidth]{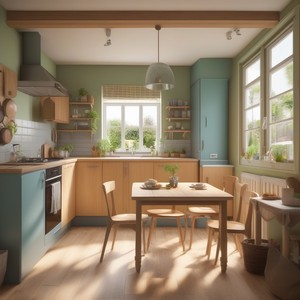} &
  \includegraphics[width=0.31\linewidth]{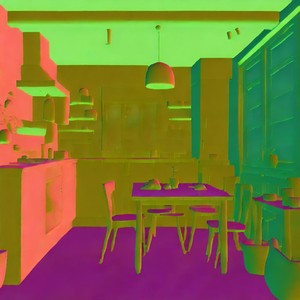} &
  \includegraphics[width=0.31\linewidth]{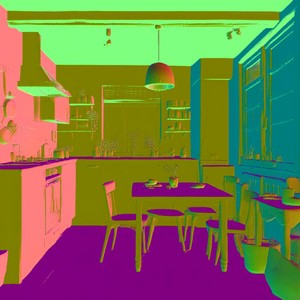} \\
& \includegraphics[width=0.31\linewidth]{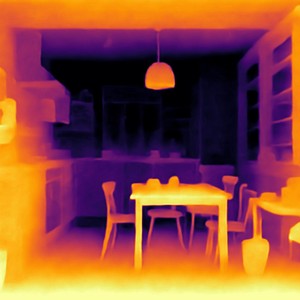} &
  \includegraphics[width=0.31\linewidth]{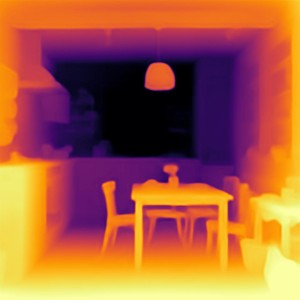} \\
& \includegraphics[width=0.31\linewidth]{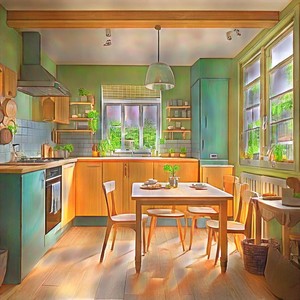} &
  \includegraphics[width=0.31\linewidth]{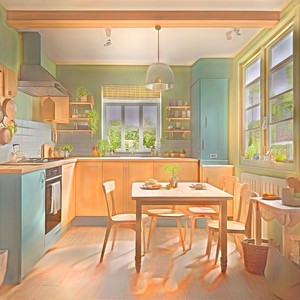} \\
& \includegraphics[width=0.31\linewidth]{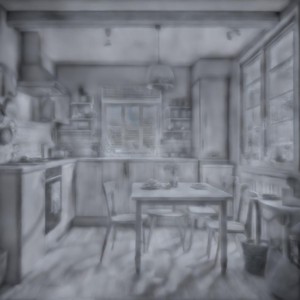} &
  \includegraphics[width=0.31\linewidth]{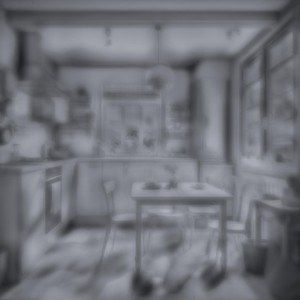}

\end{tabular}
\caption{Cont. results of \augunet models applied on unseen $1024^2$ synthetic images. Left: original image; middle: ours; right: pseudo ground truth.}
\label{fig:1024_results8}
\end{figure*}

\begin{figure*}[ht]
\centering
\scriptsize
\setlength\tabcolsep{0.pt}
\renewcommand{\arraystretch}{0.}
\begin{tabular}{ccc}
\includegraphics[width=0.31\linewidth]{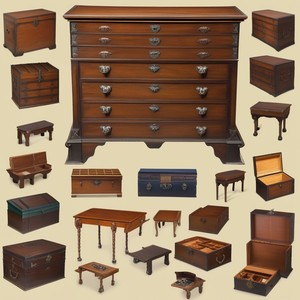} &
  \includegraphics[width=0.31\linewidth]{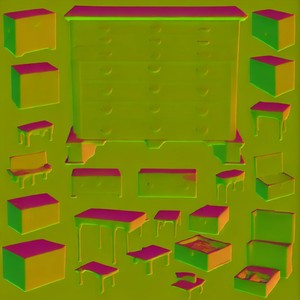} &
  \includegraphics[width=0.31\linewidth]{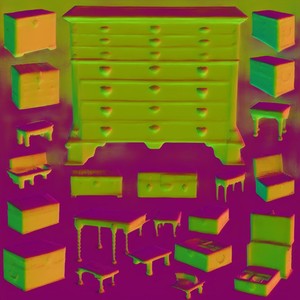} \\
& \includegraphics[width=0.31\linewidth]{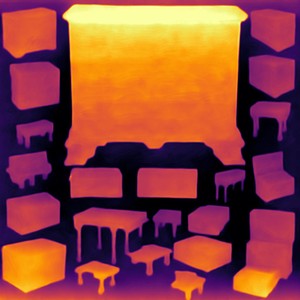} &
  \includegraphics[width=0.31\linewidth]{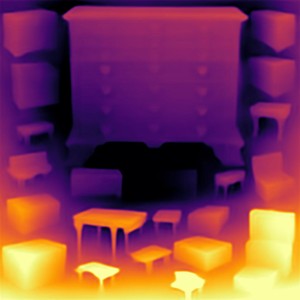} \\
& \includegraphics[width=0.31\linewidth]{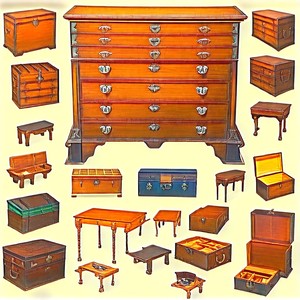} &
  \includegraphics[width=0.31\linewidth]{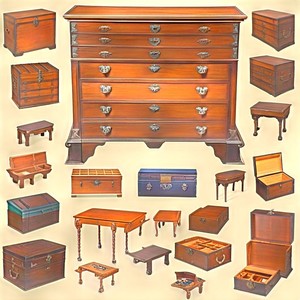} \\
& \includegraphics[width=0.31\linewidth]{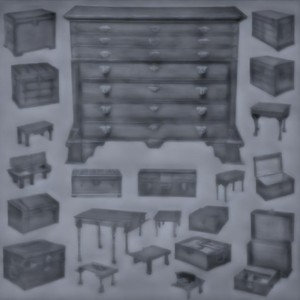} &
  \includegraphics[width=0.31\linewidth]{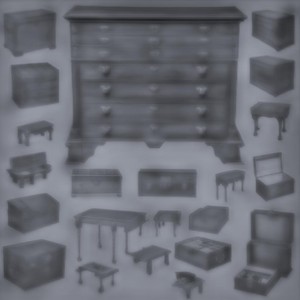}

\end{tabular}
\caption{Cont. results of \augunet models applied on unseen $1024^2$ synthetic images. Left: original image; middle: ours; right: pseudo ground truth.}
\label{fig:1024_results9}
\end{figure*}

\begin{figure*}[ht]
\centering
\scriptsize
\setlength\tabcolsep{0.pt}
\renewcommand{\arraystretch}{0.}
\begin{tabular}{ccc}
\includegraphics[width=0.31\linewidth]{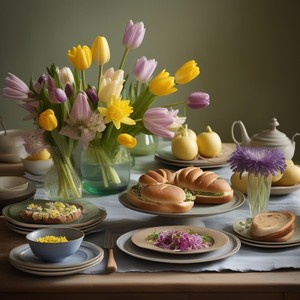} &
  \includegraphics[width=0.31\linewidth]{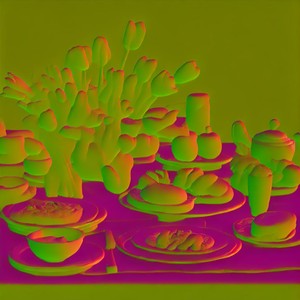} &
  \includegraphics[width=0.31\linewidth]{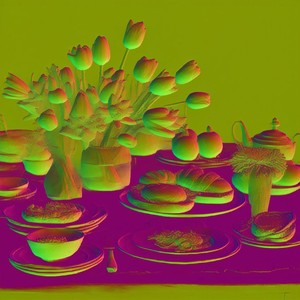} \\
& \includegraphics[width=0.31\linewidth]{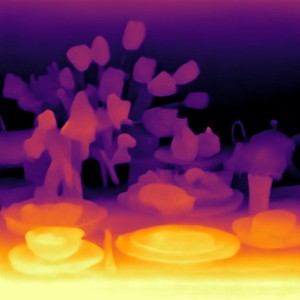} &
  \includegraphics[width=0.31\linewidth]{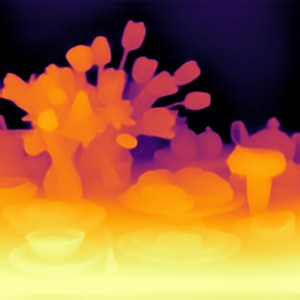} \\
& \includegraphics[width=0.31\linewidth]{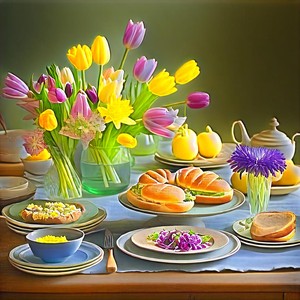} &
  \includegraphics[width=0.31\linewidth]{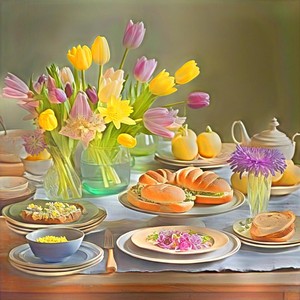} \\
& \includegraphics[width=0.31\linewidth]{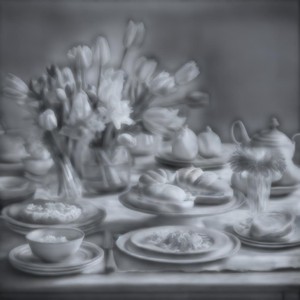} &
  \includegraphics[width=0.31\linewidth]{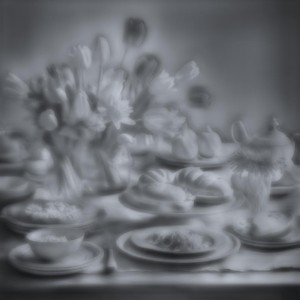}

\end{tabular}
\caption{Cont. results of \augunet models applied on unseen $1024^2$ synthetic images. Left: original image; middle: ours; right: pseudo ground truth.}
\label{fig:1024_results10}
\end{figure*}

Our multi-step \augunet models, although trained exclusively on 
$512^2$ 
  images from the DIODE dataset, demonstrate their robustness by successfully extracting intrinsic images from 
$1024^2$
  high-resolution synthetic images generated by Stable Diffusion XL~\citep{podell2023sdxl}, as shown across Figures \ref{fig:1024_results1} to \ref{fig:1024_results10}.

\end{document}